# Time-Series Forecasting in Smart Manufacturing Systems: An Experimental Evaluation of the State-of-the-art Algorithms

Mojtaba A. Farahani[ab], Fadi El Kalach[b], Austin Harper[a], M. R. McCormick[ab], Ramy Harik[c], and Thorsten Wuest[b1]

[a] West Virginia University, Morgantown, 26505 WV, U.S.A
[b] University of South Carolina, Columbia, 29208, SC, U.S.A
[c] Clemson University, Clemson, 29634, SC, U.S.A

# Abstract

Time-Series Forecasting (TSF) is a growing research domain across various domains including manufacturing. Manufacturing can benefit from Artificial Intelligence (AI) and Machine Learning (ML) innovations for TSF tasks. Although numerous TSF algorithms have been developed and proposed over the past decades, the critical validation and experimental evaluation of the algorithms hold substantial value for researchers and practitioners and are missing to-date. This study aims to fill this research gap by providing a rigorous experimental evaluation of the state-of-the-art TSF algorithms on thirteen manufacturing-related datasets with a focus on their applicability in smart manufacturing environments. Each algorithm was selected based on the defined TSF categories to ensure a representative set of state of the art algorithms. The evaluation includes different scenarios to evaluate the models using combinations of two problem categories (univariate and multivariate) and two forecasting horizons (short- and long-term). To evaluate the performance of the algorithms, the weighted average percent error was calculated for each application, and additional post hoc statistical analyses were conducted to assess the significance of observed differences. Only algorithms with accessible codes from open-source libraries were utilized, and no hyperparameter tuning was conducted. This approach allowed us to evaluate the algorithms as "out-of-the-box" solutions that can be easily implemented, ensuring their usability within the manufacturing sector by practitioners with limited technical knowledge of ML algorithms. This aligns with the objective of facilitating the adoption of these techniques in Industry 4.0 and smart manufacturing systems. Based on the results, transformer and MLP-based architectures demonstrated the best performance across different scenarios with MLP-based architecture winning the most scenarios. For univariate TSF, PatchTST emerged as the most robust algorithm, particularly for long-term horizons, while for multivariate problems, MLP-based architectures like N-HITS and TiDE showed superior results. The study revealed that simpler algorithms like XGBoost could outperform more complex transformer-based in certain tasks. These findings challenge the assumption that more sophisticated models inherently produce better results. Additionally, the research highlighted the importance of computational resource considerations, showing significant variations in runtime and memory usage across different algorithms.

[1] Corresponding Author: twuest@sc.edu (T. Wuest)

# 1. Introduction

Forecasting can be traced back to the first hunters when they looked at the sky and attempted to forecast tomorrow's temperature and determine whether it would be a good day for a hunt [1]. Forecasting has evolved dramatically since then and is particularly relevant in today's manufacturing landscape. Manufacturing industries are in the midst of digital transformation toward Industry 4.0 and smart manufacturing systems, built on three fundamental pillars: *Connectivity*, *Virtualization*, and *Data Utilization*. Connectivity facilitates the collection, communication, and storage of data from machines and operations. Virtualization creates digital assets of physical objects, enabling transparent communications between operators, operations, and decision-makers. Data utilization translates large quantities of information into meaningful insights for industrial applications [2]. While each pillar is individually crucial, their interconnectedness defines both their benefits and limitations. For example, data utilization relies on connectivity for information aggregation and transportation while requiring virtualization to bridge the gap between digital and physical assets for timely decision-making.

Within this data-driven manufacturing environment, Machine Learning (ML) and Artificial Intelligence (AI) have become invaluable in creating actionable insights for informed decision-making [3]. Time-Series Forecasting (TSF) models leverage manufacturing data, e.g., by using Statistical, ML, regression or Deep Learning (DL) algorithms, to predict future values or series of values [4]. Time-series data can be classified into two stochastic models: stationary and nonstationary series. Stationary models assume a constant process mean level while nonstationary models have a shifting mean over time. Once the stochastic model–usually nonstationary for industrial datasets–has been identified, an appropriate forecasting method can be selected [5]. When TSF models are developed with proper consideration of these stochastic foundations, their predictions are likely to better represent future data.

The flexibility of TSF has led to its adoption across diverse sectors, including healthcare [6], retail [7], finance [8], and utility networks such as electrical grids [9], telecommunications [10], and wastewater treatment [11]. TSF is also ripe for use in industrial environments due to the existing prevalence of time series data collection in manufacturing. Several production uses for this technology have been predicting customer demand for production planning [12], resource allocation [13], equipment maintenance [14], inventory requirements, and supply chain logistics [13]. In other cases, this technology can be applied to predict the Remaining Useful Life (RUL) of equipment, equipment status, in-situ production monitoring, and control, among countless other applications directly on the shop floor. The use of this technology across a breadth of industries facilitates improvements in predictive modeling for easier implementation in all fields.

Various researchers have conducted general literature reviews on TSF topics from a variety of different lenses. For instance, advances in TSF models using Artificial Neural Networks (ANN) were studied by Tealab A. in a systematic literature review [15]. Masini et al. surveyed advanced high-dimensional models for TSF such as neural networks and tree-based methods [16]. An extensive literature review on the last 50 years of forecasting combinations with references to available open-source repositories was conducted by Wang et al. [17], and DL methods for TSF was reviewed by Benidis et al. [18]. Other researchers focused on TSF tasks for specific problems and domains. These studies include the investigation of DL methods for financial time-series [8], and ML techniques for time-series energy consumption forecasting [19] among others. In the manufacturing domain, RUL prediction [20], [21] and regression tasks in predictive maintenance [22] received significant attention among reviews.

Although these reviews have holistic value and contribute to the domain's body of knowledge, none of them evaluate the performance of TSF algorithms on time-series datasets. Algorithm validation and empirical comparisons hold substantial value for practitioners by streamlining choices and revealing insights into the benefits and limitations of each model. Such evaluations were scarce among the existing literature surveyed for this study. Ahmed et al. compared several ML algorithms on the M3 competition dataset [23], and Makridakis et al. evaluated the performance of statistical, ML, and DL algorithms on the M3 competition dataset [24], [25]. Moreover, an empirical study on Recurrent Neural Networks (RNN) was conducted by Hewamalage [26], and Lara-Benıtez performed a thorough analysis of seven types of deep learning models on twelve TSF datasets [27]. Most of these papers either focus on a specific group of algorithms or perform the evaluation on a limited set of generic datasets. There were no studies which examined the TSF task from a comprehensive manufacturing perspective.

This paper addresses two significant research gaps: i) to date, there is no comprehensive experimental evaluation of TSF algorithms in a manufacturing context, and ii) to date, a comprehensive list of manufacturing datasets which can be utilized for TSF applications has not been enumerated. Our objectives are threefold: first, to explore and categorize state-of-the-art ML, and DL TSF algorithms; second, to introduce and compile a set of manufacturing-related public datasets to address the domain-specific data scarcity; and finally, to evaluate selected algorithms across various manufacturing scenarios, providing practical insights into their comparative performance. Ideally, we are able to identify a set of algorithms that would work "out-of-the-box" on a wide range of problems with limited technical knowledge and hyperparameter tuning while maintaining acceptable forecasting error. This approach aims to provide practitioners with clear, hands-on, quantitative guidance for the data-driven selection of appropriate TSF methods for specific manufacturing applications while considering performance and implementation complexity.

*The scope of this study* is to fill the identified gap by addressing mentioned objectives. It should be noted that the detailed theoretical explanation of specific algorithms is out of the scope of this study and we suggest consulting the references provided for a deeper dive if that level of detail for a particular algorithm is desired. To the best of our knowledge, this is the first study of this kind in the manufacturing domain. This study's novelty lies in pioneering this type of research for the smart manufacturing domain, setting a benchmark for quantitative comparison and providing out-of-the-box solutions for practitioners in the field. Furthermore, this work offers guidance on the application of TSF algorithms and establishes a connection between otherwise potentially isolated domains of manufacturing and general TSF. More specifically, we aim to answer the following research questions with this study

- RQ1: What are the characteristics of public datasets in manufacturing-related applications that can be utilized for TSF tasks, and different preprocessing methods can be applied to prepare them for TSF experiments?
- RQ2: What are the state-of-the-art algorithms for TSF tasks applicable to manufacturing datasets, and what features make them particularly well-suited for the manufacturing domain?

- RQ3: Which algorithms demonstrate superior performance on the mentioned datasets and across different TSF tasks? How do state-of-the-art ML and DL algorithms perform compared to benchmark and statistical methods?

The remainder of this paper is structured as follows. *Section Two* provides the necessary background and definitions used in this research, defines the addressed problems, and provides an overview of statistical, ML, and DL algorithms and public datasets that can be used for TSF tasks in manufacturing research. *Section Three* provides an overview of the material and methods used for this study. It describes the experimental evaluation methodology of this study, which led to the final selection of datasets and algorithms, preprocessing steps, as well as the comparison and evaluation metrics. In *Section Four*, the results of the experiments are presented with different scenarios based on defined evaluation metrics as well as additional discussion which highlights challenges and reflects on the limitations faced during this study. Finally, in *Section Five*, the conclusions are drawn and recommendations are provided for future works.

# 2. Background and Definitions

This section details the problems to be addressed and provides essential background information and definitions that will be used throughout this paper to communicate the investigation's findings.

A time-series is defined as a sequence of ordered real-valued observations that are recorded over a fixed interval of time. A univariate time-series is defined as $x = \{x_1, x_2, ..., x_L\}$ where L is the time-series length whereas a multivariate time-series $X = [x^1, x^2, ..., x^M]^{Transposed}$ is a matrix consisting of $M$ univariate time-series recorded simultaneously and $x^i \in R^i$. When M=1 the time-series is a univariate time-series with length L.

## 2.1 Problem Statement

Within time-series analytics several supervised predictive tasks can be formulated which serve as the basis for understanding how to deploy AI algorithms. Table 1 displays a brief overview of these tasks and their characteristics.

Table 1. Supervised predictive tasks within Time-Series Analytics.

| Dataset Name | Input Data | Output Data | Example |
|---|---|---|---|
| Time-Series Classification (TSC) | Time-Series | Discrete Values | Fault Classification |
| Time-Series Forecasting (TSF) | Time-Series | Continuous Values | Energy consumption Forecasting |
| Time-Series Regression (TSR) | Time-Series | Continuous Values | RUL prediction |

TSC is a predictive task that aims to learn time-series data patterns and categorize them into discrete labeled classes [28]. TSR tasks predict continuous numeric values as the dependent variable instead of static features and depend on the time-series data as independent variables. The difference between the two is that TSC finds the approximation function to map a time series to a finite set of discrete labels while TSR predicts a continuous value from the time series [29]. For instance, TSC might classify a vibration signal collected from a bearing in a manufacturing machine as faulty or normal [30], [31], while TSR could be utilized to predict a quantitative value such as the RUL of the bearing or a machine tool based on vibrational patterns and temperature signals [22], [32].

On the other hand, TSF aims to predict future values of a time-series based on historical values of itself and/or exogenous variables. It involves fitting a function to a time-series to predict its future values by *extrapolating into the future*. Instead of using the whole time-series history, this is usually done by considering a window of past values known as *lag or lookback window*. This window is used because typically recent observations are better indicators of the future than those from the distant past. It is also more computationally expensive to use the entire history of data, specifically in high-resolution time-series with high sampling frequency.

The main idea behind TSF is that hidden patterns in historical data of a known variable can be used to predict future values of that variable. Predicting the exact future values of the target time-series is known as a *point forecast*. However, another option is to provide a prediction interval, percentile, or prediction distribution around the forecasted value which is referred to as *probabilistic forecasting*.

On top of these considerations it is essential to examine the nature of the input variables and their relation to the target variable. *Univariate forecasts* are generated by using the information from historical values of the target time-series without leveraging cross-learning from other variables. While *multivariate forecasting* may use other time-series variables to forecast the target variable similar to time-series regression. These variable relations and other input types may be linear or involve nonlinear structures. An example of multivariate forecasting is predicting the future temperature values of a manufacturing process by utilizing the historical values of the temperature and other variables such as air temperature, rotational speed, torque, etc. [1]. Figure 1 depicts these three principal concepts and their differences.

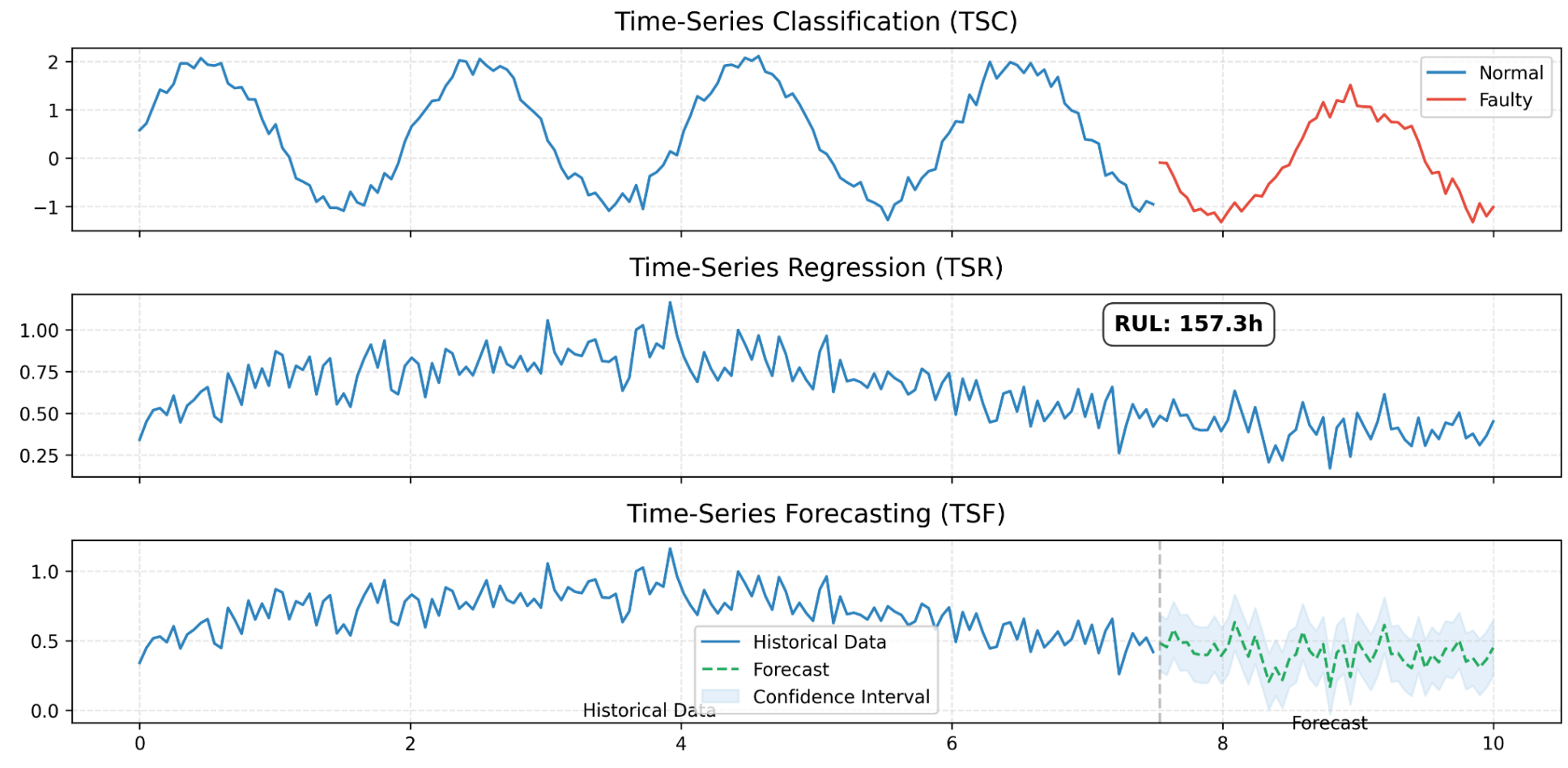

Figure 1. Time-Series analysis tasks of TSC, TSR, and TSF

This study focuses on two main problems of point TSF over four different scenarios, defined below. All of these problems and scenarios are being looked at through the manufacturing lens, by considering the type and characteristics of time-series data that might be collected in a manufacturing environment or industrial setup.

### 2.1.1. Problem 1: Univariate Time-Series Forecasting

Univariate TSF problems refer to predicting future values of one time-series variable based on its past values. Given a time-series dataset and the target variable $x = \{x_1, x_2, ..., x_L\}$, the dataset is divided into the training set $x_{train} = \{x_1, x_2, ..., x_t\}$ and the test set $x_{test} = \{x_{t+1}, x_{t+2}, ..., x_L\}$ at timestamp $t$. The problem is formulated as equation 1. In this study's experimental evaluations, two scenarios of **Short-term Univariate TSF** and **Long-term Univariate TSF** are defined for this problem.

$$\{x_{t+1}, x_{t+2}, ..., x_L\} = F(x_{t-T+1}, x_{t-T+2}, ..., x_t) + \epsilon \quad (1)$$

### 2.1.2. Problem 2: Multivariate Time-Series Forecasting

Similar to problem 1, a multivariate TSF problem refers to predicting the future values of the target variable $x = \{x_1, x_2, ..., x_L\}$ based on its past values and a set of exogenous time-series variables $x^1, x^2, ..., x^{M-1}$. In this problem, the multivariate forecasting problem is formulated as equation 2. In this study's experimental evaluations, two scenarios of **Short-term Multivariate TSF** and **Long-term Multivariate TSF** are defined for this problem.

$$\{x_{t+1}, x_{t+2}, ..., x_L\} = F(x_{t-T+1}, x_{t-T+2}, ..., x_t) + F(x^1, x^2, ..., x^{M-1}) + \epsilon \quad (2)$$

In both problems, **F** is the function approximated by fitting the model to the training set. $\epsilon$ denotes the error associated with the function approximation **F**. **M** is the number of time-series variables, and **T** is the lookback window. The model forecasts **H** timestamp each instance where H refers to the forecasting horizon. In case the test set length is bigger than the forecasting horizon **L > H**, the forecast values of the test $\{x_{t+1}, x_{t+2}, ..., x_L\}$ are generated at **N** steps where at each step the next forecasting horizon **H** is forecasted until all forecasted values for the test are generated. The goal is to achieve this by minimizing the prediction error which is the residual of the forecasted value reduced from the actual value [27], [33].

It should be noted that in this problem, only the target variable future values are being forecasted, but cross-series information is being leveraged to forecast the target variable. The idea is to develop a model by exploiting information from many time-series simultaneously. The value of one time-series is driven by other external time-varying variables, instead of developing one model for each time-series in a dataset. This approach has received increased attention recently [26] but there are other versions of multivariate TSF in the literature that are not considered in this study.

## 2.2. Time-Series Forecasting Techniques

Statistical, ML, and DL techniques are main groups of algorithms for tackling TSF problems. ML and DL techniques have received significant attention in recent years for different TSF problems and tasks. Their strength compared to statistical techniques are that there are few or no assumptions about the input data; patterns are identified and approximated automatically; they do not require multiple preprocessing and transformation steps; and no assumptions are needed regarding seasonality, trend and data distribution [34]. While these techniques are very powerful, traditional statistical methods are still widely used and even outperform ML algorithms in many studies [25]. Thus, it is important to use statistical methods as a baseline for performance comparison in novel studies to ensure improvement.

### 2.2.1. Statistical Techniques

Before the advent of ML and DL techniques, the models used for TSF tasks were mainly statistical techniques. These techniques are based on estimating linear functions from recent historical data. The advantages of these methods are that they work well with small datasets and have less parameters to choose from compared to complex ML and DL algorithms. However, there are several downsides to these methods that limit their applicability in many situations. Statistical methods often fail when they are applied directly to the dataset, without handling seasonality, trend, and applying preprocessing transformations. Moreover, these methods are being locally built for each individual time-series and they lack the cross-learning ability to share learning across different time-series instances. Therefore, it is not feasible to use these methods for very large datasets as they would require several iterations of retraining [26], [27], [35].

Statistical techniques can be categorized into different groups. Exponential smoothing models use weighted averages of past observations, with weights decreasing exponentially over time. Examples of these models include Simple Exponential Smoothing (SES) [36], Holt's method [37], and the Holt-Winters method for seasonal data [38]. Time-series regression models predict a target variable using linear relationships with other variables as regressors. The Theta model combines the last observation with various trend functions to generate forecasts [39]. AutoRegressive Integrated Moving Average (ARIMA) [5] models are the most widely used statistical methods. These models combine autoregression, differencing, and moving averages, and can be extended to handle seasonal data (SARIMA) or external variables (ARIMAX), among many other variants. Additionally, there are several specialized models that can be used for specific situations. For example, TBATS combines Fourier transforms with exponential smoothing and was designed to forecast time-series with multiple seasonal periods [40]. Prophet was designed to handle multiple seasonalities (daily, weekly, and yearly patterns) and enable forecasting at scale [41].

### 2.2.2. Machine Learning Techniques

Despite their success in machine learning tasks with tabular data, traditional ML techniques like Support Vector Machines (SVMs) and Random Forest (RF) face challenges in TSF tasks. Most ML algorithms assume independent and identically distributed (i.i.d.) instances, which conflicts with the sequential and dependent nature of time-series data. Furthermore, feature engineering for time-series data becomes complex, requiring consideration of temporal dependencies, lag effects, and seasonal patterns that may not be easily discernible. Nevertheless, researchers have successfully applied these methods to TSF tasks either directly [19] or as part of a hybrid method [42], demonstrating their relevance despite their limitations and their advantage in specific applications.

### 2.2.3. Artificial Neural Network and Deep Learning Techniques

Despite traditional ML techniques, there is an extensive body of research on ANN and DL techniques for TSF tasks due to their ability to automatically learn temporal dependencies and model non-linear patterns. MLP-based, RNN-based, CNN-based, Transformer-based, and LLMs are the four main groups of algorithms that have been proposed for different TSF tasks.

MLP-based architectures are the earliest attempt to use neural networks for TSF. They are often based on multiple stacks of fully connected dense layers with various residual connections and nonlinear activation, such as ReLU where feed-forward networks process fixed-size windows of time-series data. Their simplicity and computational efficiency make them a suitable choice for different scenarios with limited data and computational resources. While these models technically lack explicit temporal modeling mechanisms, they can capture complex non-linear relationships between the input series treated as independent features and the target variable. Neural Basis Expansion Analysis for interpretable TimeSeries (N-BEATS) was the first attempt after the vanilla MLP to demonstrate that pure DL architectures using no time-series specific components can outperform well-established statistical techniques [43] and many researchers improved upon it in recent studies following the same idea [33], [44], [45], [46].

CNN-based architectures gained popularity for time-series by applying 1D convolutional kernels and treating them as 1D signals for feature extraction. Their ability to process multivariate time-series and learn local patterns makes them effective for scenarios where multiple input sensors are available and local temporal patterns are important for prediction. The robustness, efficiency, and scalability of convolutional kernels in capturing temporal features in time-series data collected from manufacturing systems for TSC tasks have been reported in a recent study [28]. WaveNet and TCN algorithms are the earliest adaptations of CNN-based architectures for TSF tasks using stacks of convolutional layers, dilated casual convolutions, and residual blocks [47], [48]. More recent algorithms like BiTCN [49], Sample Convolution and Interaction Network (SCINet) [50], TimesNet [51], and Multi-scale Isometric Convolution Network (MICN) [52] tried to overcome TCN limitations to capture both local and global features with different techniques while keeping the main idea behind it.

RNN-based architectures were designed to handle sequential data by keeping an internal state that captures temporal dependencies. The main idea behind recurrent connections is to enable these algorithms to learn complex temporal patterns that might be missed by simpler architectures. Basic RNN architectures are difficult to train and more sophisticated algorithms such as LSTM [53] and GRU [54] are used more frequently by researchers [48]. RNN-based architectures' performance degrades when dealing with a large look-back window or forecasting for a large horizon. This is due to vanishing and exploding gradient memory constraints in their memory cells which makes it difficult to maintain dependencies between distant time stamps. Although many researchers have reverted back to other architectures for long-term forecasting tasks, like SegRNN algorithms, to overcome this issue [55].

Other works have aimed to combine aspects of CNN and RNN architectures to leverage the strengths of both architectures for discovering local dependency patterns among input variables via convolution layers and capturing long-term dependencies via recurrent layers. The Long- and Short-term Time-series Network (LSTNet) algorithm is an example of such an architecture designed for TSF tasks utilizing recurrent-skip connections [56].

Transformer-based architectures [57] are arguably the most successful DL architecture, for sequence modeling in Natural Language Processing (NLP) and Speech Recognition applications [58]. These models leverage self-attention mechanisms for capturing the dependencies in time-series and modeling long-range dependencies. Recently, there has been a great effort to propose new transformer-based solutions for TSF tasks claiming state-of-the-art forecasting performance [45]. Most notable algorithms include Informer [59], Temporal Fusion Transformer (TFT) [60], Autoformer [61], FEDformer [62], PatchTST [63], Crossformer [64], and iTransformer [65]. These algorithms utilize different techniques such as specialized temporal attention mechanisms and position encodings designed specifically for time-series data. However, recent work has shown that these transformer-based architectures may not be as powerful for TSF tasks as they are in other applications, and they can be outperformed by linear models on TSF benchmarks. This may be due to TSF's precise temporal ordering, unlike NLP where meaning can be preserved even if some words are reordered. Self-attention's permutation-invariant nature, even with positional encoding, is a hindrance when modeling time-series data that lacks semantic context [58]. Large Language Models (LLMs) are using transformers as their main element and have recently been used for TSF tasks in LLMs models like Chronos [66] and Time-LLM [67].

## 2.3. Public Manufacturing Datasets for Time-Series Forecasting

One of the biggest challenges in manufacturing is the scarcity of suitable, accessible, and public datasets for AI and ML applications [68]. Not only are these datasets scarce, but they are also scattered across different sources and not easily identifiable, requiring significant effort to locate manufacturing-related data. The standard practice in the CS community is to use or share datasets publicly to ensure reproducibility. However, this is not a common practice in manufacturing, where data sharing might be restricted due to many concerns such as cybersecurity, competitive pressures, and intellectual property protection. This makes their research non-reproducible for other researchers [42], [69], [70], [71], [72]. As a result, researchers in manufacturing must either rely on hard-to-obtain data collection from machines and conduct preprocessing from scratch, or adapt datasets from other domains. While valuable, both approaches are far from ideal as data collection and preprocessing is time-consuming and the datasets from other domains lack characteristics inherent in and defining manufacturing problems.

Currently, there are a limited number of manufacturing datasets that are publicly available to researchers for TSF and TSR tasks. However, these datasets only cover a narrow range of manufacturing applications. The UCR and UEA time-series repository [73], [74], and UCI ML repository, are examples of generic publicly available datasets which include a few manufacturing-related datasets. The Prognostics and Health Management (PHM) Society yearly competition datasets shared on NASA's Prognostics Data Repository [75] is the only domain-specific resource that we could find for this purpose.

Due to a shortage of AI-ready testbeds [76], the reluctance to share data, and the scarcity of suitable public manufacturing datasets, manufacturing is missing out on valuable advantages afforded to the computer science field by strong and plentiful data availability to advance the field as a whole. This includes faster domain-specific solutions and algorithm development, as well as late adoption of AI technologies. Public manufacturing datasets and AI testbeds are key to overcoming critical challenges in AI adoption, including the lack of technical knowledge, data, and experience with AI tools and techniques, by enabling better education [68]. An additional advantage is the ability to conduct benchmarking, where correlatable performance comparisons of algorithms can be performed using specified metrics and public datasets which is unlikely without shared datasets and testbeds. There are multiple studies in other domains while manufacturing continues to lag behind [77], [78], [79], [80]. The field urgently needs comprehensive, accessible, and scalable testbeds including multimodal data with real-time capabilities [81].

In this study, we took a step to bridge this gap by gathering several available datasets from various manufacturing applications. To find a representative list, we first gathered a list of 23 datasets shown in Table 2 from various sources. The result is a repository of manufacturing-related datasets with characteristics that can be fed into ML algorithms to investigate their performance in a smart manufacturing setting for TSF and TSR tasks.

Table 2. List of 23 publicly available datasets suitable for TSF and TSR tasks.

| Dataset Name | Domain | Application | Problem Type | Reference |
|---|---|---|---|---|
| Gas sensor temperature modulation | Chemical | Temperature Prediction | TSF | UCI[2] |
| Gas sensor dynamic gas mixtures | Chemical | Chemical Sensors | TSF | UCI |
| Appliances Energy | Energy | Pressure prediction | TSF | UCI |
| Electricity | Energy | Energy Consumption prediction | TSF | UCI |
| ETTm2 | Energy | Temperature Prediction | TSF | Zhou et al [59] |
| ETTh1 | Energy | Temperature Prediction | TSF | Zhou et al [59] |
| ECL | Energy Consumption | Energy Consumption prediction | TSF | UCI |
| ISO-NY | Chemical Sensors | Load Forecasting | TSF | NYISO[3] |

[2] https://archive.ics.uci.edu/
[3] https://www.nyiso.com/

| | | | | |
|---|---|---|---|---|
| Monroe Water Treatment Plant | Chemical Sensors | Energy Consumption prediction | TSF | Bloomington data portal[4] |
| Seoul Bike Demand | Bike Sharing | Demand Forecasting | TSF | UCI |
| AI4I 2020 | Predictive Maintenance | Process Temperature Prediction | TSF | UCI |
| Steel Industry | Steel Industry | Energy Consumption prediction | TSF | kaggle[5] |
| C-MAPSS | Aerospace | RUL | TSR | NASA[6] |
| IGBT-NASA | Electronics | RUL | TSR | NASA |
| MILLING_NASA | Milling | RUL | TSR | NASA |
| NASA_Accelerated Battery Life Testing | Electronics | RUL | TSR | NASA |
| PHM 2012 | Bearing | RUL | TSR | PHM[7] |
| PHM 2018 | Semiconductor | RUL | TSR | PHM |
| PHM 2015 | HIRF Batteries | Remaining Flying Time Prediction | TSR | NASA |
| PHM 2019 | Manufacturing | Crack Length Estimation | TSR | NASA |
| Twin gas sensor arrays | Chemical | Fault Diagnosis | TSR | UCI |
| Condition monitoring of hydraulic systems | Railway | Condition Monitoring | TSR | UCI |
| Brent Oil Prices | Oil Industry | Price Forecasting | TSF | Kaggle[8] |

# 3. Materials and Methods

This section provides an overview of the datasets used in this study, their relations with smart manufacturing use cases, and dataset-related parameter and experimental evaluation settings that must be considered in TSF tasks. Moreover, it provides the details of algorithm selection, parameter setups, algorithm-related parameters, and experimental evaluation settings. The general framework applied in this work is illustrated in Figure 2. It involves a series of preprocessing steps applied to the raw time-series data, the experimental setup, the TSF algorithms categories, and details of the evaluation steps carried out to find the top-performing TSF algorithms. The details of each step are discussed in subsequent sections. The source code of the experiments can be found in the paper's Github repository[9].

[4] https://bloomington.data.socrata.com/
[5] https://www.kaggle.com/datasets/ayushparwal2026/steel-industry-datasets
[6] https://www.nasa.gov/intelligent-systems-division/discovery-and-systems-health/pcoe/pcoe-data-set-repository/
[7] https://phmsociety.org/
[8] https://www.kaggle.com/datasets/mabusalah/brent-oil-prices

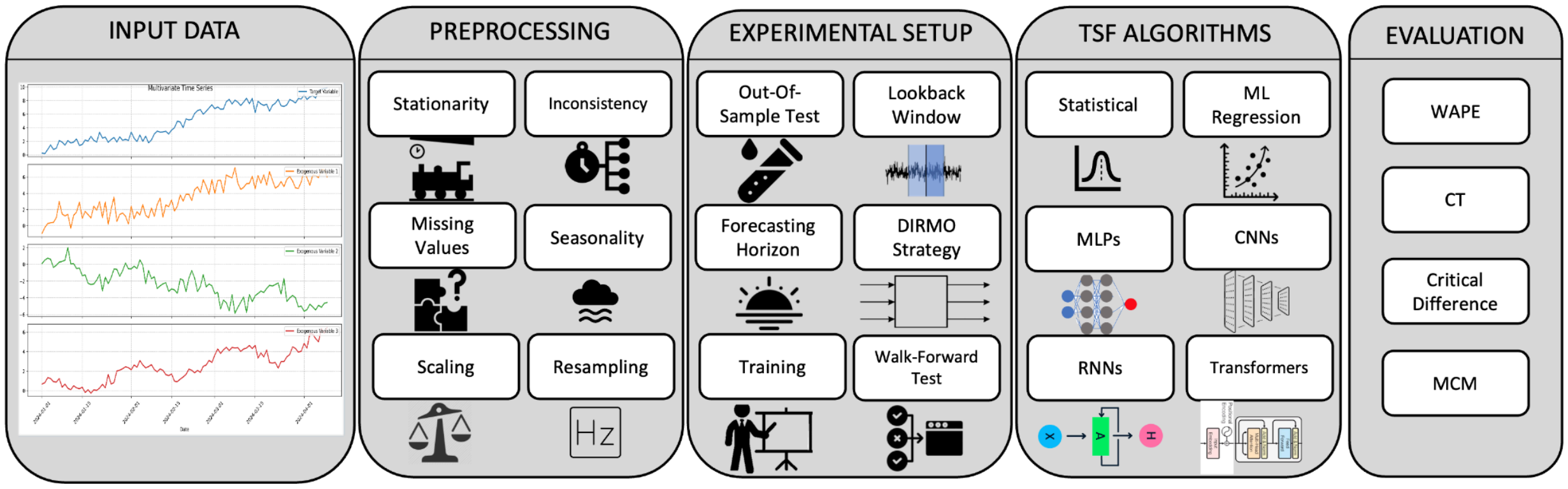

Figure 2. The General TSF Framework of this study

## 3.1. Datasets

Multiple parameters must be considered when selecting datasets for TSF tasks. First, the dataset should be examined from the problem formulation standpoint to determine whether it fits the TSR or TSF definition. TSR is a task that aims to learn an approximation function from the input time-series variables to a target scalar value. For example, RUL prediction is a typical TSR task within the manufacturing domain [20]. In contrast, TSF is a type of regression that is used to fit autoregressive models on the historical values of the target variable which is a time-series itself.

Secondly, it is critical to consider the number of algorithm input variables. With a univariate dataset, only TSF problems can be formulated since TSR inherently requires multiple variables. For multivariate datasets, there are several options: multiple input variables (either static or dynamic) with multiple dynamic outputs, multiple inputs with a single dynamic output variable for TSF, or multiple inputs with a single static output for TSR. In TSF problems with multivariate inputs, the additional variables can serve as covariates or exogenous variables to improve forecasting accuracy [29]. Table 3 displays the final list of 13 datasets suitable for TSF tasks that were used in this study alongside additional useful characterization information. More details of the datasets can be found in Appendix A.

Table 3. List of thirteen publicly available datasets suitable for TSF tasks. The target variable is used as the main forecasting variable. M denotes the number of used variables in the multivariate time-series. Granularity refers to the time difference between two subsequent observations. The scaling and transformation method, and the method to deal with missing values is also mentioned.

| Dataset Name | Target Variable | M | Granularity | Scaling | Missing Values | Transformation |
|---|---|---|---|---|---|---|
| Gas Sensor Temperature Modulation | Temperature | 5 | 30 Seconds | MiniMax | Backward Fill | None |
| Gas Sensor Dynamic Gas Mixtures | MOX sensor value | 7 | 1 Second | MiniMax | Backward Fill | None |

[9] https://github.com/tamoraji/PyTSF-MfG

| | | | | | | |
|---|---|---|---|---|---|---|
| Appliances Energy | Pressure | 6 | 10 Minutes | MiniMax | Backward Fill | None |
| Electricity | Active Power | 5 | 1 Hour | MiniMax | Backward Fill | None |
| ETTm2 | Oil Temperature | 7 | 15 Minutes | MiniMax | Backward Fill | None |
| ETTh1 | Oil Temperature | 7 | 1 Hour | MiniMax | Backward Fill | None |
| ECL | usage_KW | 8 | 1 Hour | MiniMax | Backward Fill | None |
| ISO-NY | Power Load | 1 | 15 Minutes | MiniMax | Backward Fill | None |
| Monroe Water Treatment Plant | total_kwh | 5 | 1 Day | MiniMax | Backward Fill | None |
| Seoul Bike Demand | Demand | 6 | 1 Hour | MiniMax | Backward Fill | None |
| AI4I 2020 | Process Temperature | 4 | 1 Minute | MiniMax | Backward Fill | None |
| Steel Industry | Usage_kWh | 6 | 15 Minutes | MiniMax | Backward Fill | None |
| Brent Oil Prices | Price | 1 | 1 Day | MiniMax | Backward Fill | None |

On average researchers test their models with four to six datasets to conduct experimental evaluations on their predictive capabilities. The sole exception that stood out was the research by Lara-Benitez et al., who performed their evaluation on twelve datasets [80]. This aligns with the recommendation from Demsar, J. for such studies who state that the number of datasets greater than ten is considered sufficient for this type of performance evaluation [82]. Hence, the goal was to choose between ten to twelve manufacturing-related datasets for each experiment scenario in this study. Unfortunately, it was not possible to identify this number of unique, suitable, and core manufacturing datasets due to the scarcity of datasets in the domain. This is a chronic issue in the manufacturing domain and needs to be addressed by the community [28], [81], [83].

However, each scenario addressed the data scarcity limitation by selecting datasets from adjacent domains with transferrable manufacturing applications to reach the suggested number for conclusive evaluations. The experiments were performed on the complete set of datasets to analyze and report the results. For example, although the ISO-NY dataset belongs to the power load consumption of a municipal area, it is translatable to the power consumption of an industrial site with the same variables. Moreover, when the experiment scenarios were being designed, a diverse set of target variables were carefully selected to cover a wide range of problems.

## 3.2. Data Preprocessing

Data preprocessing is the process of cleaning, formatting, transforming, and preparing the dataset for algorithm digestion. This step is one of the most time-consuming activities in many ML projects. Preprocessing is necessary to make forecasting tasks easier for the algorithm by transforming the time-series, which can have a prominent impact on TSF algorithm performance [23]. While traditional TSF techniques require stationarity of the input data mean and variance, the ML community has no general consensus on this topic. Some studies suggest that ML algorithms are technically capable of handling raw data and learning the hidden patterns without preprocessing, while other studies argue that skipping preprocessing steps like stationarity transformation can lead to instability and poor performance [24].

This section outlines the main preprocessing steps in a typical TSF framework, presenting different options for each step along with this study's chosen methods and rationale. Since this study targets researchers and practitioners in the manufacturing domain who may be new to AI and ML, one of our goals is to identify the best-performing TSF algorithms in an "out-of-the-box" fashion with minimal algorithm specific preprocessing hyperparameter tuning.

### 3.2.1. Missing Values and Time Inconsistency

ML algorithm performance tends to degrade when there are missing values or timestamp inconsistencies within the data. Thus, these issues need to be resolved through preprocessing before algorithm deployment. There are different techniques available to fill in missing values such as mean or zero substitution, linear interpolation, replacement with the dataset's minimum or maximum value, forward or backward filling, etc. [26].

In this study, the datasets were reindexed when minor timestamp duplications and inconsistencies were found. Once this was handled, all datasets were checked for missing values and remediated using the backward filling method. The exceptions to this method were for the Gas Sensor Temperature Modulation and Gas Sensor Dynamic Gas Mixtures datasets where downsampling was implemented based on the sampling rates in Table 3 in addition to previous steps. This resampling served two purposes: to manage the dataset sizes for computational efficiency and mitigating potential forecasting accuracy issues when high-frequency sampling captures small variations in the data, particularly for shorter forecasting horizons. In these cases, high sampling granularity may introduce noise rather than meaningful patterns, potentially degrading the model's forecasting performance.

### 3.2.2. Handling Seasonality and non-stationarity

Traditional statistical TSF algorithms assume linearity between a time-series' past values and the forecasted value. This assumed relationship is sensitive to seasonality and non-linear patterns. Seasonality can obscure underlying patterns and unstable variance which may lead to biased forecasts. Stationary time-series have a mean and variance that remains within a threshold over time. Thus, many time-series algorithms may perform better when a time-series is stationary without seasonal components. This is why researchers suggest preprocessing steps before applying the algorithms. Generally, three types of preprocessing can be performed to mitigate the effects of seasonality and nonlinearity: seasonal adjustments, log transformations, and power transformations are popular methods in literature for stabilizing the variance and removing the trend.

Log transformation is the simplest approach for variance stabilization. This method works by compressing high values and expanding low ones. Box-Cox transformation is a more flexible power transformation technique that has been used by many researchers [24]. The motivation for using the Box-Cox transformation is to ensure variance normality which is an assumption in many statistical techniques [1]. However, this technique is unable to handle negative numbers so the Yeo-Johnson transformation was developed to resolve this limitation [26].

Seasonality can be assumed as either additive or multiplicative depending on the assumption that the time-series is either the addition or product of its components. Researchers have developed different methods such as STL decomposition [84] to separate time-series into trend, seasonal, and remainder components to remove seasonality for further analysis.

To achieve stationarity of the mean and remove trends, differencing is the most common approach. It works by subtracting consecutive observations to eliminate the trend component. First-order differencing can handle linear trends, while higher-order differencing can be used for quadratic or polynomial trends.

In contrast to statistical TSF techniques, there is no consensus among the ML and DL literature on how to handle seasonality and nonlinearity. Some studies claim that ML methods are capable of learning any type of nonlinearity in the data and therefore can be applied to the original raw time-series data while others claim that without sufficient preprocessing steps, such methods may become unstable and yield poor results [23], [85]. In this study, since the focus is primarily on ML and DL algorithms, it was decided that the original time-series data without any transformations or seasonality adjustments will be used. This facilitated a consistent evaluation of each algorithm's inherent capabilities without preprocessing biases that could be unknowingly introduced and simplifies the TSF pipeline. Moreover, this is particularly valuable for understanding the real-world applicability of each algorithm when preprocessing might not be feasible.

### 3.2.3. Scaling

Different sensors collecting time-series data have varying scales due to inherent difference of the recorded phenomena which may have an adverse effect on DL algorithm performance. These algorithms are particularly sensitive to the scale and distribution of input time-series. For instance, if one feature ranges from 0 to 1,000 while another is from 0 to 1, the first feature might dominate the learning process, leading to poor model performance and slower convergence. Additionally, optimization algorithms like gradient descent require features to be on similar scales to work effectively. The two most common techniques used in literature are min-max scaling–also known as normalization–which transforms the data to a fixed range of 0 to 1, and standardization which centers the data around a mean of zero with unit variance.

In this study, for all ANN and ML algorithms, a normalization method was used to scale the time-series of training data between 0 and 1 to avoid these issues. We did not apply any scaling prior to using the AutoARIMA and Naive algorithm because they use different a forecasting technique than ML algorithms that do not rely on gradient descent optimization. In fact, scaling the input data might potentially degrade their forecasting performance of these algorithms.

## 3.4. Algorithms

After reviewing the literature, an initial list of 117 algorithms was compiled. While assembling the list, several features were extracted from each algorithm that later helped with designing the experimental scenarios and down-selecting a list of representative algorithms for each case. First, the algorithms were sorted by their proposed year to easily determine which algorithms were the most current when multiple options were present with similar features. The general assumption is that newer algorithms typically outperform their predecessors in the same category by addressing previous limitations. For instance, the Informer algorithm [59] was introduced in 2021 as an attempt to alleviate the limitations of the Transformer algorithm [57] for long sequence TSF and reported comparably superior performance.

Then, algorithms were categorized into univariate or multivariate based on their input data handling capabilities. Multivariate algorithms can process univariate data, while univariate algorithms cannot handle multivariate inputs which limits their applications. An evaluation of forecasting output type was also utilized to separate algorithms into point and probabilistic forecasting categories. Point forecasting algorithms generate specific predicted values, while probabilistic approaches provide specific predicted values with prediction intervals. Probabilistic algorithms can be simplified to point forecasting if desired, whereas point forecasting algorithms cannot generate probability distributions without modifications.

The algorithms' modeling scope was also analyzed to determine whether they fit into local or global approaches. Traditional statistical algorithms such as ETS [36], ARIMA, and Theta [39] are called local models and consider each time-series instance as an independent regression task to fit a function. On the other hand, ANN and DL algorithms are called global models that fit a function to all the time-series instances in the dataset. Local algorithms have been found to perform better with smaller datasets and short-horizon forecasting tasks, while global algorithms are deemed more effective for larger datasets and long forecasting horizons. This characteristic influences algorithm selection based on the dataset size and forecasting requirements [86].

Next, the algorithms' main architecture was documented, identifying how they process temporal patterns. Moreover, each algorithm's primary design purpose was extracted to understand specific target problems or limitations. This combined information enables grouping algorithms with similar underlying mechanisms while highlighting key differences. As a practical consideration, code availability was documented, noting algorithms with open-source implementations and Python libraries, as this directly impacts implementation feasibility across different applications.

Finally, algorithms without publicly available codes were excluded and a pool of 53 algorithms was created based on the scope of the study and were categorized into eight groups based on common characteristics. The ML and statistical algorithms were pruned to retain only the representative algorithms from each family. For example, from the Exponential Smoothing family (which includes SES, Holt's ES, Holt-Winters ES, Damped ES, and ETS), only ETS was retained because it was the latest instance in that family and intuitively was the most advanced algorithm among them. The eight groups include Statistical, ML regression, MLP-based, CNN-based, RNN-based, Transformers, and LLMs. This categorization ensures representative coverage across different methodological approaches while considering practical implications. Table 4 presents this study's algorithm pool.

Table 4. The final list of 53 algorithms pool. The Library shows the Python library that includes the open-source codes for the algorithms and the Reference highlights the paper that proposed the algorithm.

| Algorithm | Category | Library | Year | Algorithm | Category | Library | Year |
|---|---|---|---|---|---|---|---|
| Transformer [57] | Transformers | Darts [87] | 2017 | LSTM-ATT [88] | RNNs | Github Repo [10] | 2021 |
| Informer [59] | Transformers | NeuralForecast[11] | 2021 | SegRNN [55] | RNNs | TSLiB [89] | 2023 |
| TFT [60] | Transformers | NeuralForecast | 2021 | CNN [90] | CNNs | TimeSeriesForecasting[12] | 1998 |
| Autoformer [61] | Transformers | NeuralForecast | 2022 | WaveNet [47] | CNNs | GluonTS [91] | 2016 |
| FEDformer [62] | Transformers | NeuralForecast | 2022 | TCN [48] | CNNs | Darts | 2018 |
| ETSformer [92] | Transformers | TSLiB | 2022 | DeepGLO [35] | CNNs | DeepGLO[13] | 2019 |
| Pyraformer [93] | Transformers | TSLiB | 2022 | SCINet [50] | CNNs | TSLiB | 2022 |
| PatchTST [63] | Transformers | NeuralForecast | 2023 | BiTCN [49] | CNNs | NeuralForecast | 2023 |
| Crossformer [64] | Transformers | TSLiB | 2023 | TIMESNET [51] | CNNs | NeuralForecast | 2023 |
| iTransformer [65] | Transformers | NeuralForecast | 2024 | MICN [52] | CNNs | TSLiB | 2023 |
| Time-LLM [67] | LLMs | NeuralForecast | 2024 | LSTNet [56] | CNN-RNNs | GluonTS | 2018 |
| Chronos [66] | LLMs | Chronos | 2024 | TPA-LSTM [94] | CNN-RNNs | Github Repo[14] | 2019 |
| MLP | MLPs | NeuralForecast | 1958 | Naive | Statistical | n/a | n/a |
| N-BEATS [43] | MLPs | NeuralForecast | 2020 | ETS [95] | Statistical | StatsForecast[15] | 2002 |
| N-BEATSX [96] | MLPs | NeuralForecast | 2021 | AutoARIMA [97] | Statistical | StatsForecast | 2008 |
| N-HiTS [44] | MLPs | NeuralForecast | 2022 | TBATS [40] | Statistical | StatsForecast | 2011 |
| LightTS [33] | MLPs | TSLiB | 2022 | AutoTheta [98] | Statistical | Sktime | 2018 |

[10] https://github.com/ZhenghuaNTU/RUL-prediction-using-attention-based-deep-learning-approach/tree/master
[11] https://github.com/Nixtla/neuralforecast
[12] https://github.com/pedrolarben/TimeSeriesForecasting-DeepLearning
[13] https://github.com/rajatsen91/deepglo
[14] https://github.com/gantheory/TPA-LSTM
[15] https://github.com/Nixtla/statsforecast

| | | | | | | | |
|---|---|---|---|---|---|---|---|
| DLinear [58] | MLPs | Darts | 2022 | Prophet [41] | Statistical | Darts | 2018 |
| NLinear [58] | MLPs | Darts | 2022 | AR-Net [99] | Statistical | Triebe et al., | 2019 |
| TSMixerX [46] | MLPs | NeuralForecast | 2023 | Koopa [100] | Other | TSLiB | 2023 |
| TiDE [45] | MLPs | NeuralForecast | 2024 | KNNR | ML regression | Sktime | 1951 |
| Block RNN | RNNs | Darts | 1986 | SVR | ML regression | Sktime | 1998 |
| ERNN | RNNs | NeuralForecast | 1990 | RFRegressor | ML regression | Darts | 2001 |
| ESN [101] | RNNs | TimeSeriesForecasting | 2001 | XGBoost [102] | ML regression | Darts | 2016 |
| Block LSTM [53] | RNNs | Darts | 2014 | LightGBM | ML regression | Darts | 2017 |
| Block GRU [103] | RNNs | Darts | 2014 | RocketRegressor [104] | ML regression | Sktime | 2019 |
| Dilated RNN [105] | RNNs | NeuralForecast | 2017 | | | | |

This table can be used as a valuable reference for a representative list of the state-of-the-art TSF algorithms from different categories with available open-source implementation. It can be used as a baseline by both practitioners and researchers for various applications and use cases within smart manufacturing systems.

## 3.5. Experimental Setup

This section explicates different aspects of the proposed experimental scenarios. Each experiment was run in a Python virtual environment with Python 3.12 installed and major DL and ML python packages such as Neuralforecast 1.7.5, scikit-learn 1.5.2, torch 2.4.1, xgboost 2.1.1, darts 0.30.0, statsforecast 1.7.7.1 were used in this study. All experiments have been done on an AMD Threadripper Pro 5975WX CPU with 32 cores (64 threads), coupled with 2x RTX A5000 (24 GB) GPUs, and 512GB of memory.

### 3.5.1. Training, Test and Evaluation scheme

The consensus among TSF researchers is that the performance of TSF algorithms should be assessed using *out-of-sample* tests on the hold-out set rather than in-sample tests and goodness of fit to the training set. For a given TSF algorithm, in-sample errors are likely to be lower than testing errors due to issues like overfitting and low generalization so algorithms selected by the best in-sample fit may not best predict post-sample data [106]. When splitting the dataset, it is very important to ensure that it does not leak future data to the model, so techniques like k-fold cross-validation that mix future and past observations, cannot be used [107]. However, it is important to split the dataset into a *training set* and a *test set* in experimental evaluation studies. The training set is considered as the historical data and used to train the algorithm and to find the approximation function modeling the dataset. The final time in the training set (*t*) is the point from which the out-of-sample forecasts are generated and is called the *forecasting origin*. The test set is used for performance evaluation and comparison between different algorithms. The number of time periods between the origin and the time being forecast is the *forecasting horizon*.

In an out-of-sample test, we can use either a *fixed-origin evaluation* or a *rolling-origin evaluation*. Standing at origin (t) in a fixed-origin setup, the forecast is generated for the next forecasting horizon (H) and the forecasting error is calculated. The main issue with the fixed-origin evaluation is that if the test set is equal to the forecasting horizon and only one forecast is generated, that forecast is susceptible to biases due to occurrences unique to that origin. In a *rolling-origin evaluation*, the origin is successively updated and new forecasts are generated from each new origin [106].

In this study, we first split the dataset into two separate lists of training and test sets with an 80:20 ratio. Then a walk-forward performance evaluation is used on the test set to assess efficacy. Walk-forward validation is a variation of the rolling-origin evaluation commonly used to evaluate the performance of TSF algorithms. The basic idea of walk-forward validation is to train the algorithm using the historical data up to the forecasting origin, make predictions for the immediate future equal to the forecasting horizon, and then update the history with the true outcome for that horizon. This process is repeated sequentially, iteratively advancing the evaluation window with steps equal to the forecasting horizon until forecasts are generated for the complete test set [75]. This is equivalent to the blocked cross-validation suggested by Bergmeir and Benítez [76].

The next step is to define each scenario's forecasting horizon and lookback window length. This study aims to evaluate both short- and long-term scenarios by examining short-term forecasting horizons of 3, 6, and 12 time steps and long-term forecasting horizons of 96, 288, and 672 time steps. These time periods are common when evaluating forecasting horizons across research domains and various applications in literature. There is no definite rule on how to choose the best lookback window for TSF tasks. However, the general rule is that large lookback windows provide improved results due to having more information and should be slightly bigger than the forecasting horizon while including the seasonality period. A window of 1.25 to 3 times of the forecasting horizon seems to be the accepted heuristic for this task [26]. In this study, we used a fixed lookback window for each problem and we set 50 as the lookback window for our short-term forecasting scenarios and 1,000 as the lookback window for our long-term forecasting scenarios. These numbers were chosen empirically making sure a balance between providing enough historical time steps and maintaining computational efficiency.

### 3.5.2. Multiple output strategies

Finally, when forecasting the future for multiple timestamps there are several strategies that can be deployed. Each of these strategies has unique advantages and disadvantages with the choice between them depending on the problem formulation, use case, and algorithm capabilities. There are five types of multistep ahead forecasting strategies commonly used throughout the literature. In this study, DIRMO strategy has been used.

**Recursive strategy**–also known as iterative or autoregressive–is the oldest and most commonly used forecasting strategy. It works by training a single model **F** to make one-step-ahead predictions, then uses each prediction as input for forecasting the next step. This process continues iteratively until reaching the desired forecasting length **L**. While this method is straightforward to implement, its main limitation is the accumulation of prediction errors that occurs over longer horizons, as each new forecast builds upon previous predictions [109]. Figure 3 shows a visualization of how the recursive strategy works.

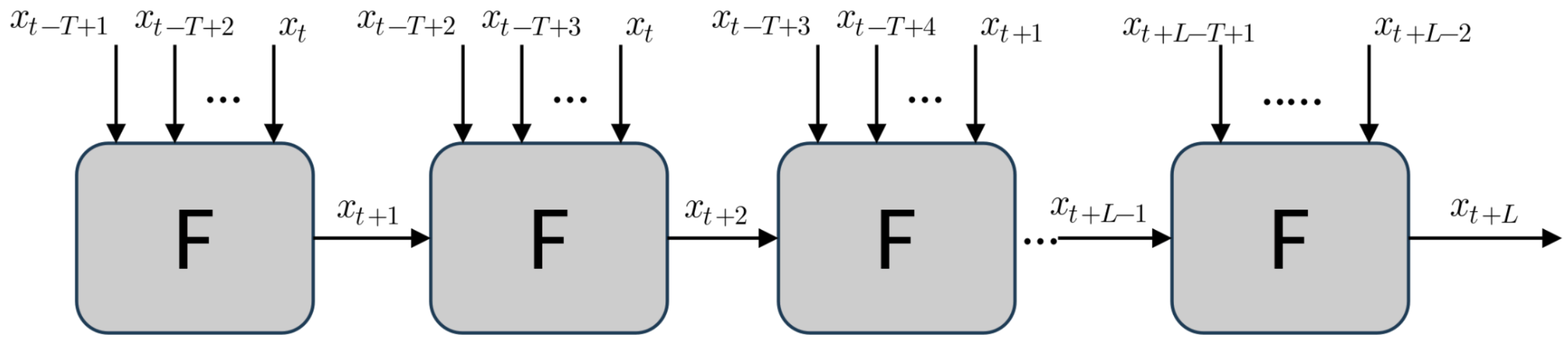

Figure 3. Recursive or autoregressive Strategy

**The Direct strategy**–also known as independent–uses a different approach by training separate models for each time horizon independently instead of using a single iterative model. In other terms, **L** models are learned from the time series for the forecasting length **L**. While this method eliminates the problem of accumulating errors of the recursive strategy, the independent nature of this strategy means they cannot capture complex relationships between time horizons and it may affect the forecasting accuracy [110]. Figure 4 shows a visualization of how the direct strategy works.

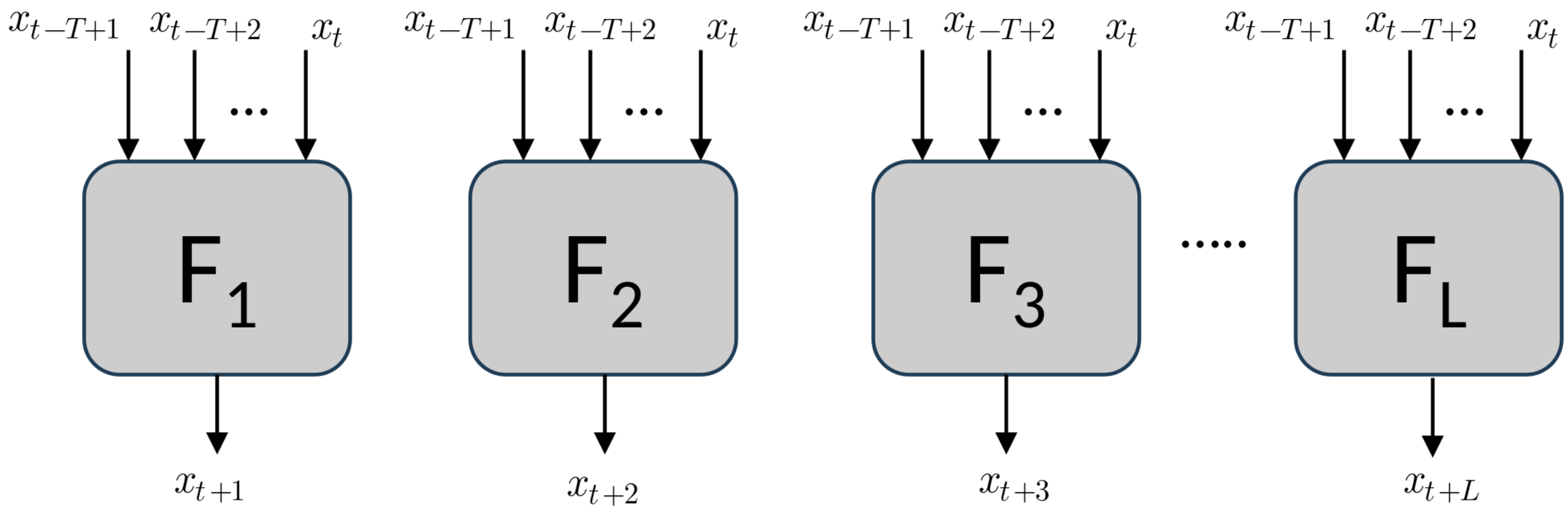

Figure 4. Direct or independent Strategy

**The DirRec strategy** combines the principles of the Direct and the Recursive strategies. DirRec computes the forecasts with different models for every horizon, same as the Direct strategy, and feedback the predictions to the model like the Recursive strategy. A limited number of studies have been done regarding this strategy due to its complicated implementation [111]. A limitation of DiRec strategy is that the lookback window expands with each prediction step, which can be a limiting factor as it may lead to scalability challenges for certain applications and algorithms, particularly those with limited computational resources. This expanding window Figure 5 shows a visualization of how the DirRec strategy works.

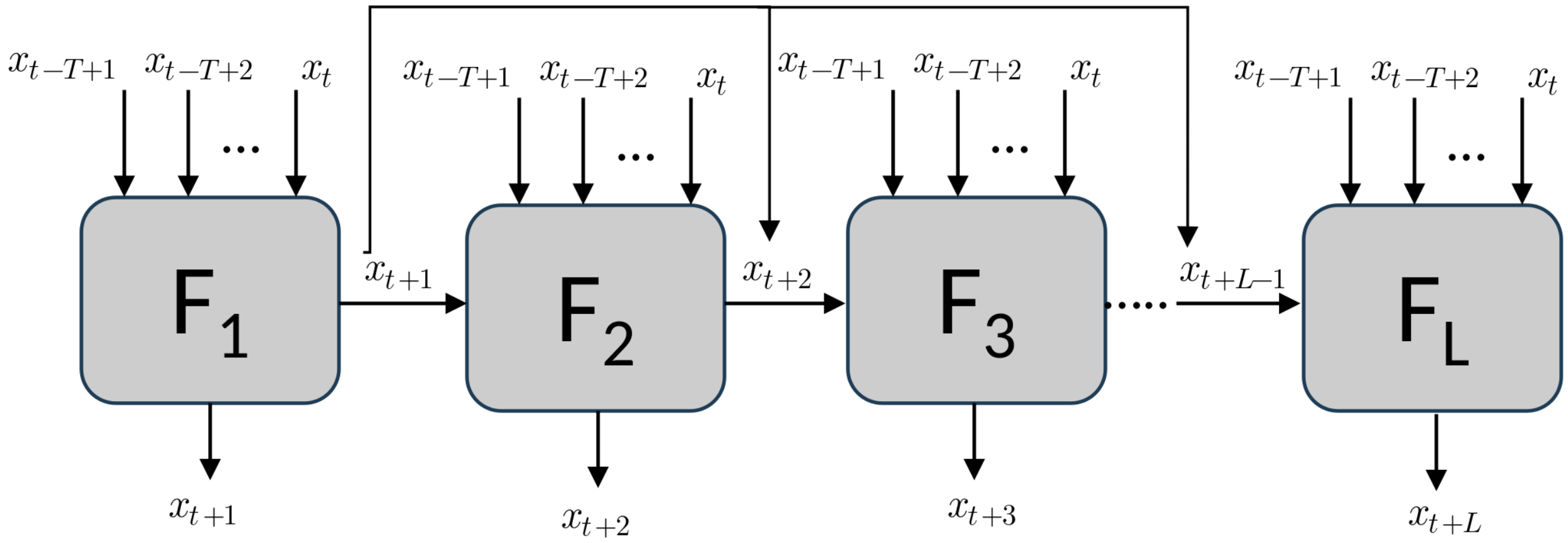

Figure 5. DiRec Strategy

**The Multi-Input Multi-Output (MIMO) strategy** learns an array of multiple outputs from the input time series. Unlike the previous strategies, the MIMO strategy returns a vector of values in a single step. The advantage of this strategy comes from producing the forecasts for the whole output window at once, thus incorporating the inter-dependencies between each time step and preventing error accumulation over predictions [26], [80]. MIMO strategy is widely used in the literature using DL algorithms. Figure 6 shows a visualization of how the MIMO strategy works.

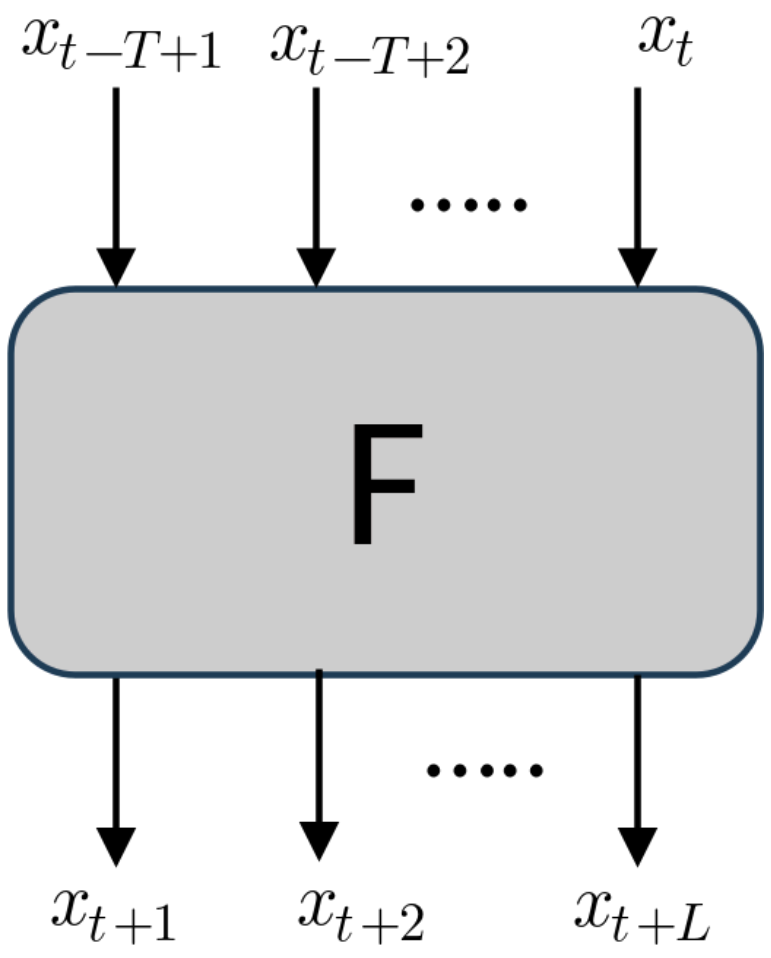

Figure 6. MIMO Strategy

**The DirMO strategy** forecasts the forecasting length in blocks, where each block is forecasted with MIMO method. Thus, the forecasting task of the next **L** steps is divided into **N** MIMO tasks of length **H** for each step [111]. Figure 7 shows a visualization of an adaptation DirMO strategy used in this work where the test set is divided into **N** blocks where the length of each block is equal to the designed forecasting horizon. Each block is then predicted using a fixed number of inputs with the MIMO method. After predicting each block, the training set gets updated with **H** observations equal to the length of the forecasting horizon and then the next block is predicted.

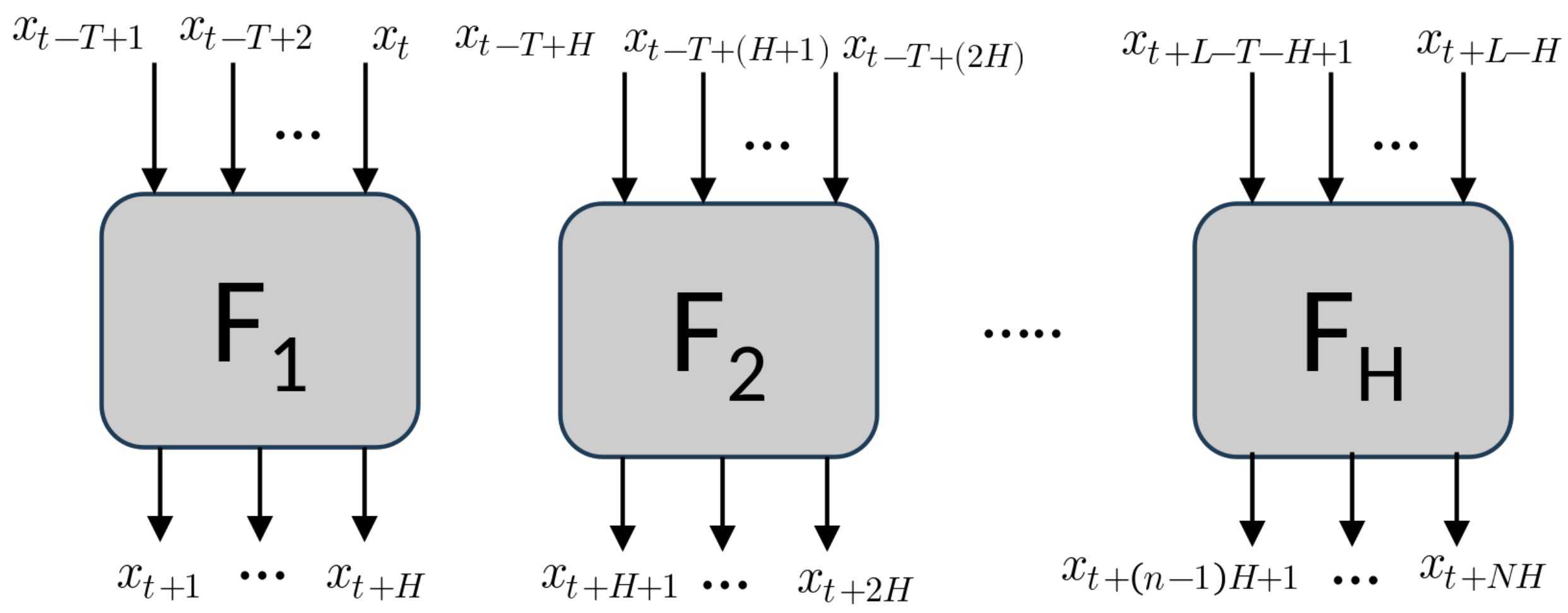

Figure 7. DirIMO strategy

## 3.6. Evaluation metrics

Once the forecasts are generated, that value or series must be evaluated to assess the validity of the model's output. There are four groups of statistical evaluation metrics which provide quantitative insights about the quality of a forecasting model: error-based metrics, percentage-based metrics, relative performance metrics, and goodness-of-fit metrics.

*Error-based error metrics* calculate prediction error as compared to the ground truth to evaluate the deviation from the expected values. This method is one of the most popular and straightforward evaluation methods. Examples of these metrics are: Mean Square Error (MSE), Root Mean Square Error (RMSE), and Mean Absolute Error (MAE). One study by Dudukcu et al., calculated MSE to evaluate the performance of LSTM and WaveNet to predict jet engine RUL [9]. Others have used both MAE and MSE to provide deeper error-based performance assessments [62].

*Percentage error metrics* provide a similar comparison to error-based metrics but expressed as a percentage of forecast error compared to the ground truth. While these methods are also popular they provide more intuitive and interpretable results for evaluating the efficacy of forecasting models. Mean Absolute Percentage Error (MAPE), symmetric Mean Absolute Percentage Error (sMAPE), and Weighted Absolute Percentage Error (WAPE) are examples of the metrics from this category. In their recommended forecasting approach, Taylor et al. suggest using MAPE due to its simple interpretability [41]. Along with several other studies that have utilized these metrics, the percentage error metrics are widely used across several domains such as computer science [23], energy [19], and economics [112].

*Goodness-of-fit error metrics* are extracted from statistical summaries to assess how well forecasted values fit the ground truth. These are not simple comparisons, rather detailed statistical regression analyses to determine correlations between forecasting models and the ground truth using metrics such as R^2 and Adjusted R^2. These values are usually accompanied by other error assessment metrics studies like Jeong et al., using R^2 with Normalized Root Mean Square Error (NRMSE), MSE, RMSE, and MAE to evaluate deep learning models for organic waste biogas emission prediction [113].

*Relative performance error metrics* assess the performance of a forecasting model when compared to a baseline model such as a Naive forecaster. These metrics differ from the previous categories in that they are comparing forecasted models to a baseline model rather than the ground truth. While these metrics are not as commonly used they are helpful in circumstances where assessing the validity of a forecasting model requires a different perspective. Some examples of these metrics are Relative Absolute Error (RAE), Relative Squared Error (RSE), and Mean Absolute Scaled Error (MASE). When applying several forecasting algorithms to datasets from various domains, Shih et al. used both RSE and RAE with the empirical correlation coefficient (CORR) to evaluate their models [94].

Many studies take a hybrid approach to evaluating their algorithms when it comes to the four categories of evaluation metrics by including several methods. Pabcleylla et al., utilized this hybrid approach by calculating the MAE, RMSE, sMAPE, and MASE to measure forecasting error while deploying LSTM models across several datasets for long and short term supply chain management predictions [114]. Through the deployment of several evaluation criterias, a holistic perspective can be gained on the performance of a forecasting model in a number of scenarios.

For this study, WAPE was used as the main metric of interest when evaluating algorithm performance with RMSE also reported in the Appendix B. WAPE leverages the rescaling of forecast error to make it easier to compare results that feature time series with differing scales [27]. This calculation is made by dividing the MAE by the mean such as in equation 3 where A denotes Actual values and P denotes Predicted values.

$$WAPE = \frac{MAE(A,P)}{Mean(A)} = \frac{Mean(|A-P|)}{Mean(A)} \quad (3)$$

WAPE is widely used as an accuracy measurement for forecasting because of its stability, scale independence and simple interpretability. However, it can be skewed if the actual values have a wide range and underemphasize errors on smaller values in comparison to larger ones [27]. Utilizing this metric will provide this paper with a robust assessment of forecasting performance.

Additional evaluation metrics may be required in some situations to provide more robust decision making information. This might be important when two algorithms have similar performance evaluations without significant differences. The *Computational Time (CT)* is directly related to the algorithms' computational complexity and can be used in this situation and [25]. While error metrics reflect an algorithm's ability to correctly forecast the future, CT can offer a different dimension of evaluation. In real-world applications, where time-sensitive decisions are required or computational resources are limited, the computational speed and efficiency of a TSF algorithm can be important factors [28].

There are many metrics designed to measure the computational complexity of the algorithms such as Floating Point Operations Per Second (FLOPS), Big O notation, Model Fitting (MF), memory bandwidth, etc. Although all these metrics are valuable, given the experimental nature of this study, CT was chosen as the additional metric due being more intuitive, straightforward, and interpretable for real-world applications.

## 3.7. Statistical Analysis

Two post-hoc statistical analyses are used in this study. They allow us to compare the performance of the algorithms across different scenarios. For the first analysis, the recommendation of Demsar was followed to generate a *Critical Difference (CD)* diagram to visualize the various algorithm's performance compared to each other [82]. To do this, a non-parametric *Friedman test* was carried out to test the null hypothesis that all algorithms perform the same and the observed differences are by chance. Then, the *Wilcoxon signed-rank* test with Holm correction ($\alpha = 0.05$) was done to measure the significance of differences between different algorithms.

For the second analysis, a recent tool named the *Multiple Comparison Matrix (MCM)* was used which emphasizes pairwise comparisons over global rankings. MCM is robust and the key advantage of MCM is that pairwise comparisons remain stable and cannot be manipulated by adding or removing other algorithms from the comparison set [115]. The MCM presents pairwise comparisons between algorithms in a matrix format, where each cell contains three key statistics comparing the row algorithm to the column algorithm. The first statistic is the mean difference in performance between the two algorithms across all tasks. The difference is visually represented through a color gradient where red indicates the column algorithm performs better and blue indicates the row algorithm performs better. The second statistic is a Win-Tie-Loss count showing how many times the column algorithm performed better, equal to, or worse than the row algorithm across all tasks. The third statistic is the p-value from a Wilcoxon signed-rank test comparing the two algorithms, with bold text indicating statistical significance. The algorithms in both rows and columns are ordered by their average performance measure across all tasks, ensuring that the relative ordering of any two algorithms remains stable regardless of which other algorithms are included in the comparison. The MCM was generated for the top-performing algorithms identified in the CD diagram for this study.

## 3.8. Experiment Scenarios

Four experiment scenarios were developed for two problems on the dataset pool. *Problem one* is defined as Univariate TSF and *problem two* is Multivariate TSF. Since running all 53 algorithms for each scenario was infeasible due to time and computational resource constraints, a list of eighteen algorithms representative of all categories was selected to run for four scenarios. For univariate experiments, twelve datasets were selected and only the target variable from Table 3 was kept in the preprocessing step while the rest of the variables were dropped. For multivariate experiments, eleven datasets were selected. For each dataset, alongside the target variable, the remaining available variables (equal to M-1 in Table 3) were incorporated into the model as exogenous variables. The total number of variables for each dataset can be found in Table 3. Each scenario includes two sets of algorithms: i) benchmark algorithms of traditional TSF methods to establish baselines for performance comparison and ii) more sophisticated ML and DL algorithms which are the focus of this study. Table 5 illustrates the details of the benchmark algorithms and their properties for this experimental evaluation.

Table 5. Seven Benchmark algorithms used for all four experiment scenarios.

| Algorithm Name | Category | Scenario | Parameters |
|---|---|---|---|
| Naive | Benchmark | 1,2,3,4 | n/a |
| MLP | MLPs | 1,2,3,4 | num_layers=2, hidden_size=1024, batch_size =32 |
| LSTM | RNNs | 1,2 | hidden_dim=32, n_rnn_layers=2, batch_size = 32 |
| XGBoost | ML regression | 1,2,3,4 | Default parameters |
| TCN | CNNs | 1,2,3,4 | num_layers=2, num_filters=64, dilation_base=2, kernel_size=6, batch_size = 32 |
| AutoARIMA | Statistical | 1,2 | max_p=5, max_q=5, max_order=5, max_d=2, stationary=False, seasonal=True, ic='aicc' |
| Block GRU | RNNs | 3,4 | hidden_dim=32, n_rnn_layers=2, batch_size = 32 |

The second set of more advanced TSF algorithms were chosen based on their capabilities and design purposes. We tried to select the most recent algorithms among all categories to have problem-specific representative lists. With the recent interest in transformers their representation has been increased for the list of algorithms for this study. *Scenario one* is defined as *Short-term Univariate TSF* and includes running twelve algorithms on twelve datasets with three forecasting horizons of 3, 6, and 12 time steps. *Scenario two* is defined as *Long-term Univariate TSF* and includes running the same twelve algorithms as scenario one on the same datasets, with three long-term forecasting horizons of 96, 288, and 672. Table 6 shows the details of the TSF algorithms used for scenario one and two in addition to the benchmark algorithms, and their properties for this experimental evaluation.

Table 6. Six TSF algorithms used for scenarios one and two.

| Algorithm Name | Category | Scenario | Parameters |
|---|---|---|---|
| Block GRU | RNNs | 1,2 | Hidden_dim=32, n_rnn_layers=2, batch_size=32 |
| TimesNet | CNNs | 1,2 | Hidden_size=64, conv_hidden_size=64, top_k=5, num_kernels =6, encoder_layers=2, batch_size=32 |
| N-BEATS | MLPs | 1,2 | N_blocks=[1, 1, 1], mlp_units=[[512, 512], [512, 512], [512, 512]], activation=ReLU, batch_size=32 |
| Informer | Transformers | 1,2 | Hidden_size=64, factor=3, n_head=4, conv_hidden_size=32, activation=gelu, encoders=2, decoders=1, batch_size=16 |
| Itransformer | Transformers | 1,2 | n_series=1, hidden_size=512, n_heads=8, e_layers=2, d_layers=1, d_ff=2048, factor=1, batch_size=32 |
| PatchTST | Transformers | 1,2 | Encoder_layers=3, n_heads=16, hidden_size=128, linear_hidden_size=256, patch_len=16, stride=8, batch_size=32 |

*Scenario Three* is defined as *Short-term Multivariate TSF* and includes running eleven algorithms on eleven datasets with three forecasting horizons of 3, 6, and 12 time steps. *Scenario Four* is defined as *Long-term Multivariate TSF* and includes running the same eleven algorithms as scenario one on the same datasets, with three long-term forecasting horizons of 96, 288, and 672. Table 7 displays the details of the TSF algorithms used for scenario three and four in addition to the benchmark algorithms, and their properties for this experimental evaluation. These algorithms are the state-of-the-art in their respective category based on the literature and are expected to outperform the benchmark algorithms.

Table 7. Six TSF algorithms used for scenarios three and four.

| Algorithm Name | Category | Scenario | Parameters |
|---|---|---|---|
| DLinear | MLPs | 3,4 | kernel_size=25, batch_size=32 |
| BiTCN | CNNs | 3,4 | hidden_size=16, dropout=0.5, batch_size=32 |
| TSMixerX | Transformers | 1,2 | n_series = 1, n_block=2, ff_dim=64, revin=True, batch_size = 32 |
| TiDE | MLPs | 1,2 | hidden_size=512, decoder_dim=32, temporal_decoder_dim=128, num_encoders=1, num_decoders=1, temporal_width=4, batch_size = 32 |
| N-HITS | MLPs | 1,2 | n_blocks=[1, 1, 1], mlp_units=[[512, 512], [512, 512], [512, 512]], n_pool_kernel_size=[2, 2, 1], n_freq_downsample=[4, 2, 1], activation=ReLU, batch_size = 32 |
| TFT | Transformers | 3,4 | hidden_size=128, n_head:int=4, grn_activation =ELU, batch_size = 32 |

# 4. Results and Discussion

This section provides the results and analysis of the implemented experimental evaluation for different scenarios and offers additional insights. The evaluation's objective was to assess TSF algorithm performance based on the original authors' (or default parameters in the Python package) recommended configurations without any optimization to see how well they can generalize to different problems. Each algorithm's hyperparameters can be tailored to function more accurately on challenging datasets but for this study the default hyperparameters were used to avoid introducing bias and make the evaluation as aligned with practitioners requirements as possible. All algorithms were trained for 100 epochs. In total, 1,590 experiments were run with a total runtime of 1,192 hours. Next, the summarized results for four experimental scenarios are presented. All post-hoc statistical analyses are based on the WAPE metric. The results are analyzed from different perspectives with a summary presented at the end. The full details of these results can be found in Appendix C.

First, each dataset was examined as an individual problem to investigate which algorithms produce the best results for different situations. The results are summarized in Table 8. For each dataset, the WAPE error was averaged over all forecasting horizons. For univariate forecasting, PatchTST and N-BEATS are identified as the most robust algorithms with six and four wins respectively. For multivariate forecasting, more diverse results are observed, and no specific algorithm dominates. However, XGBoost shows the highest number of wins, with N-HITS as the runner-up, achieving four and three wins, respectively.

Table 8. Best WAPE obtained for each of thirteen datasets for different problems.

| Dataset Name | Target Variable | Univariate forecasting | | Multivariate Forecasting | |
|---|---|---|---|---|---|
| | | Algorithm | WAPE % | Algorithm | WAPE % |
| Gas Sensor Temperature Modulation | Temperature | N-NEATS | 0.34 | Naive | 0.36 |
| Gas Sensor Dynamic Gas Mixtures | MOX sensor value | PatchTST | 15.44 | BiTCN | 5.00 |
| Appliances Energy | Pressure | PatchTST | 0.23 | N-HITS | 0.23 |
| Electricity | Active Power | N/A | N/A | BiTCN | 44.77 |
| ETTm2 | Oil Temperature | PatchTST | 8.28 | N-HITS | 9.39 |
| ETTh1 | Oil Temperature | N-NEATS | 20.57 | N-HITS | 20.65 |
| ECL | usage_KW | PatchTST | 5.15 | XGBoost | 5.83 |
| ISO-NY | Power Load | PatchTST | 5.27 | N/A | N/A |
| Monroe Water Treatment Plant | total_kwh | PatchTST | 7.13 | XGBoost | 7.27 |
| Seoul Bike Demand | Demand | N-NEATS | 40.77 | XGBoost | 41.22 |
| AI4I 2020 | Process Temperature | N-NEATS | 0.15 | Naive | 0.14 |
| Steel Industry | Usage_kWh | XGBoost | 44.09 | XGBoost | 42.25 |
| Brent Oil Prices | Price | Naive | 9.29 | N/A | N/A |

The results in Table 8 highlight some interesting observations, particularly the Naive forecasting algorithm's performance as the best model in certain problems. Naive forecasting predicts the last observed value in the lookback window for the next horizon and is commonly used as a baseline for comparison in forecasting tasks. Although the Naive forecaster might perform well for short horizons, due to the autocorrelation between successive values, this outcome suggests that the underlying patterns in the data were not distinguishable by more complex algorithms. This result is potentially due to high noise-to-signal ratio or insufficient training, and under or overfitting. Furthermore, it may indicate that the problem at hand could benefit from additional preprocessing steps, such as decomposing trend or seasonality, to enhance the signal available for forecasting. This showcases the importance of preprocessing for TSF tasks for manufacturing-related datasets.

Another interesting result is that, for some problems, the WAPE did not improve or increase in certain cases when transitioning from univariate to multivariate problems. The initial expectation was that the error should decrease when more data is available to the model for learning. However, this was not always the case based on these results. This could be because the introduced covariates lack predictive power for the future values of the target variable, reflecting mere correlation rather than causation. Consequently, instead of aiding forecasting, the additional variables increase the dimensionality of the feature space, making it more challenging for the algorithms to learn meaningful patterns. This phenomenon aligns with Clive Granger's 1969 work on economic time-series, which introduced the concept of Granger causality. According to Granger causality, if a time-series covariate x1 causes a target time-series x2, the past values of x1 should contain information that improves the prediction of x2 beyond what is contained in the past values of x2 alone. These findings highlight the importance of feature selection in TSF tasks and challenge the assumption that "the more, the merrier" applies universally, especially in manufacturing-related problems. Additionally, this underscores the potential risks of overfitting and multicollinearity when expanding the feature space without ensuring the relevance of covariates.

## 4.1. Results for different scenarios

Next, different scenarios are analyzed. These results are calculated by averaging different forecasting horizons for brevity. The detailed results of each forecasting horizon can be found in Appendix C. Table 9 summarizes the scenario one results in which consists of twelve univariate TSF algorithms on twelve univariate datasets for all short-term horizons. This represents a summary of 12*12*3=432 experiments. The *AVG ERR* row shows the average of errors over all 432 experiments. The *WIN* row denotes the number of datasets that the corresponding algorithm was the best performer out of the total of 36 runs on a dataset. The average rank denotes the ranking of the given algorithm compared to the others overall in the scenario.

Table 9. Scenario one summary of 12 algorithms on 12 datasets for all short-term horizons

| Metric | Algorithms | | | | | | | | | | | |
|---|---|---|---|---|---|---|---|---|---|---|---|---|
| | Naive | MLP | LSTM | XGBoost | TCN | AutoARIMA | Block GRU | Informer | TimesNet | N-BEATS | iTransformer | PatchTST |
| AVG ERR | 0.1155 | 0.1129 | 0.1038 | 0.0983 | 0.1005 | 0.3558 | **0.0801** | 0.1710 | 0.1508 | 0.1058 | 0.1473 | 0.0883 |
| WIN | 8 | 0 | 3 | 0 | 0 | 0 | **10** | 0 | 0 | 9 | 0 | 6 |
| AVG Rank | 6.44 | 4.18 | 4.25 | 7.45 | 6.32 | 11.85 | **2.96** | 9.75 | 8.22 | 3.18 | 9.72 | 3.66 |

Figure 8 shows the CD diagram for scenario one. The cliques are formed using a pairwise Wilcoxon test. The existence of a clique between a pair of algorithms means that they are not significantly different from each other over tested datasets. Figure 9 is the MCM matrix illustrating the pairwise comparison test between six top-performing algorithms in this scenario. Block GRU emerged as the top-performer in all tests showing its dominance in short-term univariate forecasting. The fact that GRU could beat advanced and sophisticated algorithms is notable. Another interesting result here is that the Naive forecaster won eight out of the twelve experiments. While it did not emerge as a top-performer in CD and MCM diagrams, the number of wins shows that many algorithms performed worse than the Naive forecaster in some cases. It also indicates the importance of choosing a good algorithm for a TSF task to be able to capture the underlying patterns. N-BEATS and PatchTST algorithms are also among the top-performers in this scenario and can be considered for such use-cases.

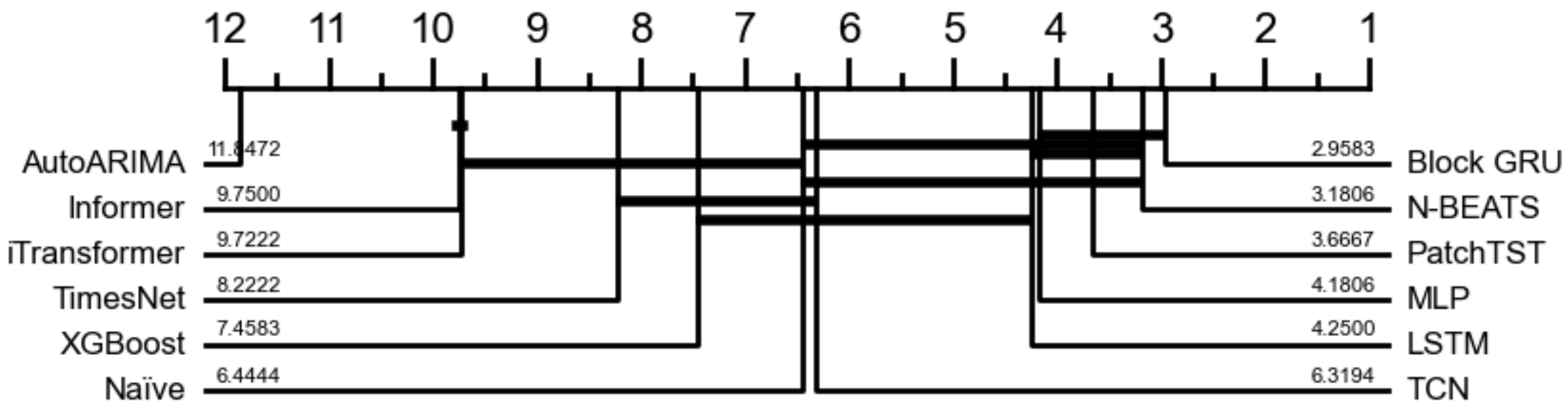

Figure 8. CD diagrams for 12 univariate algorithms on 12 datasets for all short-term horizons

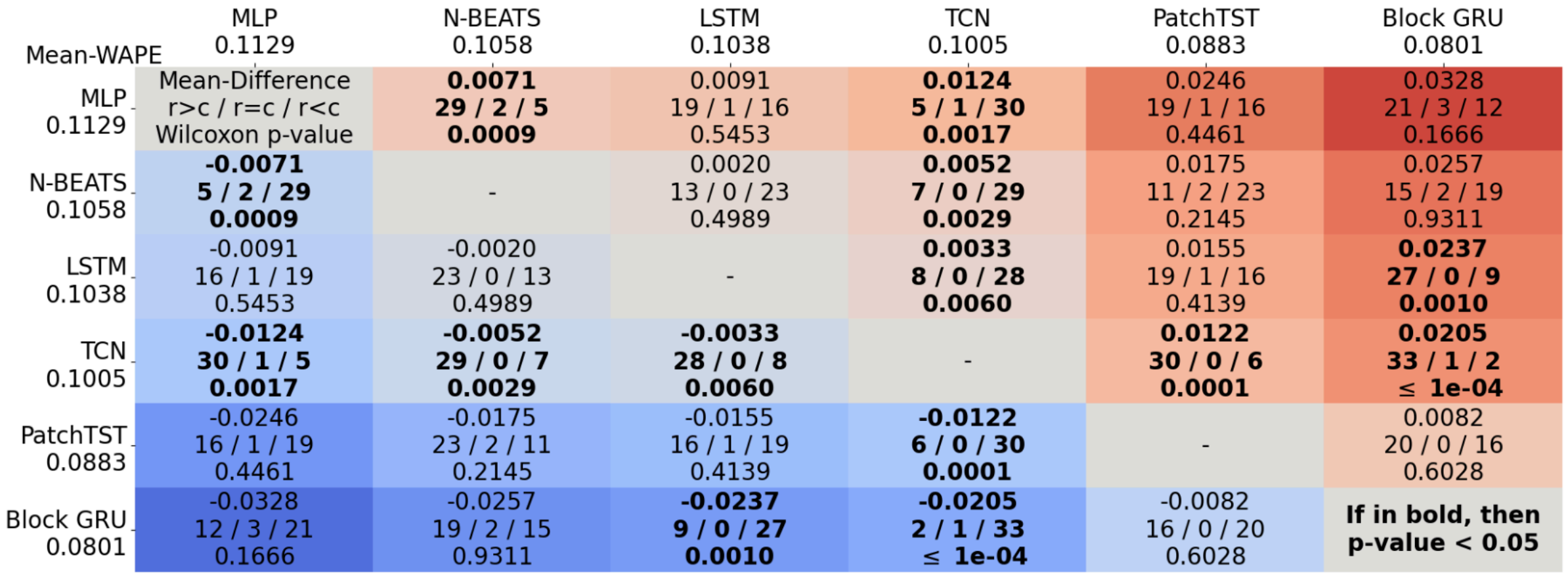

Figure 9. MCM for top six univariate algorithms on 12 datasets for all short-term horizons

Table 10 summarizes the scenario with two results. PatchTST is the best-performing algorithm in this scenario with 15 wins and lowest average rank. The MLP algorithm is also among top-performers in this scenario. An interesting result in this scenario is that Block GRU performance has degraded, showing that although it performed well for short-term forecasting, it struggled in capturing long-term dependencies.

Table 10. Scenario two summary of 12 algorithms on 12 datasets for all long-term horizons

| Metric | Algorithms | | | | | | | | | | | |
|---|---|---|---|---|---|---|---|---|---|---|---|---|
| | Naive | MLP | LSTM | XGBoost | TCN | AutoARIMA | Block GRU | Informer | TimesNet | N-BEATS | iTransformer | PatchTST |
| AVG ERR | 0.2701 | **0.2035** | 0.3262 | 0.2229 | 0.2293 | 0.3717 | 0.2388 | 0.2957 | 0.2241 | 0.2217 | 0.2187 | 0.2070 |
| WIN | 5 | 2 | 2 | 5 | 1 | 0 | 1 | 1 | 0 | 2 | 2 | **15** |
| AVG Rank | 6.92 | 4.21 | 8.28 | 6.97 | 6.85 | 10.21 | 6.74 | 8.71 | 5.69 | 4.76 | 6.03 | **2.64** |

MLP, PatchTST, and iTransformer algorithms are the top-performers based on the results illustrated in Figure 10 and Figure 11. This shows the power of transformer-based architectures for capturing long-term dependencies and long-term forecasting.

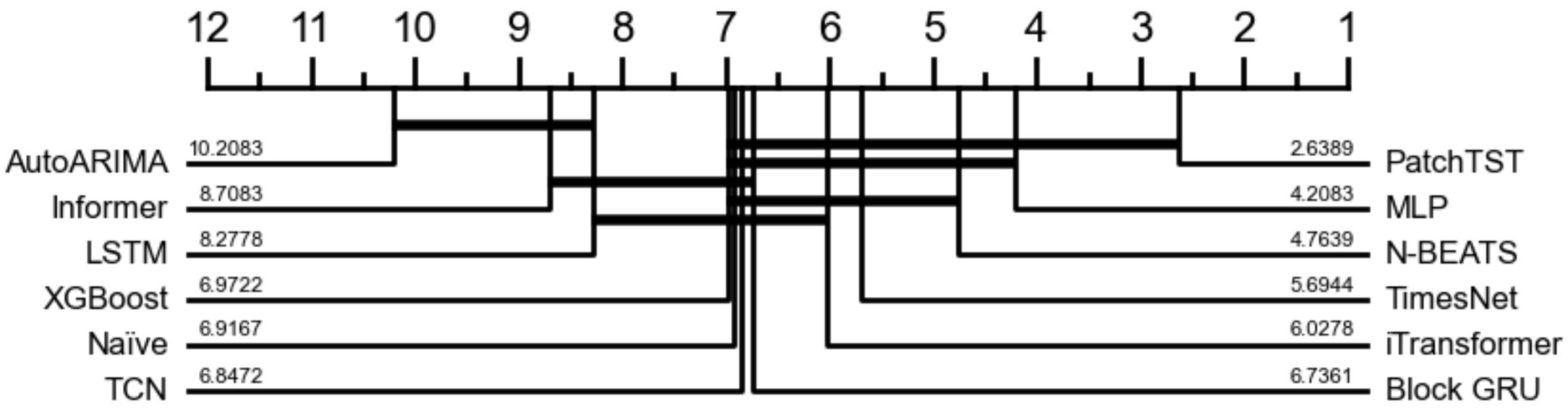

Figure 10. CD diagrams for 12 univariate algorithms on 12 datasets for all long-term horizons

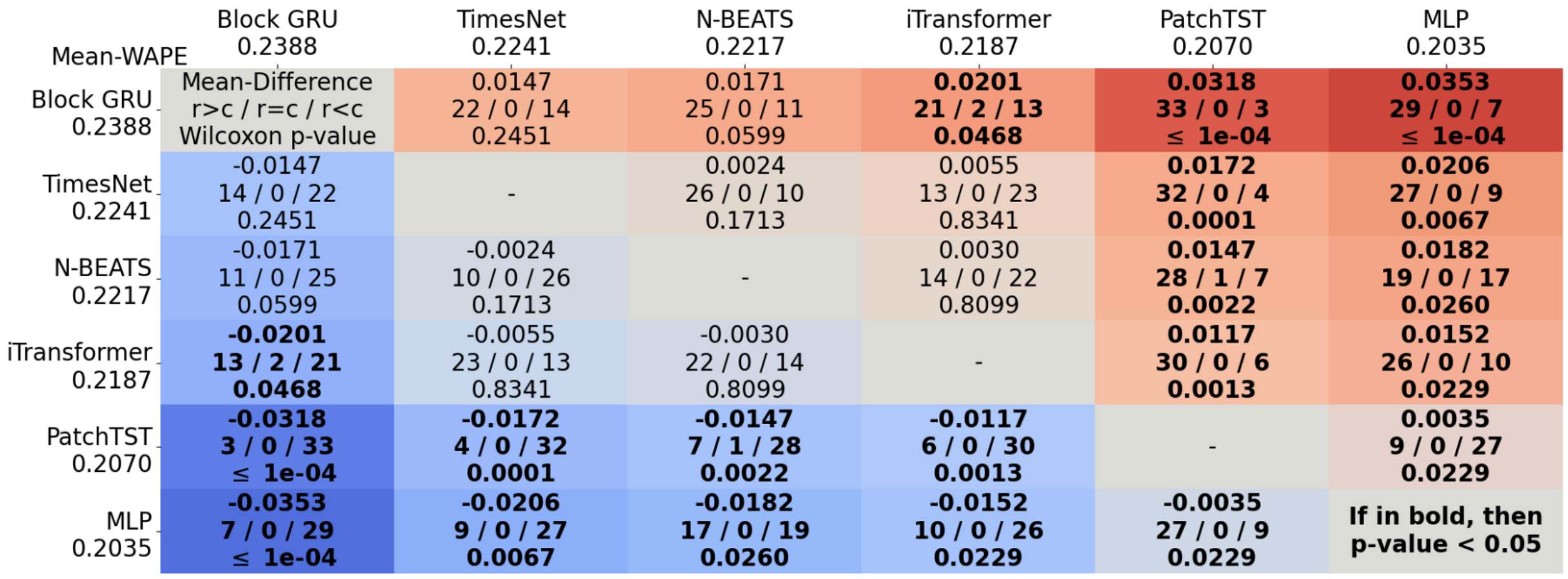

Figure 11. MCM for top six univariate algorithms on 12 datasets for all long-term horizons

Overall, for univariate TSF problems, transformers and MLP-based architectures are showing the best results and PatchTST can be chosen as the most robust algorithm for all horizons. These results are consistent with the box plots in Figure 12 showing PatchTST with the lowest overall mean. Figure 12 also showcases the robustness of different algorithms, where smaller boxes indicate more robustness.

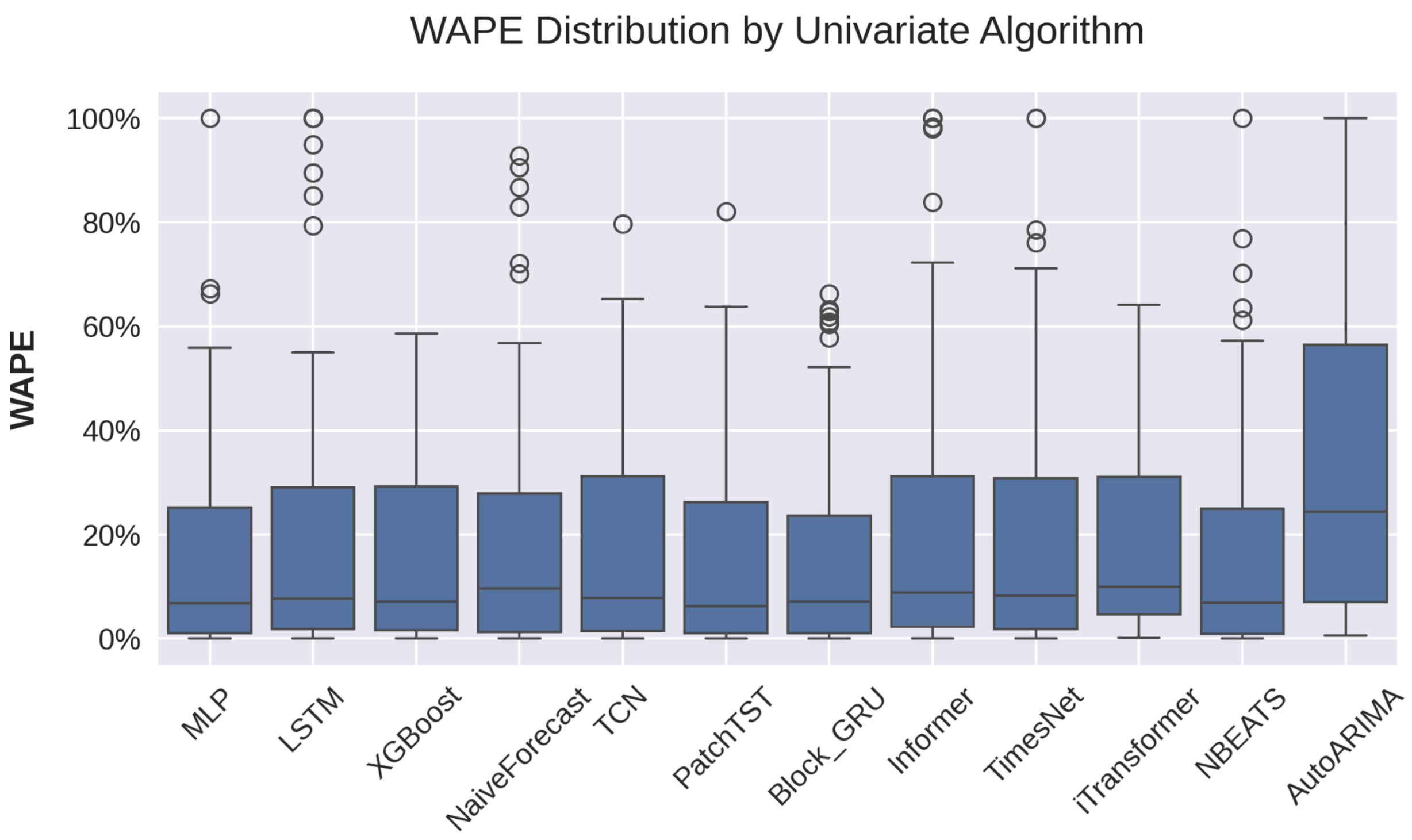

Figure 12. WAPE distribution box-plots for univariate forecasting problem

Table 11 summarizes the experimental results from scenario three. In this scenario, MLP-based architectures are showing the best results having N-HITS and MLP algorithms among the top-performers in all metrics. Figure 13 and Figure 14 agree with this result. TFT and XGboost are showing close performance for the third place.

Table 11. Scenario three summary of 11 algorithms on 11 datasets for all multivariate short-term horizons

| Metric | Algorithms | | | | | | | | | | |
|---|---|---|---|---|---|---|---|---|---|---|---|
| | Naive | MLP | TCN | Block GRU | XGBoo st | Dlinear | BiTCN | TSMix erX | TiDE | N-HIT S | TFT |
| AVG ERR | 0.1767 | 0.1389 | 0.3678 | 0.1705 | 0.1394 | 0.6474 | 0.2088 | 0.1931 | 0.2245 | **0.1324** | 0.1537 |
| WIN | 5 | 3 | 0 | 5 | 3 | 0 | 0 | 0 | 0 | **13** | 2 |
| AVG Rank | 4.94 | 3.07 | 8.71 | 6.50 | 4.86 | 10.15 | 8.06 | 5.48 | 7.86 | **2.20** | 4.15 |

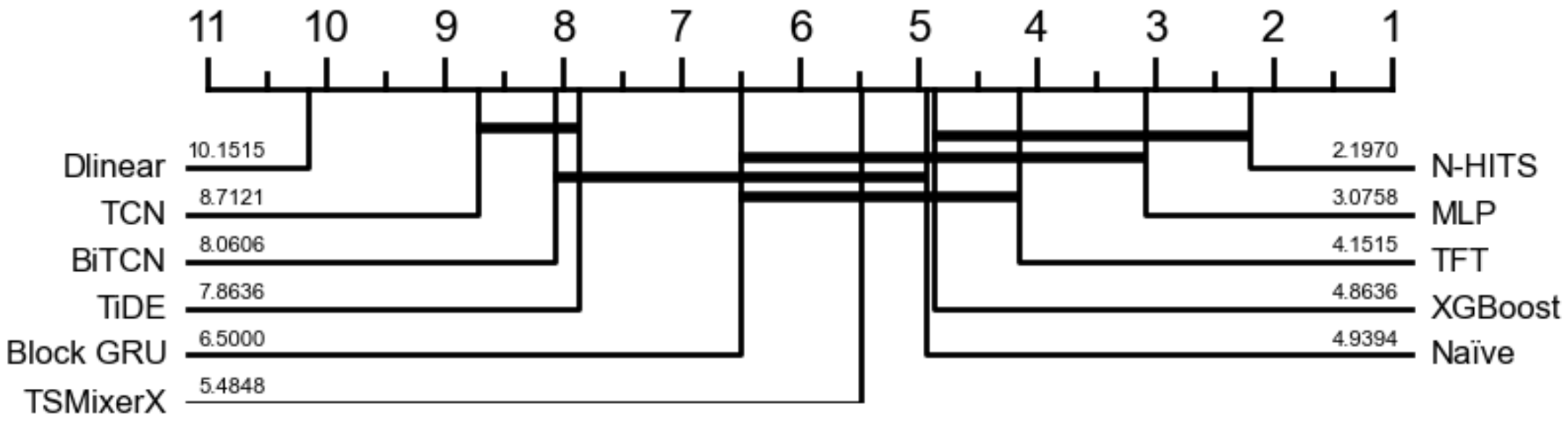

Figure 13: CD diagrams for 11 multivariate algorithms on 11 datasets for all multivariate short-term horizons

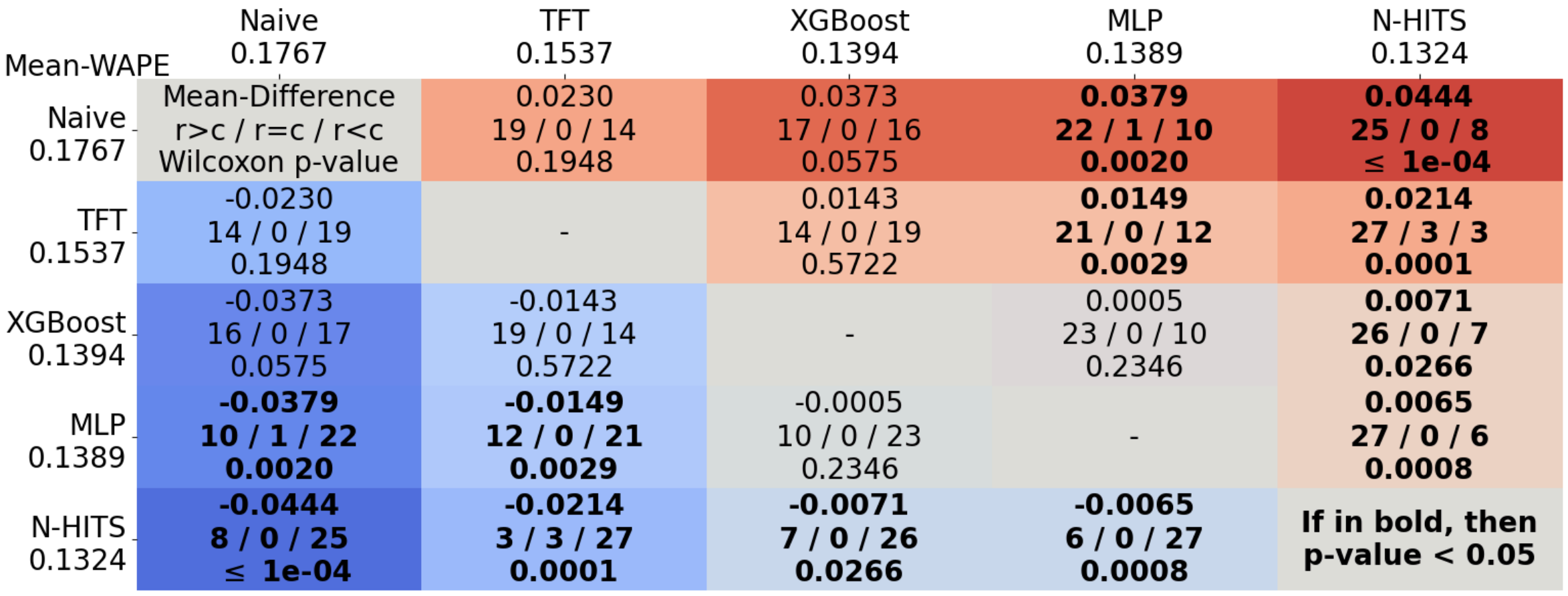

Figure 14: MCM for top five multivariate algorithms on 11 datasets for all multivariate short-term horizons

Table 12 summarizes the results from scenario four. This scenario is the most complex for this study, requiring the handling of both long-term forecasting and multivariate input time-series. However in this scenario, MLP-based architectures dominated the results and showed the best results having TiDE and TSMixerX algorithms among the top-performers in all metrics. XGBoost's performance also shows promising results in this scenario. It should be noted that the experimental hardware ran out of memory running the TFT algorithm for forecasting horizon of 672 and WAPE equal to 1 while calculating the averages.

Table 12. Scenario three summary of 11 algorithms on 11 datasets for all long-term horizons

| Metric | Algorithms | | | | | | | | | | |
|---|---|---|---|---|---|---|---|---|---|---|---|
| | Naive | MLP | TCN | Block GRU | XGBoost | Dlinear | BiTCN | TSMixerX | TiDE | N-HITS | TFT |
| AVG ERR | 0.3338 | 0.2789 | 0.4969 | 0.2977 | 0.2413 | 0.9287 | 0.2802 | 0.2400 | **0.2397** | 0.2742 | 0.5455 |
| WIN | 3 | 1 | 1 | 0 | **11** | 0 | 9 | 3 | 2 | 2 | 1 |
| AVG Rank | 6.10 | 5.09 | 8.11 | 6.36 | 4.82 | 10.62 | 5.68 | 3.77 | **3.75** | 4.32 | 7.36 |

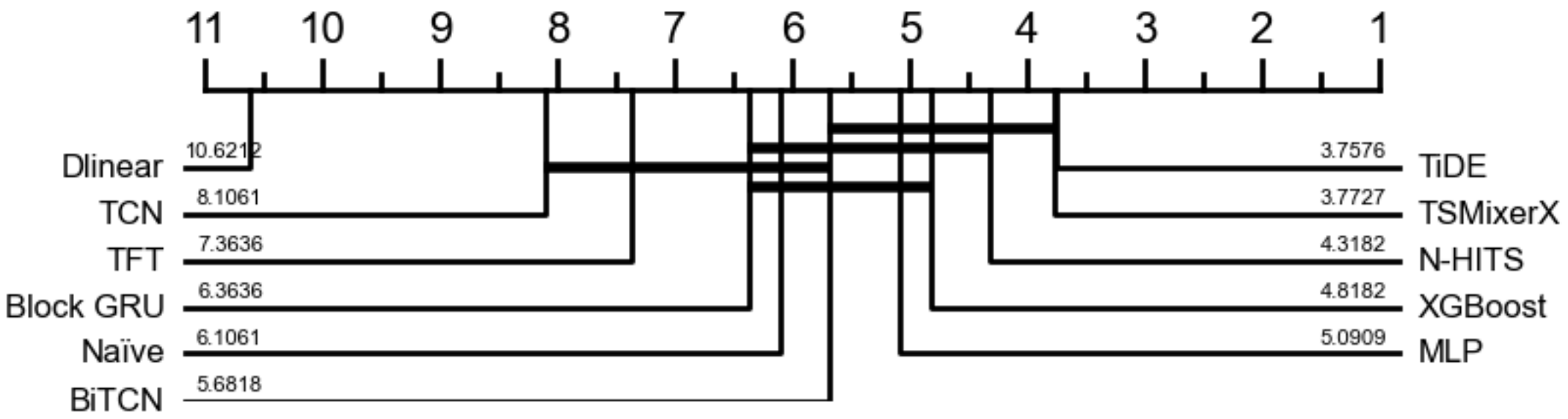

Figure 15: CD diagrams for 11 multivariate TSF algorithms on 11 datasets for all long-term horizons

| Mean-WAPE | MLP 0.2789 | N-HITS 0.2742 | XGBoost 0.2413 | TSMixerX 0.2400 | TiDE 0.2397 |
|---|---|---|---|---|---|
| MLP 0.2789 | Mean-Difference<br>r>c / r=c / r<c<br>Wilcoxon p-value | 0.0047<br>18 / 0 / 15<br>0.9578 | 0.0376<br>16 / 1 / 16<br>0.2312 | **0.0389<br>22 / 0 / 11<br>0.0137** | **0.0392<br>22 / 1 / 10<br>0.0212** |
| N-HITS 0.2742 | -0.0047<br>15 / 0 / 18<br>0.9578 | - | 0.0329<br>16 / 0 / 17<br>0.2141 | **0.0342<br>19 / 0 / 14<br>0.0074** | **0.0345<br>21 / 0 / 12<br>0.0021** |
| XGBoost 0.2413 | -0.0376<br>16 / 1 / 16<br>0.2312 | -0.0329<br>17 / 0 / 16<br>0.2141 | - | 0.0013<br>18 / 1 / 14<br>0.3000 | 0.0016<br>18 / 0 / 15<br>0.7509 |
| TSMixerX 0.2400 | **-0.0389<br>11 / 0 / 22<br>0.0137** | **-0.0342<br>14 / 0 / 19<br>0.0074** | -0.0013<br>14 / 1 / 18<br>0.3000 | - | 0.0003<br>14 / 0 / 19<br>0.2798 |
| TiDE 0.2397 | **-0.0392<br>10 / 1 / 22<br>0.0212** | **-0.0345<br>12 / 0 / 21<br>0.0021** | -0.0016<br>15 / 0 / 18<br>0.7509 | -0.0003<br>19 / 0 / 14<br>0.2798 | **If in bold, then<br>p-value < 0.05** |

Figure 16: MCM for top five multivariate TSF algorithms on 11 datasets for all long-term horizons

Overall, for multivariate TSF problems, MLP-based architectures showed the best results with no single algorithm dominating for all horizons. These results are consistent with the Figure 15 CD diagram and the MCM matrix in Figure 16. Figure 17 also shows the superiority and robustness of MLP-based algorithms such as N-HITS for multivariate forecasting problems.

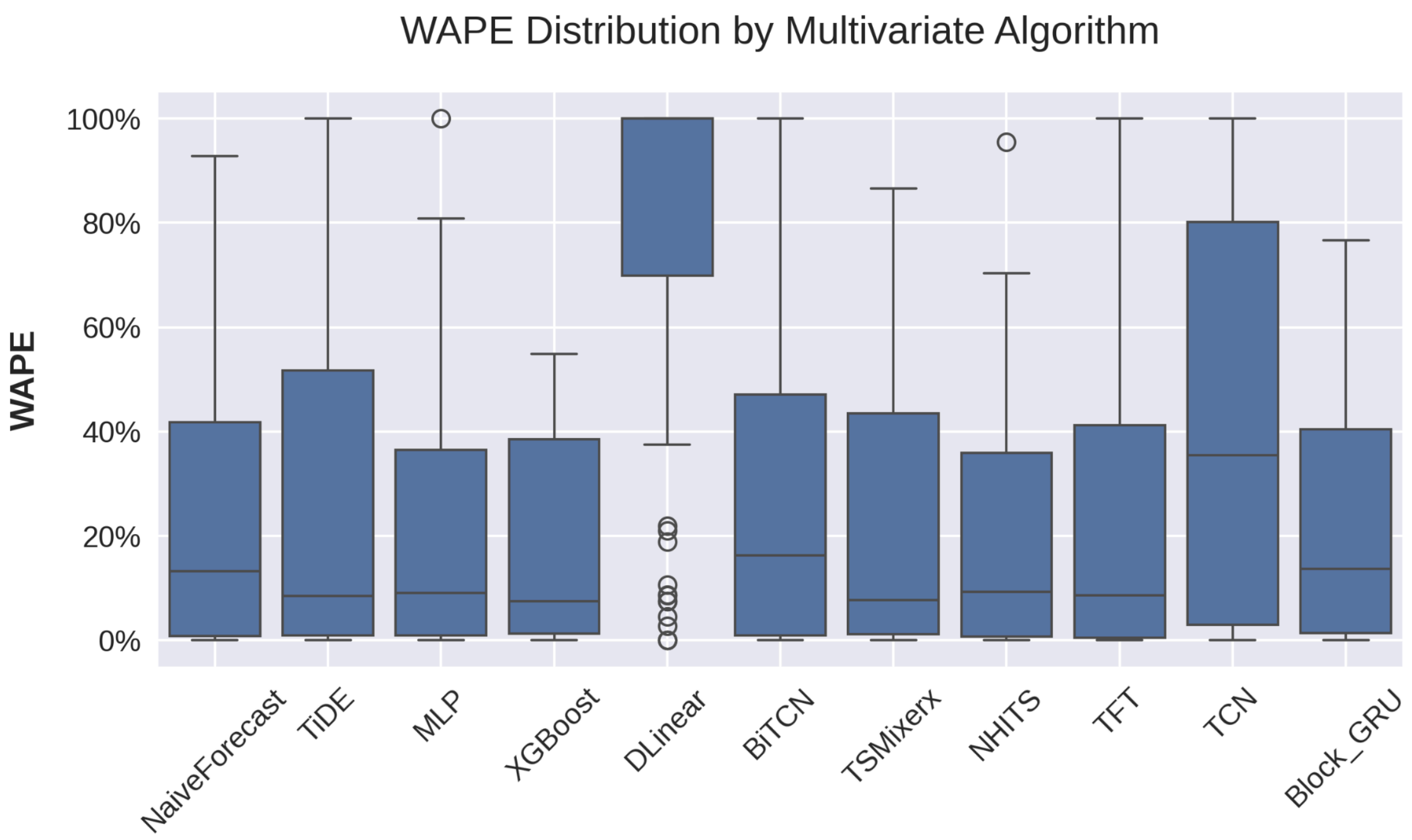

Figure 17. WAPE distribution box-plots for multivariate forecasting problem

## 4.2. Computational Time and Expense Evaluation

As highlighted in previous sections results, there are some situations where no specific algorithm shows significant superiority over other candidates. To improve decision making other evaluation metrics should also be considered in addition to the forecasting error. First, the total runtime of the algorithms are calculated for both univariate and multivariate forecasting problems. This metric includes both training and testing time of a given algorithm, averaged over all datasets. As depicted in the vertical axis, the total range of runtime varied from 100 milliseconds of Naive forecasting to about two days for Block GRU algorithm. These results show that although some algorithms such as Block GRU might deliver low forecasting errors, they might not be feasible for large datasets which are prevalent in the manufacturing domain. The results are shown in Figure 18.

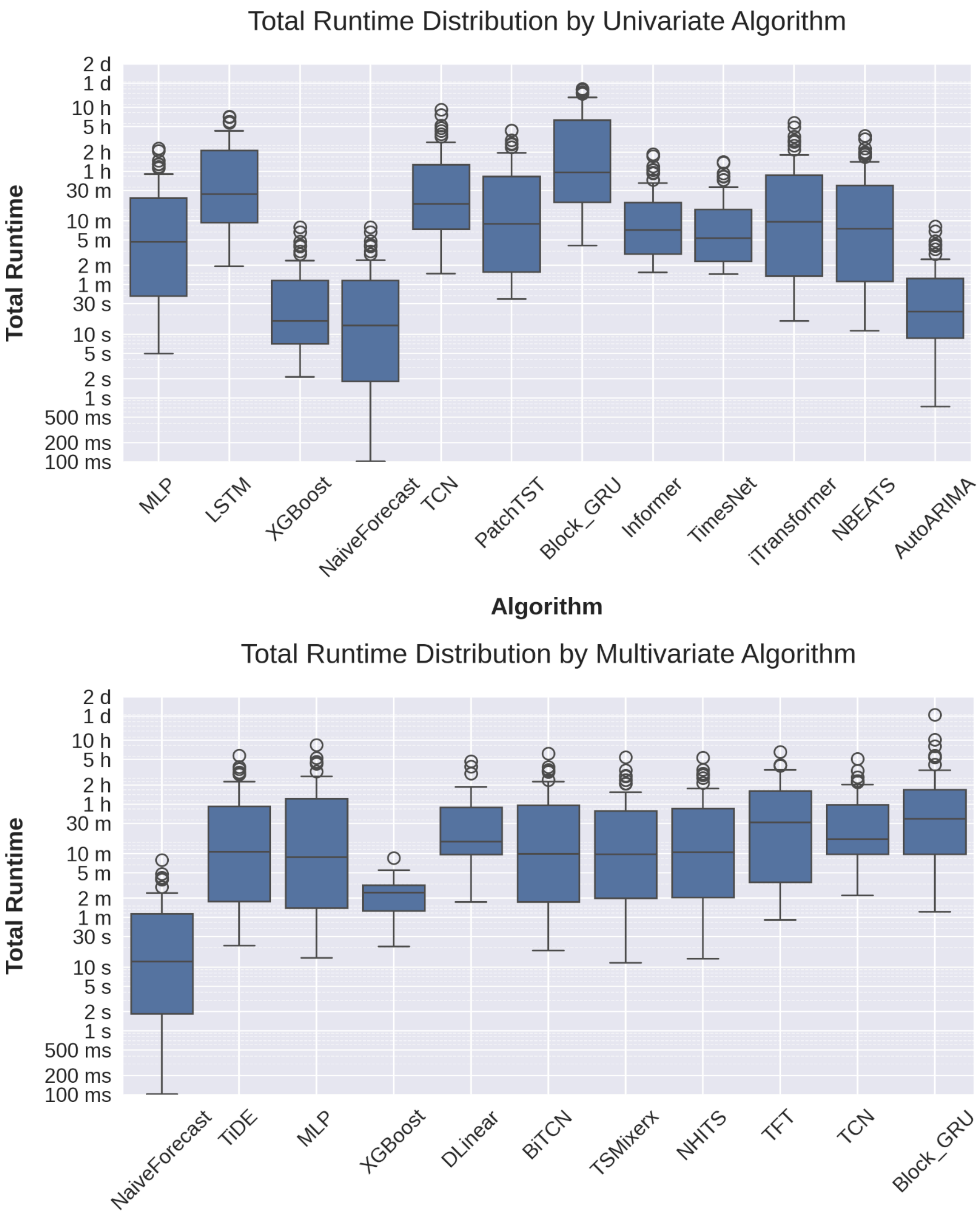

Figure 18. Total runtime comparison of Algorithms in different problems

It is difficult to compare runtimes and computational expenses of different algorithms for several reasons such as differences in Python package software and available hardware resources. Moreover, some algorithms had been run on CPU while some others ran on GPU and some ran on both in parallel which makes it difficult to know how this affects the final runtime. Keeping these considerations in mind, this approach is practical to gain insight into relative algorithm performance.

Next, the average runtime of the algorithms was plotted against the average WAPE and average peak memory to visually compare the algorithms in Figure 19. These results can be considered only as an indicator for scalability. If the problem is associated with a large dataset or if a need to scale up the system in the future is anticipated, the forecaster' scalability should be considered. The lower-left part of the plots in Figure 19 is the ideal location featuring both low computational cost and low forecasting errors. For example, in the univariate problem XGBoost is showing competitive performance with very low runtime. However, a trade off can be made to have lower WAPE with the PatchTST algorithm at the expense of both computational runtime and memory. Choosing between these options will depend on several factors in the application and the use case at hand; such as available computational resources and operational constraints. For instance, in real-time forecasting scenarios, a faster runtime may take precedence over marginal improvements in the forecasting error, making XGBoost a more practical choice. Conversely, in scenarios where the performance is critical and computational resources are not a limiting factor, investing in more resource-intensive algorithms like PatchTST could be justified.

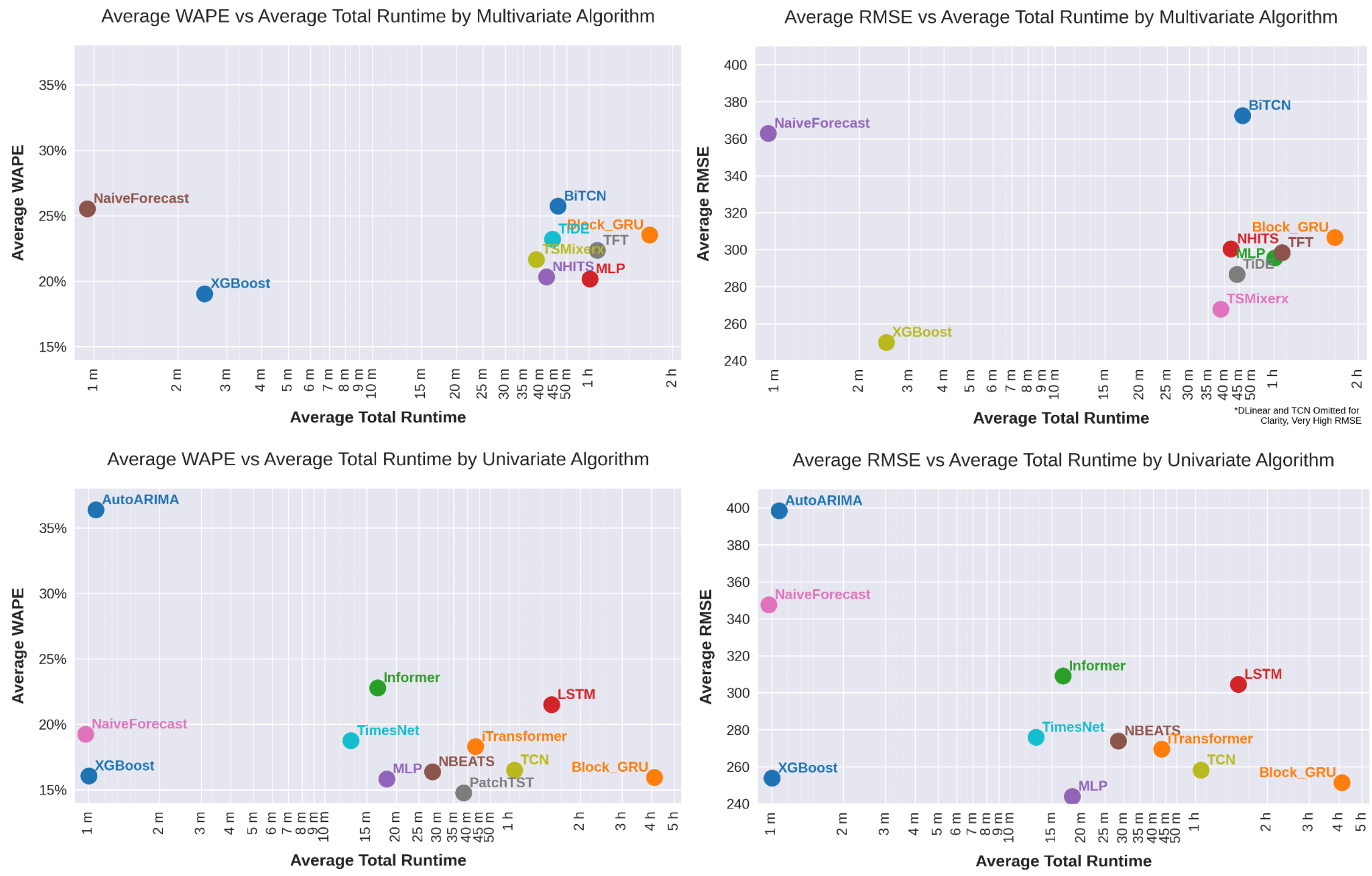

Figure 19. Average peak memory vs total runtime

## 4.3. Application domains Analysis

In this subsection, the results are analyzed for different groups of datasets to see if there are any meaningful differences between the best algorithms for different groups. The categorization can be found in Table 13. One major insight derived from this investigation is the performance of forecasting algorithms across different manufacturing domains. This analysis can aid in determining the feasibility of utilizing forecasting for different use cases in manufacturing. Within the datasets utilized in the previous experiments, three major manufacturing domains were extracted. The first domain includes that of *demand forecasting*, which focuses on predicting the market demand for products in the future. This domain is crucial for manufacturers to plan accordingly to satisfy market requirements. The second domain includes *forecasting sensor data*, this area is pivotal to predicting the proper functioning of a smart manufacturing system. The final domain is that of *energy consumption prediction*. This domain is crucial in minimizing energy consumption, decarbonisation efforts, and lowering overall manufacturing costs.

Table 13. Dataset categorization

| Dataset Name | Category | Scenarios |
|---|---|---|
| Gas Sensor Temperature Modulation | Sensors | 1, 2, 3, 4 |
| Gas Sensor Dynamic Gas Mixtures | Sensors | 1, 2, 3, 4 |
| Appliances Energy | Energy Consumption | 1, 2, 3, 4 |
| Electricity | Energy Consumption | 3, 4 |
| ETTm2 | Sensors | 1, 2, 3, 4 |
| ETTm1 | Sensors | 1, 2, 3, 4 |
| ECL | Energy Consumption | 1, 2, 3, 4 |
| ISO-NY | Energy Consumption | 1, 2 |
| Monroe Water Treatment Plant | Energy Consumption | 1, 2, 3, 4 |
| Seoul Bike Demand | Demand | 1, 2, 3, 4 |
| AI4I 2020 | Sensors | 1, 2, 3, 4 |
| Steel Industry | Energy Consumption | 1, 2, 3, 4 |
| Brent Oil Prices | Demand | 1,2 |

To evaluate the performance of each algorithm across the three domains, the average WAPE was calculated across all three forecasting horizons for each dataset and for all datasets in each domain. The results are exhibited in Table 14.

Table 14. Scenario 1 and 2 Domain Results

| Domain | Naive | | MLP | | LSTM | | XGBoost | | TCN | | AutoARIMA | |
|---|---|---|---|---|---|---|---|---|---|---|---|---|
| | S1 | S2 | S1 | S2 | S1 | S2 | S1 | S2 | S1 | S2 | S1 | S2 |
| Demand | 0.273 | 0.454 | 0.170 | 0.439 | 0.171 | 0.696 | 0.195 | 0.460 | 0.181 | 0.437 | 0.636 | 0.636 |

| Domain | Naive | | MLP | | TCN | | XGBoost | | TiDE | | DLinear | |
|---|---|---|---|---|---|---|---|---|---|---|---|---|
| Sensors | 0.037 | 0.196 | 0.027 | 0.156 | 0.027 | 0.233 | 0.041 | 0.204 | 0.038 | 0.185 | 0.310 | 0.358 |
| Energy Consumption | 0.131 | 0.270 | 0.175 | 0.156 | 0.153 | 0.271 | 0.117 | 0.146 | 0.132 | 0.190 | 0.289 | 0.289 |

| Domain | Block GRU | | Informer | | TimesNet | | N-BEATS | | iTransformer | | PatchTST | |
|---|---|---|---|---|---|---|---|---|---|---|---|---|
| | S1 | S2 | S1 | S2 | S1 | S2 | S1 | S2 | S1 | S2 | S1 | S2 |
| Demand | 0.172 | 0.506 | 0.328 | 0.540 | 0.242 | 0.407 | 0.154 | 0.447 | 0.285 | 0.427 | 0.175 | 0.479 |
| Sensors | 0.027 | 0.188 | 0.053 | 0.173 | 0.040 | 0.172 | 0.028 | 0.156 | 0.088 | 0.177 | 0.028 | 0.151 |
| Energy Consumption | 0.096 | 0.183 | 0.226 | 0.321 | 0.225 | 0.203 | 0.165 | 0.197 | 0.151 | 0.176 | 0.113 | 0.154 |

Within Scenarios one and two, univariate forecasting, the domain with the lowest WAPE across the multiple experiments was that of Sensors. A total of 17 of the 24 experiments across the different algorithms achieved the best results with datasets relating to sensors. In some experiments the winning margin was very narrow. Nonetheless, these results might be indicative of the specific use cases in manufacturing that forecasting algorithms can be beneficial for. Predicting sensor data can be useful in manufacturing to determine the operational status of the system as well as the ability to predict any defects to the final product. Experiments attempting to predict demand yielded the highest error percentages showcasing the difficulty in predicting fluctuating market demands.

The same analysis was conducted to the results obtained from Scenarios 3 and 4. Following the aggregation of the error percentage for each scenario, Table 15 showcases the results based on each category of dataset.

Table 15. Scenario 3 and 4 Domain Results

| Domain | Naive | | MLP | | TCN | | Block GRU | | XGBoost | | Dlinear | |
|---|---|---|---|---|---|---|---|---|---|---|---|---|
| | S3 | S4 | S3 | S4 | S3 | S4 | S3 | S4 | S3 | S4 | S3 | S4 |
| Demand | 0.518 | 0.751 | 0.383 | 0.803 | 0.814 | 0.920 | 0.586 | 0.739 | 0.332 | 0.492 | 1.000 | 1.000 |
| Sensors | 0.035 | 0.196 | 0.032 | 0.193 | 0.063 | 0.332 | 0.060 | 0.201 | 0.045 | 0.206 | 0.424 | 1.000 |
| Energy Consumption | 0.250 | 0.388 | 0.196 | 0.260 | 0.584 | 0.577 | 0.198 | 0.306 | 0.195 | 0.226 | 0.800 | 0.841 |

| Domain | BiTCN | | TSMixerX | | TiDE | | N-HITS | | TFT | |
|---|---|---|---|---|---|---|---|---|---|---|
| | S3 | S4 | S3 | S4 | S3 | S4 | S3 | S4 | S3 | S4 |
| Demand | 0.488 | 0.706 | 0.444 | 0.622 | 0.564 | 0.585 | 0.349 | 0.728 | 0.519 | 0.809 |
| Sensors | 0.089 | 0.146 | 0.041 | 0.163 | 0.059 | 0.160 | 0.027 | 0.162 | 0.034 | 0.455 |
| Energy Consumption | 0.273 | 0.329 | 0.294 | 0.240 | 0.322 | 0.250 | 0.195 | 0.296 | 0.200 | 0.583 |

These results are similar to those in scenario 1 and 2 where the sensor domain witnessed the lowest average error. From these analyses, we can see that forecasting, short term and long term, operate best in the manufacturing domain when dealing with sensor forecasting use cases.

The results also show the best performing algorithms for each domain. Within the Demand category, N-Beats yielded the lowest WAPE for univariate forecasting while XGBoost was the top performer in multivariate forecasting. Similarly, N-Beats and N-HITS were the best performers in the Sensors category for univariate and multivariate respectively. It is worth noting that the performance of LSTM was almost identical to N-Beats within the univariate results. Finally, Block GRU and N-HITS performed the best within the energy consumption univariate and multivariate forecasting. While no concrete conclusion can be made from these results, the results from these investigations can be a pivotal platform to narrow down further the algorithm selection process across multiple manufacturing use cases.

# 5. Conclusion, Future Works, and Limitations

In addition to algorithm development, the effort of experimentally validating the numerous existing models and comparing their performance is equally important. This is of immense value to researchers and practitioners, as it narrows down their choices, and provides insight into the advantages and disadvantages of different models and provides the ability to further customize their choice by comparing their problem space (dataset) with the performance on the most similar evaluation dataset. This kind of study also sets benchmarks which look to channel research directions into more promising avenues.

This study provides an experimental evaluation of state-of-the-art TSF algorithms in the context of smart manufacturing systems. To the best of the authors knowledge, tt is the largest empirical study of TSF algorithms in the manufacturing domain to-date. By applying these algorithms on thirteen manufacturing-related datasets, this work addresses a critical gap in the literature and offers valuable insights for smart manufacturing systems. The field of TSF has remarkably evolved in recent years, shifting from traditional statistical techniques to advanced ML ones. In the manufacturing domain, this evolution has been driven by the transformation towards a data-rich Industry 4.0, where various sensors and enhanced connectivity have eliminated many of the data constraints of the past.

While statistical techniques such as ARIMA and exponential smoothing are reliable approaches, they often fall short when dealing with complex, multivariate, and non-stationary data without proper preprocessing. The emergence of ML methods has addressed many of their limitations by automatically learning the underlying patterns and estimating the data relationships without any assumptions. This is particularly valuable in the manufacturing domain where the input data may be multivariate and non-linear.

This analysis also reveals an important point regarding model complexity. As demonstrated in the results, simple techniques like XGBoost and MLP-based architectures can outperform more complex Transformer-based architectures in certain scenarios, particularly for short-term forecasting tasks. For long-term forecasting, transformer-based models like PatchTST showed competitive performance in capturing long term temporal dependencies. These findings confirm the claim that the current Transformer-based algorithms are not generating the best performance for TSF tasks despite their massive success in other fields like NLP and their advanced architecture [58]. This apparent and partially unexpected underperformance may be attributed to subtle differences between time-series and natural language data. While transformers initially built high expectations through their impressive performance in NLP, they have not fully delivered in the manufacturing time-series domain. Thus, there is a need for more sophisticated, domain-specific transformer architectures tailored to the unique characteristics of time-series forecasting in manufacturing.

Additionally, the results of this study show that model complexity should be carefully balanced against the specific requirement of the forecasting task at hand. The study's emphasis on out-of-the-box algorithm implementations highlights the practical applicability of these algorithms for Industry 4.0. By minimizing preprocessing and hyperparameter tuning, the solutions that are effective and accessible for practitioners with limited ML expertise are prioritized.

Finally, comparing this study to our previous work on experimental evaluations of TSC algorithms [28], it can be concluded that TSF is generally a broader, more complex problem and a less mature field compared to TSC within the manufacturing domain. This highlights the need for more research efforts to advance this promising area further.

Several topics and subtopics were not discussed in this work and are considered in *future works*. These topics include but are not limited to investigating preprocessing techniques and their effect on manufacturing forecasting performance, investigation of LLM-based techniques for manufacturing TSF tasks, and integration of domain specific knowledge into TSF frameworks to improve forecasting performance in manufacturing contexts.

There are several *limitations* that must be considered when interpreting the results of this paper. This paper was written under certain assumptions, timelines, and resource limitations, with a focus on providing a comprehensive TSF framework in smart manufacturing systems. While the complete removal of subjectivity is impossible, we intend to articulate the used process and methodology transparently, to enable the audience to understand our intent, understanding, potential biases, and their influence on the content of this paper. In particular, the following limitations are worth noting:

A significant challenge is the scarcity of suitable, accessible, and public datasets for AI and ML applications in manufacturing. This limitation is a substantial barrier to research reproducibility and algorithm validation.

In the algorithm selection, it was assumed that newer algorithms within the same categories would outperform their predecessors. While this assumption is intuitively correct, it may not hold true in all cases, as some algorithms developed for specific problems may exist that defy this assumption.

However, studies like this involve making a number of decisions about how information is collected and analyzed, how experiments are conducted, etc. Often there is no one "correct" approach. Instead, this study focuses on being as transparent as possible in explaining all the steps to increase the clarity and reproducibility of this work.

Although different evaluation metrics had been utilized in this study, there might be some limitations and biases in the used evaluation metrics. Finally, the results discussed in this paper are only applicable to the studied domain and problem and the findings are not generalizable to other domains.

Finally, several preprocessing steps were carried out for the datasets to make them ready for algorithm ingestion. These steps include reindexing, resampling, scaling, interpolation, etc. These steps were done to prevent algorithm errors and to manage the dataset sizes for computational efficiency. We made the decision to choose the best technique in each step to the best of our knowledge in a standard way. However, we acknowledge there could be numerous other ways to choose these techniques, and our choice might have introduced some biases to the final results. We tried to transparently explain all the steps so the reader could understand potential biases.

# Declaration of competing interest

The authors declare the following financial interests/personal relationships which may be considered as potential competing interests: Thorsten Wuest, Ramy Harik reports financial support was provided by National Science Foundation. If there are other authors, they declare that they have no known competing financial interests or personal relationships that could have appeared to influence the work reported in this paper.

# Data availability

Data will be made available upon request.

# Acknowledgment

This material is based upon work supported by the National Science Foundation under Grant No. 2119654. Any opinions, findings, conclusions, or recommendations expressed in this material are those of the author(s) and do not necessarily reflect the views of the National Science Foundation.

# Appendix A

In Appendix A, more detail of the datasets used in this study is provided;

**AI4I 2020:** This dataset is a Predictive Maintenance Dataset is a synthetic dataset that reflects real predictive maintenance data encountered in industry. It has 6 different features that monitor the Air temperature, process temperature, rotational speed, torque, and tool wear of a simulated machine. It has two target variables to predict machine and tool wear failure.

**Appliance Energy:** It consists of 138 time series derived from the Appliances Energy Prediction dataset available in the UCI repository. Each time series includes 24 features, such as temperature and humidity readings from nine rooms in a house, collected using a ZigBee wireless sensor network. Additionally, it incorporates weather and climate data, including temperature, pressure, humidity, wind speed, visibility, and dew point, sourced from Chievres airport. The dataset represents 10-minute averaged data over a period of 4.5 months.

**Brent Oil Prices:** The aim of this dataset and work is to predict future Crude Oil Prices based on the historical data available in the dataset. The dataset contains daily Brent oil prices from 17th of May 1987 until the 13th of November 2022. It is a univariate dataset with the oil price as the only feature with 9011 instances.

**ECL:** This dataset focuses on electricity consumption of 370 different clients. There are 140,256 instances in this dataset that were created for energy consumption prediction purposes.

**ETTh1 and ETTm2:** These datasets are collected from different electric transformers from two regions of a province of China with the aim of studying the oil temperature. ETTm2 data was collected every 15 minutes between July 2016 and July 2018 and ETTh1 data was collected each hour. Both datasets have 8 features to outline the load of the transformers and the oil temperature.

**Gas sensor dynamic gas mixtures:** This dataset contains time series data from 16 chemical sensors exposed to gas mixtures of varying concentrations, specifically Ethylene, Methane, Ethylene, and CO in air. The data, collected over 12-hour experiments at the ChemoSignals Laboratory at UC San Diego, includes continuous sensor conductivity readings sampled at 100 Hz under randomized concentration changes. The dataset was designed to challenge sensor response analysis, featuring varying concentration levels, complex transitions, and repeated sensor types to explore reproducibility and variability. It is intended to support sensor and AI research in tasks such as continuous monitoring, calibration transfer, and sensor failure analysis, and is available solely for non-commercial research purposes.

**Gas sensor temperature modulation**: This dataset contains time series data from 14 temperature-modulated metal oxide semiconductor (MOX) gas sensors exposed to dynamic mixtures of carbon monoxide (CO) and humid synthetic air in a controlled gas chamber. The platform utilized commercially available sensors (7 Figaro TGS 3870-A04 and 7 FIS SB-500-12) with voltage-modulated heating cycles, and sensor responses were sampled at 3.5 Hz using high-precision data acquisition systems. Gas mixtures were delivered through mass flow controllers (MFCs) from pressurized CO, dry air, and wet air streams, with CO concentrations ranging from 0–20 ppm and relative humidity levels between 15% and 75%. Each experiment lasted 25 hours, with 100 measurements collected over 13 sessions to analyze gas concentration and humidity effects, providing valuable data for sensor performance evaluation and AI-based analysis in chemical detection.

**Monroe Water Treatment Plant:** This is publically available data about the energy consumption of the Monroe Water Treatment Plant in the city of Bloomington, Indiana. It has eight features that reflect the different metrics of energy consumptions such as kilowatt hr.

**Steel Industry:** This dataset was created to analyze the trends in energy consumption of the steel industry. This dataset has 9 features such as Usage_kWh , Lagging_Current_Reactive, Leading_Current_Reactive_Power_kVarh , and CO2(tCO2). This data was collected across a year with a sampling rate of 15 minutes leading to 35041 instances.

**ISO-NY:** This is publically available data about the energy consumption of 11 different regions in New York city, NY. The energy consumption of the CENTRAL region was chosen for this study. The data was collected every 5 minutes. It is a univariate dataset containing only the energy load.

**Seoul Bike Demand:** This dataset contains weather information (Temperature, Humidity, Windspeed, Visibility, Dewpoint, Solar radiation, Snowfall, Rainfall), the number of bikes rented per hour and date information in a bike renting station in an urban area in seoul, Korea.

**Electricity:** This dataset contains 2075259 measurements collected every minute gathered in a house located in France between December 2006 and November 2010. It's a multivariate dataset with 7 features of global_active_power, global_reactive_power, voltage, intensity, sub_metering_1, sub_metering_2, and sub_metering_3.

# Appendix B

In Appendix B, additional evaluation metrics of RMSE are provided:

Table 16. Scenario 1 Forecasting Horizon 3 RMSE Results

| Domain | Naive | MLP | LSTM | XGBoost | TCN | AutoARIMA | Block GRU | Informer | TimesNet | N-BEATS | iTransformer | PatchTST |
|---|---|---|---|---|---|---|---|---|---|---|---|---|
| AI4I 2020 | 0.112 | 0.117 | 0.109 | 0.156 | 0.135 | 0.290 | 0.113 | 0.186 | 0.142 | 0.113 | 0.454 | 0.113 |
| Appliances Energy | 0.103 | 0.078 | 0.101 | 0.122 | 0.075 | 6.059 | 0.060 | 0.452 | 0.242 | 0.054 | 1.887 | 0.052 |
| Brent Oil Prices | 1.886 | 1.945 | 1.948 | 2.720 | 2.363 | 23.289 | 1.879 | 3.283 | 2.706 | 1.929 | 8.963 | 1.910 |
| ECL | 273.919 | 166.714 | 208.634 | 200.199 | 227.572 | 567.450 | 173.278 | 297.057 | 280.741 | 161.364 | 271.985 | 175.879 |
| ETTh1 | 0.945 | 0.881 | 0.908 | 1.140 | 1.107 | 5.169 | 0.899 | 1.415 | 1.093 | 0.870 | 1.961 | 0.887 |
| ETTm2 | 0.739 | 0.426 | 0.384 | 0.635 | 0.522 | 16.318 | 0.368 | 1.950 | 1.281 | 0.386 | 2.963 | 0.401 |
| Gas sensor dynamic gas mixtures | 47.914 | 26.784 | 18.223 | 27.331 | 22.513 | 908.774 | 17.794 | 93.104 | 67.885 | 23.908 | 325.721 | 27.418 |
| Gas sensor temperature modulation | 0.013 | 0.013 | 0.038 | 0.061 | 0.018 | 0.382 | 0.016 | 0.023 | 0.024 | 0.013 | 0.183 | 0.013 |
| ISO-NY | 46.657 | 40.992 | 40.586 | 47.864 | 43.653 | 456.751 | 37.610 | 97.642 | 69.917 | 37.910 | 132.236 | 39.335 |
| Monroe Water Treatment Plant | 1,844.642 | 1,323.545 | 1,267.229 | 1,391.462 | 1,332.033 | 1,835.131 | 1,259.302 | 2,007.284 | 1,570.189 | 1,369.634 | 1,379.673 | 1,323.715 |
| Seoul Bike Demand | 485.709 | 299.387 | 218.465 | 299.732 | 282.827 | 971.010 | 257.755 | 475.490 | 420.252 | 292.079 | 455.055 | 310.267 |
| Steel_industry_Usage_kWh | 17.912 | 43.010 | 16.235 | 12.813 | 14.005 | 43.834 | 12.010 | 25.759 | 28.998 | 20.925 | 19.757 | 22.803 |

Table 17. Scenario 1 Forecasting Horizon 6 RMSE Results

| Domain | Naive | MLP | LSTM | XGBoost | TCN | AutoARIMA | Block GRU | Informer | TimesNet | N-BEATS | iTransformer | PatchTST |
|---|---|---|---|---|---|---|---|---|---|---|---|---|
| AI4I 2020 | 0.145 | 0.149 | 0.148 | 0.217 | 0.201 | 2.903 | 0.149 | 0.201 | 0.176 | 0.152 | 0.603 | 0.161 |
| Appliances Energy | 0.187 | 0.142 | 0.150 | 0.235 | 0.161 | 6.092 | 0.154 | 0.450 | 0.275 | 0.118 | 2.128 | 0.218 |
| Brent Oil Prices | 2.648 | 2.783 | 2.629 | 3.896 | 3.363 | 23.288 | 2.597 | 4.612 | 3.376 | 2.721 | 8.851 | 2.825 |
| ECL | 470.433 | 218.164 | 260.687 | 258.374 | 308.378 | 567.400 | 237.074 | 368.915 | 304.829 | 214.434 | 274.394 | 179.155 |
| ETTh1 | 1.190 | 1.080 | 1.172 | 1.541 | 1.477 | 5.267 | 1.147 | 1.688 | 1.34 | 1.078 | 1.944 | 1.058 |
| ETTm2 | 1.286 | 0.719 | 0.730 | 1.134 | 1.079 | 16.814 | 0.632 | 1.530 | 1.162 | 0.675 | 2.873 | 0.757 |
| Gas sensor dynamic gas mixtures | 83.623 | 41.235 | 33.561 | 56.993 | 49.965 | 912.220 | 28.736 | 121.895 | 125.909 | 51.035 | 369.122 | 62.972 |
| Gas sensor temperature modulation | 0.017 | 0.019 | 0.026 | 0.076 | 0.026 | 0.382 | 0.026 | 0.041 | 0.026 | 0.018 | 0.193 | 0.019 |
| ISO-NY | 72.009 | 58.630 | 60.181 | 78.523 | 74.385 | 460.427 | 52.988 | 116.570 | 91.674 | 55.706 | 150.243 | 48.520 |
| Monroe Water Treatment Plant | 1,730.997 | 1,372.360 | 1,362.435 | 1,449.625 | 1,412.797 | 1,829.535 | 1,354.332 | 1,777.662 | 1,924.699 | 1,372.451 | 1,518.759 | 1,369.007 |
| Seoul Bike Demand | 568.694 | 361.272 | 418.777 | 392.034 | 350.061 | 970.647 | 391.712 | 660.472 | 478.729 | 340.448 | 449.334 | 350.516 |
| Steel_industry_Usage_kWh | 22.403 | 54.900 | 23.451 | 16.972 | 18.653 | 41.550 | 14.761 | 35.315 | 46.590 | 59.261 | 20.281 | 14.134 |

Table 18. Scenario 1 Forecasting Horizon 12  RMSE Results

| Domain | Naive | MLP | LSTM | XGBoost | TCN | AutoARIMA | Block GRU | Informer | TimesNet | N-BEATS | iTransformer | PatchTST |
|---|---|---|---|---|---|---|---|---|---|---|---|---|
| AI4I 2020 | 0.194 | 0.201 | 0.212 | 0.310 | 0.267 | 2.903 | 0.196 | 0.243 | 0.215 | 0.203 | 0.566 | 0.213 |
| Appliances Energy | 0.329 | 0.229 | 0.330 | 0.455 | 0.320 | 6.149 | 0.233 | 0.440 | 0.317 | 0.208 | 1.904 | 0.255 |
| Brent Oil Prices | 3.690 | 3.814 | 3.394 | 5.325 | 4.603 | 23.288 | 3.363 | 5.064 | 4.001 | 3.889 | 9.526 | 3.978 |
| ECL | 863.001 | 305.135 | 351.841 | 320.558 | 345.969 | 567.126 | 355.638 | 401.137 | 360.283 | 293.905 | 265.566 | 212.552 |
| ETTh1 | 1.696 | 1.400 | 1.378 | 1.978 | 1.831 | 5.352 | 1.472 | 1.928 | 1.528 | 1.399 | 2.056 | 1.393 |
| ETTm2 | 2.342 | 1.341 | 1.413 | 2.179 | 1.997 | 17.990 | 1.244 | 2.091 | 1.664 | 1.289 | 3.837 | 1.300 |
| Gas sensor dynamic gas mixtures | 152.654 | 92.563 | 78.502 | 140.716 | 122.790 | 918.897 | 68.305 | 235.723 | 204.728 | 121.172 | 391.901 | 115.554 |
| Gas sensor temperature modulation | 0.028 | 0.029 | 0.034 | 0.107 | 0.059 | 0.382 | 0.055 | 0.050 | 0.034 | 0.027 | 0.189 | 0.029 |
| ISO-NY | 121.001 | 99.867 | 101.535 | 137.813 | 141.718 | 457.983 | 90.797 | 141.026 | 120.184 | 97.131 | 158.093 | 66.949 |
| Monroe Water Treatment Plant | 2,122.758 | 1,464.403 | 1,412.601 | 1,558.190 | 1,525.345 | 1,822.697 | 1,391.341 | 1,914.327 | 1,836.850 | 1,488.703 | 1,582.345 | 1,492.688 |
| Seoul Bike Demand | 583.045 | 433.348 | 487.638 | 472.760 | 423.851 | 970.445 | 440.350 | 631.203 | 526.740 | 481.555 | 500.798 | 410.179 |
| Steel_industry_Usage_kWh | 24.560 | 126.737 | 35.388 | 21.559 | 23.491 | 38.889 | 18.154 | 73.772 | 72.891 | 150.706 | 20.535 | 16.071 |

Table 19. Scenario 2 Forecasting Horizon 96 RMSE Results

| Domain | Naive | MLP | LSTM | XGBoost | TCN | AutoARIMA | Block GRU | Informer | TimesNet | N-BEATS | iTransformer | PatchTST |
|---|---|---|---|---|---|---|---|---|---|---|---|---|
| AI4I 2020 | 0.469 | 0.607 | 0.519 | 1.228 | 0.751 | 2.903 | 0.570 | 0.817 | 0.901 | 0.501 | 0.792 | 0.516 |
| Appliances Energy | 1.972 | 2.167 | 7.270 | 3.831 | 2.600 | 6.391 | 1.605 | 5.036 | 2.814 | 1.937 | 4.198 | 1.546 |
| Brent Oil Prices | 8.410 | 11.226 | 60.420 | 15.644 | 14.537 | 23.288 | 18.735 | 17.383 | 16.990 | 14.930 | 18.999 | 11.292 |
| ECL | 815.399 | 318.963 | 502.879 | 298.199 | 421.945 | 566.444 | 327.617 | 548.614 | 514.534 | 346.098 | 314.594 | 252.878 |
| ETTh1 | 3.462 | 2.734 | 3.150 | 3.593 | 3.615 | 5.231 | 3.411 | 2.818 | 3.524 | 2.746 | 3.103 | 2.690 |
| ETTm2 | 8.141 | 3.703 | 4.128 | 5.160 | 4.764 | 22.765 | 4.452 | 5.771 | 4.402 | 3.693 | 4.735 | 3.718 |
| Gas sensor dynamic gas mixtures | 777.327 | 618.201 | 638.814 | 816.032 | 779.354 | 945.829 | 723.622 | 763.978 | 741.206 | 689.258 | 818.962 | 592.574 |
| Gas sensor temperature modulation | 0.099 | 0.124 | 0.192 | 0.412 | 0.375 | 0.383 | 0.139 | 0.171 | 0.138 | 0.094 | 0.279 | 0.090 |
| ISO-NY | 243.417 | 163.621 | 154.898 | 184.555 | 175.132 | 377.355 | 191.495 | 294.272 | 185.157 | 161.044 | 206.773 | 149.424 |
| Monroe Water Treatment Plant | 2,332.733 | 1,522.205 | 1,654.262 | 1,644.128 | 1,747.507 | 1,781.728 | 1,569.557 | 1,725.637 | 1,607.561 | 2,104.021 | 1,605.218 | 1,602.75 |
| Seoul Bike Demand | 711.159 | 525.487 | 1,290.694 | 555.845 | 534.104 | 970.949 | 633.164 | 620.562 | 532.384 | 518.224 | 531.257 | 518.278 |
| Steel_industry_Usage_kWh | 37.734 | 19.494 | 74.124 | 19.608 | 24.591 | 34.631 | 24.739 | 30.181 | 24.217 | 19.857 | 22.321 | 20.643 |

Table 20. Scenario 2 Forecasting Horizon 288 RMSE Results

| Domain | Naive | MLP | LSTM | XGBoost | TCN | AutoARIMA | Block GRU | Informer | TimesNet | N-BEATS | iTransformer | PatchTST |
|---|---|---|---|---|---|---|---|---|---|---|---|---|
| AI4I 2020 | 0.940 | 1.055 | 0.864 | 1.823 | 1.270 | 2.903 | 1.196 | 1.689 | 1.040 | 0.867 | 1.168 | 0.985 |
| Appliances Energy | 4.227 | 4.321 | 21.095 | 6.809 | 5.304 | 6.441 | 7.577 | 6.911 | 4.641 | 4.031 | 4.614 | 3.991 |
| Brent Oil Prices | 13.532 | 22.365 | 18.131 | 36.867 | 23.609 | 23.288 | 30.298 | 22.247 | 16.207 | 30.321 | 20.201 | 29.494 |
| ECL | 827.297 | 412.890 | 561.472 | 335.966 | 425.942 | 566.572 | 438.310 | 561.755 | 425.308 | 412.787 | 321.634 | 305.162 |
| ETTh1 | 3.621 | 3.203 | 5.280 | 3.388 | 3.143 | 5.224 | 4.354 | 3.365 | 3.188 | 3.092 | 2.908 | 3.116 |
| ETTm2 | 8.725 | 5.108 | 9.751 | 6.609 | 7.464 | 22.930 | 6.848 | 6.693 | 5.337 | 5.001 | 5.526 | 4.844 |
| Gas sensor dynamic gas mixtures | 1041.765 | 781.925 | 982.279 | 909.600 | 869.300 | 951.292 | 843.251 | 838.995 | 806.029 | 841.563 | 876.566 | 768.973 |
| Gas sensor temperature modulation | 0.222 | 0.506 | 3.027 | 0.921 | 0.549 | 0.383 | 0.480 | 0.307 | 0.234 | 0.238 | 0.336 | 0.192 |
| ISO-NY | 271.961 | 217.042 | 219.175 | 246.600 | 252.761 | 375.049 | 261.700 | 302.873 | 245.016 | 211.327 | 230.368 | 198.437 |
| Monroe Water Treatment Plant | 2,587.926 | 1,700.458 | 2,389.363 | 1,581.009 | 1,758.932 | 1,766.174 | 2,014.424 | 1,787.676 | 1,776.605 | 2,688.406 | 1,763.561 | 1,655.959 |
| Seoul Bike Demand | 741.047 | 540.182 | 959.856 | 604.089 | 608.096 | 970.943 | 618.853 | 647.468 | 554.374 | 543.700 | 544.353 | 546.879 |
| Steel_industry_Usage_kWh | 38.557 | 19.593 | 80.066 | 19.671 | 28.196 | 32.503 | 22.575 | 32.417 | 25.961 | 24.790 | 22.800 | 20.451 |

Table 21. Scenario 2 Forecasting Horizon 672 RMSE Results

| Domain | Naive | MLP | LSTM | XGBoost | TCN | AutoARIMA | Block GRU | Informer | TimesNet | N-BEATS | iTransformer | PatchTST |
|---|---|---|---|---|---|---|---|---|---|---|---|---|
| AI4I 2020 | 1.627 | 1.958 | 2.380 | 2.417 | 1.716 | 2.903 | 1.734 | 2.330 | 1.742 | 1.817 | 2.395 | 1.542 |
| Appliances Energy | 6.782 | 7.084 | 21.724 | 7.583 | 5.682 | 6.456 | 10.000 | 30.874 | 8.178 | 9.444 | 8.185 | 7.351 |
| Brent Oil Prices | 15.537 | 41.553 | 21.509 | 30.359 | 36.755 | 23.288 | 47.509 | 36.046 | 26.841 | 44.562 | 28.442 | 74.282 |
| ECL | 851.768 | 446.725 | 637.740 | 357.734 | 358.322 | 566..585 | 401.466 | 3.995 | 436.725 | 567.020 | 346.614 | 319.767 |
| ETTh1 | 3.396 | 3.465 | 8.006 | 5.252 | 3.372 | 5.223 | 5.056 | 7.316 | 3.696 | 3.575 | 3.347 | 3.697 |
| ETTm2 | 10.277 | 6.261 | 11.227 | 7.308 | 7.654 | 22.961 | 6.940 | 7.400 | 6.132 | 6.589 | 6.212 | 6.203 |
| Gas sensor dynamic gas mixtures | 1,042.539 | 765.925 | 976.746 | 986.233 | 825.483 | 952.258 | 997.934 | 883.878 | 811.900 | 808.237 | 940.570 | 787.522 |
| Gas sensor temperature modulation | 0.399 | 0.374 | 0.835 | 1.887 | 0.886 | 0.383 | 0.626 | 0.458 | 0.419 | 0.351 | 0.511 | 0.429 |
| ISO-NY | 279.378 | 227.200 | 294.348 | 275.987 | 278.362 | 374.572 | 270.472 | 322.780 | 222.767 | 252.818 | 240.914 | 209.443 |
| Monroe Water Treatment Plant | 1,874.743 | 1,893.164 | 2,385.271 | 1,758.885 | 1,942.684 | 1,763.438 | 1,701.306 | 2,138.523 | 1,607.877 | 2,069.791 | 1,449.539 | 1,599.981 |
| Seoul Bike Demand | 869.602 | 704.534 | 1393.425 | 615.069 | 672.575 | 970.949 | 668.959 | 778.634 | 667.653 | 654.478 | 644.807 | 709.500 |
| Steel_industry_Usage_kWh | 39.443 | 21.446 | 120.229 | 20.711 | 21.190 | 31.900 | 23.672 | 32.360 | 25.193 | 27.460 | 22.254 | 20.710 |

Table 22. Scenario 3 Forecasting Horizon 3 RMSE Results

| Domain | Naive | MLP | TCN | Block GRU | XGBoost | Dlinear | BiTCH | TSMixerX | TiDE | N-HITS | TFT |
|---|---|---|---|---|---|---|---|---|---|---|---|
| AI4I 2020 | 0.112 | 0.117 | 7.186 | 1.254 | 0.152 | 27.050 | 0.256 | 0.148 | 0.235 | 0.113 | 0.113 |
| Appliances Energy | 0.103 | 0.101 | 0.078 | 0.090 | 0.158 | 0.073 | 0.245 | 0.153 | 0.388 | 0.065 | 0.088 |
| ECL | 273.919 | 216.651 | 163.041 | 568.557 | 212.996 | 700,922.500 | 385.971 | 274.164 | 349.933 | 181.127 | 224.317 |
| Electricity | 0.780 | 0.575 | 1.531 | 0.584 | 0.614 | 3.178 | 0.689 | 0.623 | 0.697 | 0.564 | 0.641 |
| ETTh1 | 0.945 | 0.917 | 1.319 | 1.294 | 1.143 | 8.086 | 1.844 | 1.147 | 1.609 | 0.862 | 0.939 |
| ETTm2 | 0.739 | 0.792 | 1.081 | 0.405 | 0.697 | 3.001 | 2.169 | 0.953 | 2.323 | 0.454 | 0.844 |
| Gas sensor dynamic gas mixtures | 47.914 | 30.667 | 72.610 | 308.425 | 29.565 | 20,430.440 | 139.288 | 79.922 | 172.470 | 22.320 | 41.419 |
| Gas sensor temperature modulation | 0.013 | 0.014 | 0.703 | 0.044 | 0.053 | 2.158 | 0.032 | 0.021 | 0.030 | 0.013 | 0.013 |
| Monroe Water Treatment Plant | 1,844.642 | 1,335.430 | 1,9671.541 | 1,674.750 | 1,370.073 | 1,022,371.375 | 2,060.596 | 1,374.428 | 1,653.879 | 1,420.620 | 1,480.469 |
| Seoul Bike Demand | 485.709 | 382.498 | 568.698 | 606.197 | 307.231 | 18,673.560 | 538.860 | 416.933 | 557.836 | 326.413 | 434.350 |
| Steel_industry_Usage_kWh | 17.912 | 14.110 | 17.466 | 11.993 | 12.992 | 33.732 | 31.843 | 31.214 | 25.672 | 13.703 | 17.710 |

Table 23. Scenario 3 Forecasting Horizon 6 RMSE Results

| Domain | Naive | MLP | TCN | Block GRU | XGBoost | Dlinear | BiTCH | TSMixerX | TiDE | N-HITS | TFT |
|---|---|---|---|---|---|---|---|---|---|---|---|
| AI4I 2020 | 0.145 | 0.154 | 0.304 | 1.255 | 0.218 | 10.233 | 0.575 | 0.175 | 0.228 | 0.146 | 0.147 |
| Appliances Energy | 0.187 | 0.160 | 0.164 | 1.673 | 0.288 | 0.157 | 2.197 | 0.192 | 0.394 | 0.133 | 0.151 |
| ECL | 470.433 | 228.405 | 977,219.900 | 569.427 | 236.063 | 1,027,736.000 | 477.374 | 325.616 | 355.616 | 235.851 | 301.413 |
| Electricity | 0.981 | 0.595 | 0.642 | 0.586 | 0.636 | 4.040 | 0.621 | 0.651 | 0.704 | 0.594 | 0.694 |
| ETTh1 | 1.190 | 1.124 | 1.473 | 1.326 | 1.563 | 7.051 | 2.677 | 1.298 | 1.6716 | 1.0633 | 1.169 |
| ETTm2 | 1.285 | 1.110 | 1.948 | 1.018 | 1.227 | 20.992 | 5.639 | 1.311 | 2.613 | 0.764 | 1.289 |
| Gas sensor dynamic gas mixtures | 83.623 | 52.291 | 139.729 | 219.292 | 71.989 | 54,033.37 | 352.0374 | 144.283 | 192.945 | 38.473 | 87.481 |
| Gas sensor temperature modulation | 0.017 | 0.029 | 0.089 | 0.244 | 0.069 | 1.427 | 0.063 | 0.0241 | 0.035 | 0.019 | 0.019 |
| Monroe Water Treatment Plant | 1,730.997 | 1,430.234 | 2,460.161 | 1,721.519 | 1,419.084 | 268,052.813 | 1,516.980 | 1,461.190 | 1,661.926 | 1,421.607 | 1,448.572 |
| Seoul Bike Demand | 568.694 | 394.389 | 5,311.956 | 587.302 | 361.742 | 3,920.023 | 550.464 | 472.898 | 582.788 | 375.142 | 514.572 |
| Steel_industry_Usage_kWh | 22.403 | 17.230 | 20.756 | 14.152 | 17.626 | 48.872 | 21.508 | 31.222 | 35.470 | 17.218 | 20.402 |

Table 24. Scenario 3 Forecasting Horizon 12 RMSE Results

| Domain | Naive | MLP | TCN | Block GRU | XGBoost | Dlinear | BiTCH | TSMixerX | TiDE | N-HITS | TFT |
|---|---|---|---|---|---|---|---|---|---|---|---|
| AI4I 2020 | 0.194 | 0.200 | 11.348 | 1.372 | 0.328 | 35.805 | 0.623 | 0.216 | 0.267 | 0.203 | 0.199 |
| Appliances Energy | 0.329 | 0.253 | 0.361 | 0.853 | 0.491 | 0.274 | 1.056 | 0.296 | 0.452 | 0.252 | 0.263 |
| ECL | 863.001 | 273.288 | 13,949.810 | 574.188 | 259.560 | 2,937,608.000 | 444.807 | 336.727 | 439.046 | 279.109 | 0.263 |
| Electricity | 0.925 | 0.611 | 1.316 | 0.618 | 0.645 | 3.663 | 0.650 | 0.663 | 0.710 | 0.621 | 371.196 |
| ETTh1 | 1.697 | 1.436 | 1.936 | 1.518 | 2.054 | 3.565 | 2.407 | 1.514 | 1.819 | 1.379 | 1.563 |
| ETTm2 | 2.342 | 1.636 | 3.753 | 1.307 | 2.327 | 12.445 | 3.457 | 1.909 | 3.245 | 1.465 | 1.740 |
| Gas sensor dynamic gas mixtures | 152.654 | 154.159 | 233.497 | 562.820 | 174.615 | 74,141.680 | 265.127 | 218.636 | 260.168 | 85.064 | 137.194 |
| Gas sensor temperature modulation | 0.028 | 0.035 | 0.135 | 0.038 | 0.101 | 3.600 | 0.120 | 0.032 | 0.043 | 0.030 | 0.033 |
| Monroe Water Treatment Plant | 2,122.758 | 1,474.392 | 76,759.039 | 1,702.053 | ,555.267 | 1,397,646.125 | 1,547.804 | 1,526.968 | 1,717.179 | 1,547.918 | 1,542.667 |
| Seoul Bike Demand | 583.045 | 454.260 | 787.195 | 555.717 | 427.629 | 13,746.920 | 502.941 | 522.496 | 562.673 | 446.497 | 592.634 |
| Steel_industry_Usage_kWh | 24.560 | 21.037 | 25.386 | 16.156 | 20.191 | 138.079 | 18.459 | 30.007 | 53.901 | 21.899 | 24.937 |

Table 25. Scenario 4 Forecasting Horizon 96 RMSE Results

| Domain | Naive | MLP | TCN | Block GRU | XGBoost | Dlinear | BiTCH | TSMixerX | TiDE | N-HITS | TFT |
|---|---|---|---|---|---|---|---|---|---|---|---|
| AI4I 2020 | 0.469 | 0.648 | 5.839 | 1.439 | 1.240 | 525.241 | 0.666 | 0.637 | 0.669 | 0.457 | 0.481 |
| Appliances Energy | 1.972 | 3.096 | 2.503 | 1.667 | 3.935 | 216.604 | 3.076 | 2.396 | 2.376 | 2.070 | 1.550 |
| ECL | 815.399 | 340.465 | 845,342.900 | 627.597 | 277.136 | 43,977,316.000 | 565.115 | 268.834 | 311.583 | 397.297 | 557.537 |
| Electricity | 0.976 | 0.632 | 0.719 | 0.704 | 0.671 | 45.879 | 3.266 | 0.632 | 0.666 | 0.816 | 0.732 |
| ETTh1 | 3.462 | 2.615 | 3.579 | 3.897 | 3.679 | 80.203 | 5.945 | 2.841 | 2.765 | 3.030 | 3.104 |
| ETTm2 | 8.141 | 4.330 | 6.128 | 7.371 | 4.950 | 293.427 | 744.365 | 3.877 | 3.924 | 4.684 | 5.220 |
| Gas sensor dynamic gas mixtures | 777.327 | 706.707 | 2,161.759 | 710.758 | 898.503 | 1,385,569.000 | 0.149 | 742.443 | 678.862 | 613.875 | 884.067 |
| Gas sensor temperature modulation | 0.099 | 0.146 | 0.374 | 0.607 | 0.357 | 35.254 | 2,575.223 | 0.123 | 0.124 | 0.101 | 0.101 |
| Monroe Water Treatment Plant | 2,332.733 | 1,798.083 | 23,858.262 | 1,773.311 | 1,509.838 | 17,871,648.000 | 2,613.757 | 1,630.354 | 1,650.897 | 2,207.930 | 2,259.522 |
| Seoul Bike Demand | 711.159 | 638.428 | 716.765 | 695.497 | 530.324 | 358,149.600 | 628.433 | 504.596 | 548.857 | 615.316 | 656.925 |
| Steel_industry_Usage_kWh | 37.734 | 21.173 | 24.936 | 25.846 | 20.419 | 1284.105 | 28.244 | 21.410 | 21.123 | 21.143 | 36.155 |

Table 26. Scenario 4 Forecasting Horizon 288 RMSE Results

| Domain | Naive | MLP | TCN | Block GRU | XGBoost | Dlinear | BiTCH | TSMixerX | TiDE | N-HITS | TFT |
|---|---|---|---|---|---|---|---|---|---|---|---|
| AI4I 2020 | 0.940 | 1.170 | 3.770 | 1.362 | 1.691 | 469.110 | 1.082 | 1.040 | 1.130 | 0.806 | 0.952 |
| Appliances Energy | 4.227 | 4.050 | 5.478 | 7.819 | 5.408 | 222.044 | 4.943 | 5.906 | 4.866 | 4.851 | 4.392 |
| ECL | 827.297 | 379.738 | 3,093.021 | 618.896 | 297.537 | 104,000,000.000 | 560.620 | 309.501 | 330.068 | 355.739 | 573.416 |
| Electricity | 1.070 | 0.650 | 0.942 | 0.721 | 0.691 | 41.953 | 3.558 | 0.642 | 0.678 | 0.852 | 0.788 |
| ETTh1 | 3.621 | 2.861 | 3.061 | 4.096 | 3.493 | 120.172 | 6.574 | 2.856 | 3.154 | 3.200 | 3.288 |
| ETTm2 | 8.725 | 5.547 | 11.575 | 6.484 | 6.696 | 196.049 | 820.747 | 5.257 | 5.193 | 5.598 | 6.841 |
| Gas sensor dynamic gas mixtures | 1,041.765 | 792.706 | 2,374.841 | 931.829 | 862.483 | 1,630,580.000 | 0.254 | 872.426 | 800.094 | 986.251 | 1,379.796 |
| Gas sensor temperature modulation | 0.222 | 0.288 | 0.380 | 0.425 | 0.840 | 44.914 | 2,573.926 | 0.340 | 0.249 | 0.234 | 0.219 |
| Monroe Water Treatment Plant | 2,587.926 | 2,424.051 | 723,298.688 | 1,773.311 | 1,550.615 | 26,261,786.000 | 3285.364 | 1,634.634 | 1,710.275 | 3,079.217 | 1,801.941 |
| Seoul Bike Demand | 741.047 | 804.193 | 6,987.228 | 647.109 | 558.022 | 271,398.800 | 632.855 | 562.4708 | 571.832 | 638.706 | 654.370 |
| Steel_industry_Usage_kWh | 38.557 | 21.937 | 25.743 | 27.903 | 19.883 | 1,334.863 | 31.966 | 20.847 | 21.050 | 24.027 | 31.811 |

Table 27. Scenario 4 Forecasting Horizon 672 RMSE Results

| Domain | Naive | MLP | TCN | Block GRU | XGBoost | Dlinear | BiTCH | TSMixerX | TiDE | N-HITS | TFT |
|---|---|---|---|---|---|---|---|---|---|---|---|
| AI4I 2020 | 1.627 | 1.572 | 19.080 | 1.492 | 1.842 | 544.723 | 1.684 | 2.085 | 1.974 | 1.783 | OOM |
| Appliances Energy | 6.782 | 8.087 | 5.965 | 6.457 | 7.328 | 204.895 | 7.378 | 8.118 | 7.241 | 6.357 | OOM |
| ECL | 851.768 | 407.866 | 1,780.550 | 605.931 | 323.801 | 127,000,000.000 | 602.381 | 326.592 | 355.060 | 469.620 | OOM |
| Electricity | 0.922 | 0.633 | 1.015 | 0.694 | 0.729 | 50.069 | 4.439 | 0.648 | 0.681 | 0.746 | OOM |
| ETTh1 | 3.396 | 3.356 | 3.192 | 4.096 | 5.140 | 73.018 | 8.307 | 3.137 | 3.633 | 3.228 | OOM |
| ETTm2 | 10.277 | 7.139 | 11.714 | 7.502 | 7.016 | 279.536 | 891.682 | 6.461 | 6.223 | 6.575 | OOM |
| Gas sensor dynamic gas mixtures | 1,042.539 | 840.229 | 9,505.217 | 964.469 | 938.223 | 2,496,048.000 | 0.422 | 912.202 | 813.262 | 859.778 | OOM |
| Gas sensor temperature modulation | 0.399 | 0.497 | 0.736 | 1.210 | 1.861 | 45.938 | 3,482.110 | 0.523 | 0.448 | 0.450 | OOM |
| Monroe Water Treatment Plant | 1,874.743 | 2,639.862 | 6,935.542 | 1,815.089 | 1,588.512 | 24,332,164.000 | 3,309.435 | 1,605.669 | 1,682.614 | 2,092.845 | OOM |
| Seoul Bike Demand | 869.602 | 1,130.535 | 5,431.269 | 695.121 | 548.284 | 340,124.800 | 698.299 | 937.987 | 716.738 | 952.726 | OOM |
| Steel_industry_Usage_kWh | 39.443 | 22.234 | 23.618 | 25.157 | 20.684 | 1,817.259 | 31.763 | 20.749 | 21.315 | 25.734 | OOM |

# Appendix C

In Appendix C, the detailed results for different forecasting horizons are presented in three formats and different evaluation metrics:

## Exp Scenario 1: Short-term Univariate TSF

Table 28. Scenario one Performance summary of 12 algorithms on 12 datasets for FH=3

| Metric | Algorithms | | | | | | | | | | | |
|---|---|---|---|---|---|---|---|---|---|---|---|---|
| | Naive | MLP | LSTM | XGBoost | TCN | AutoARIMA | Block GRU | Informer | TimesNet | N-BEATS | iTransformer | PatchTST |
| AVG ERR | 0.0852 | 0.0846 | 0.0644 | 0.0705 | 0.0715 | 0.3528 | **0.0595** | 0.1398 | 0.1281 | 0.0709 | 0.1410 | 0.0851 |
| WIN | 3 | 0 | 2 | 0 | 0 | 0 | 3 | 0 | 0 | **4** | 0 | 0 |
| AVG Rank | 5.70 | 4.87 | 4.12 | 7.17 | 5.54 | 11.92 | 2.87 | 10.17 | 8.83 | **2.75** | 10.25 | 3.79 |

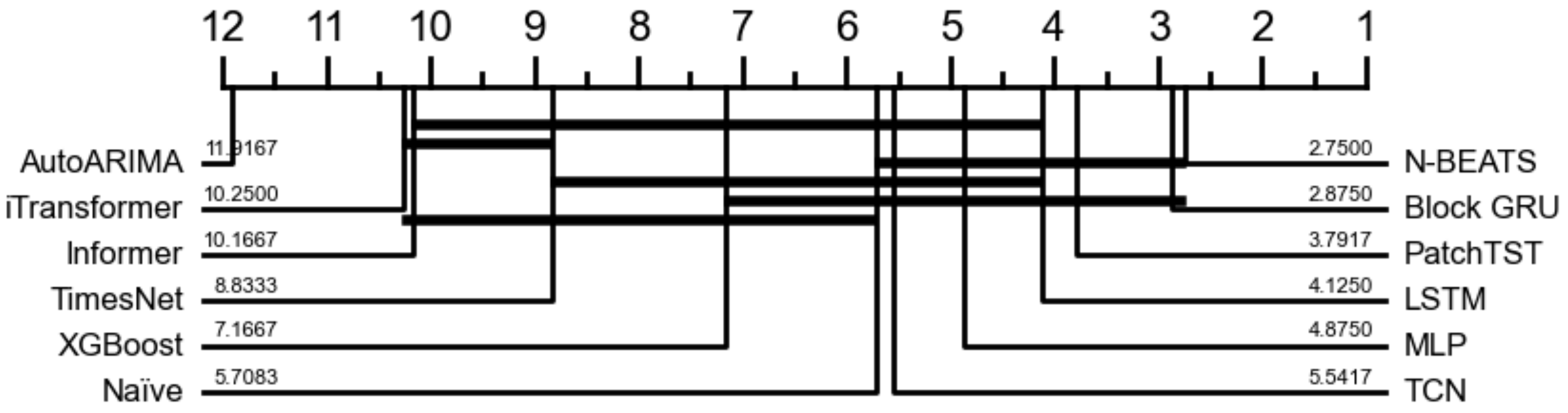

Figure 20: Critical difference diagrams for 12 univariate TSF algorithms on the 12 datasets for FH=3

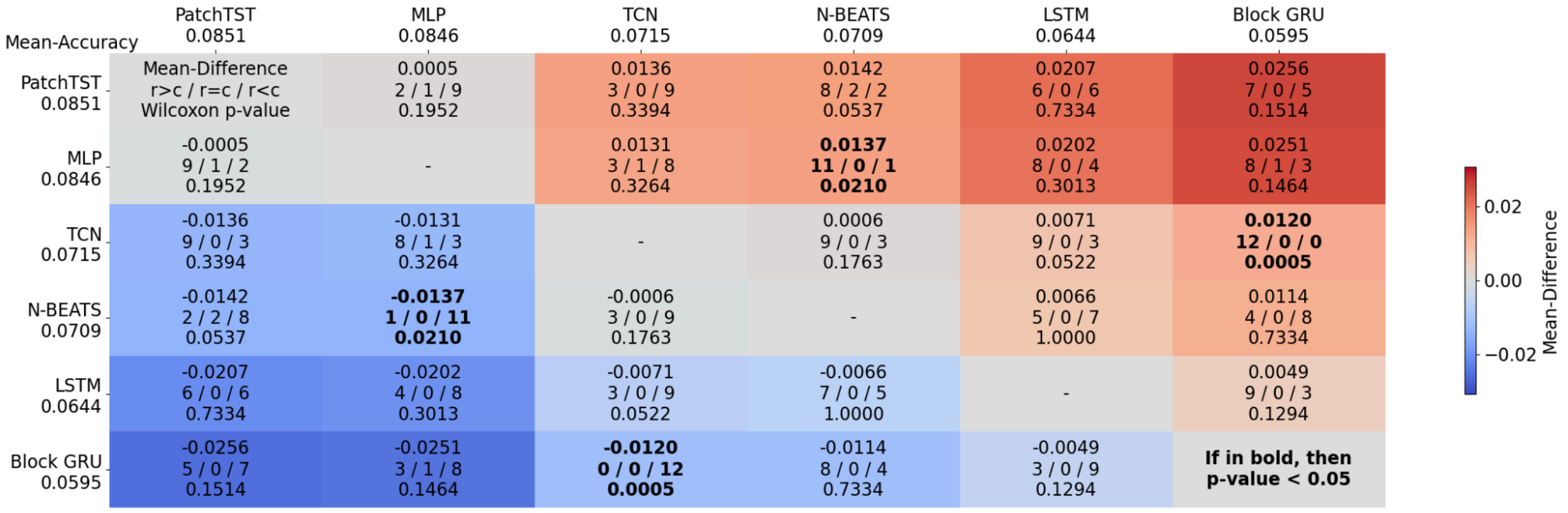

Figure 21: MCM for top six univariate TSF algorithms on the 12 datasets for FH=3

The results for scenario one forecasting horizon of 6 are:

Table 29. Scenario one Performance summary of 12 algorithms on 12 datasets for FH=6

| Metric | Algorithms | | | | | | | | | | | |
|---|---|---|---|---|---|---|---|---|---|---|---|---|
| | Naive | MLP | LSTM | XGBoost | TCN | AutoARIMA | Block GRU | Informer | TimesNet | N-BEATS | iTransformer | PatchTST |
| AVG ERR | 0.1158 | 0.1054 | 0.0995 | 0.0959 | 0.1002 | 0.3552 | **0.0782** | 0.1843 | 0.1573 | 0.0996 | 0.1494 | 0.0815 |
| WIN | **3** | 0 | 1 | 0 | 0 | 0 | **3** | 0 | 0 | 2 | 0 | **3** |
| AVG Rank | 5.92 | 3.87 | 3.96 | 7.50 | 6.29 | 11.96 | 3.21 | 9.75 | 8.62 | **3.08** | 9.92 | 3.92 |

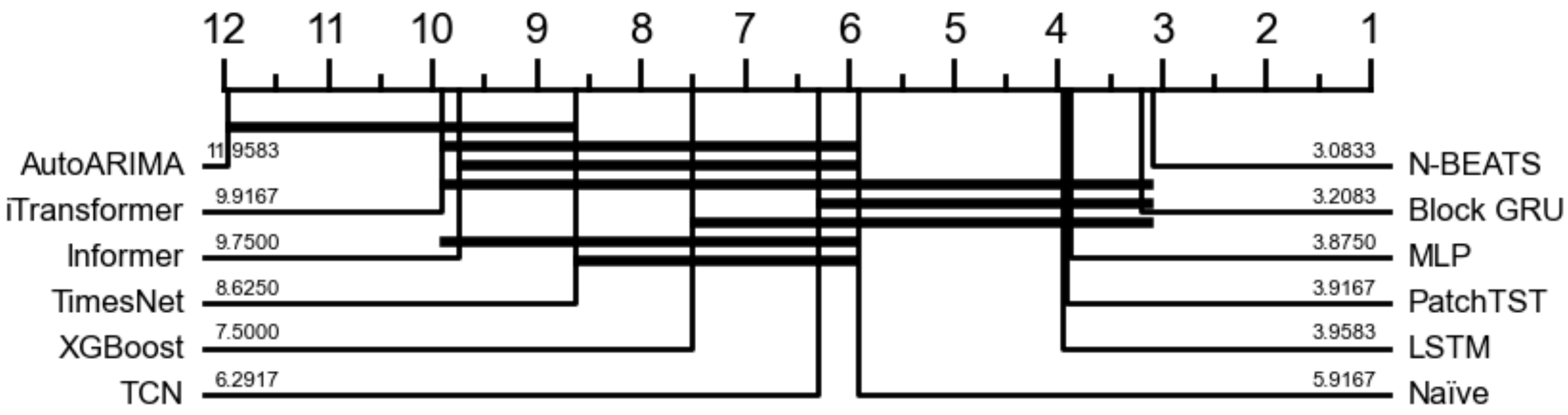

Figure 22: Critical difference diagrams for 12 univariate TSF algorithms on the 12 datasets for FH=6

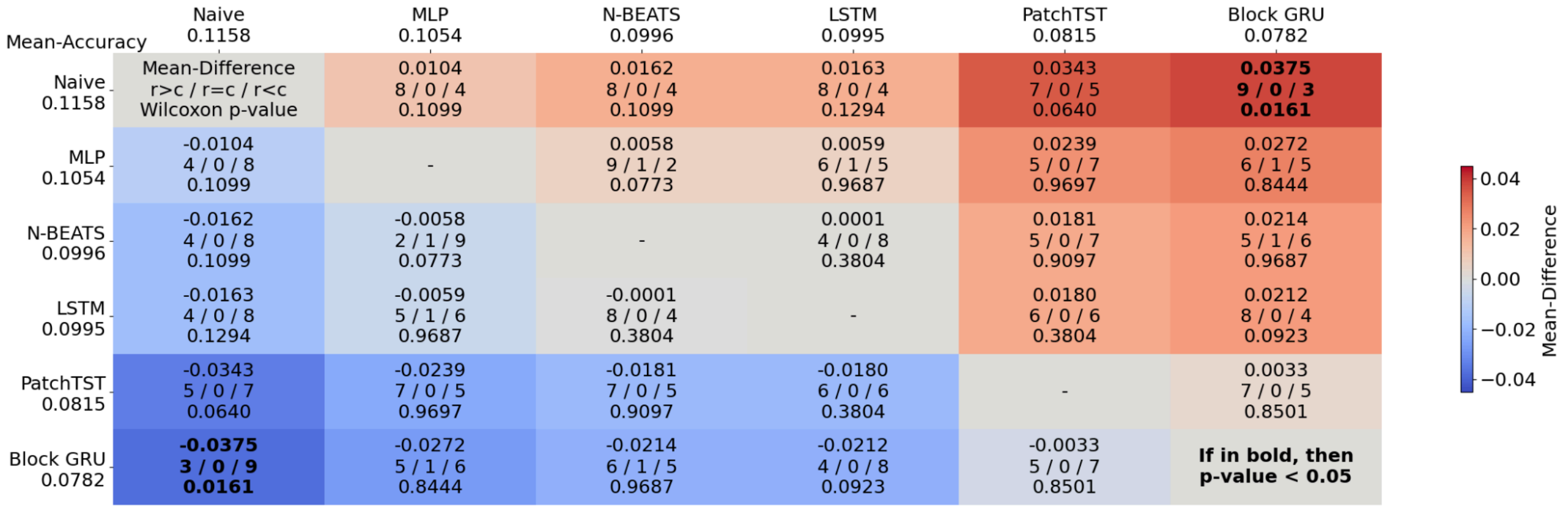

Figure 23: MCM for top six univariate TSF algorithms on the 12 datasets for FH=6

The results for scenario one forecasting horizon of 12 are:

Table 30. Scenario one Performance summary of 12 algorithms on 12 datasets for FH=12

| Metric | Algorithms | | | | | | | | | | | |
|---|---|---|---|---|---|---|---|---|---|---|---|---|
| | Naive | MLP | LSTM | XGBoost | TCN | AutoARIMA | Block GRU | Informer | Times Net | N-BEATS | iTransformer | PatchTST |
| AVG ERR | 0.1434 | 0.1488 | 0.1476 | 0.1284 | 0.1299 | 0.3595 | 0.1025 | 0.1890 | 0.1670 | 0.1468 | 0.1515 | **0.0983** |
| WIN | 2 | 0 | 0 | 0 | 0 | 0 | **4** | 0 | 0 | 3 | 0 | 3 |
| AVG Rank | 6.5 | 4.04 | 4.58 | 8.08 | 7.33 | 11.75 | **3.00** | 9.42 | 7.46 | 3.21 | 9.08 | 3.54 |

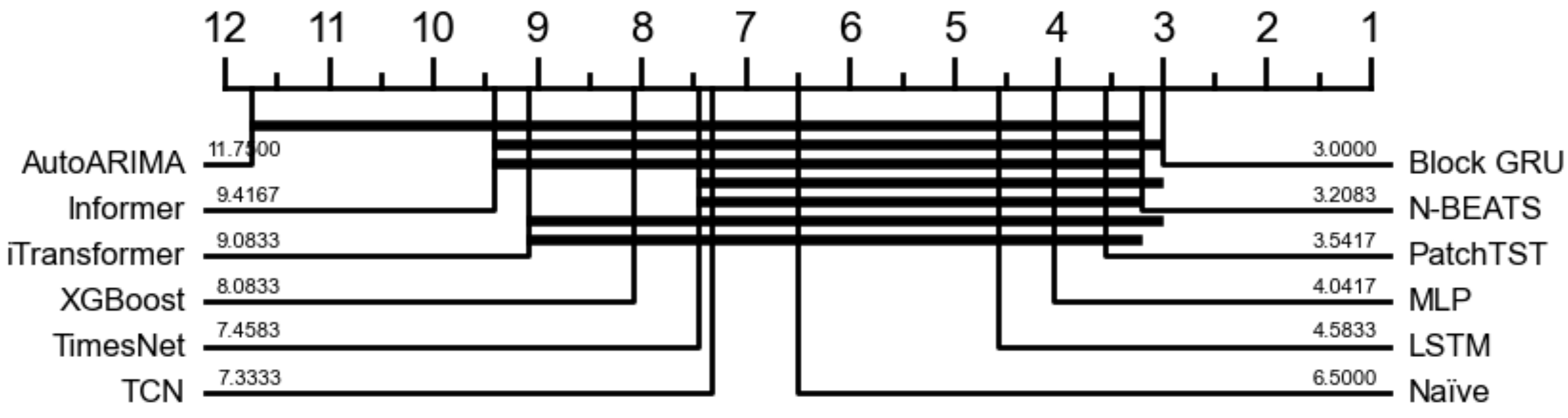

Figure 24: Critical difference diagrams for 12 univariate TSF algorithms on the 12 datasets for FH=12

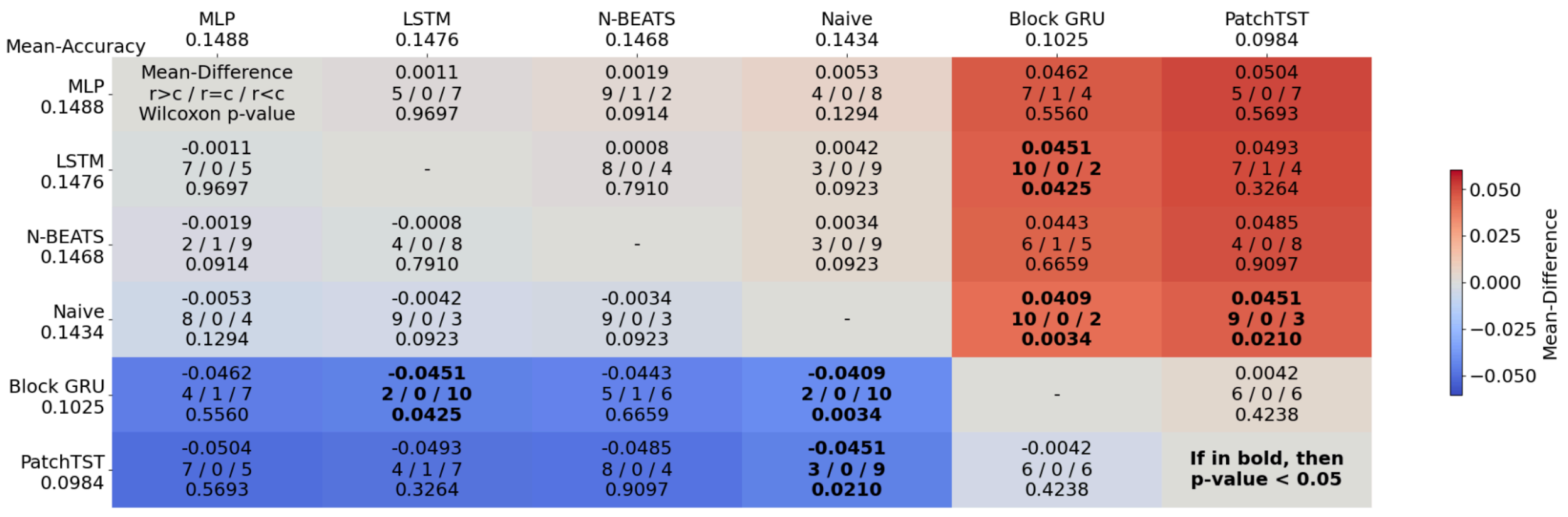

Figure 25: MCM for top six univariate TSF algorithms on the 12 datasets for FH=12

# Exp Scenario 2: Long-term Univariate TSF

The results for scenario two forecasting horizon of 96 are:

Table 31. Scenario two Performance summary of 12 algorithms on 12 datasets for FH=96

| Metric | Algorithms | | | | | | | | | | | |
|---|---|---|---|---|---|---|---|---|---|---|---|---|
| | Naive | MLP | LSTM | XGBoost | TCN | AutoARIMA | Block GRU | Informer | TimesNet | N-BEATS | iTransformer | PatchTST |
| AVG ERR | 0.2467 | 0.1622 | 0.3070 | 0.1847 | 0.2010 | 0.3715 | 0.1998 | 0.2524 | 0.2075 | 0.1709 | 0.2024 | **0.1610** |
| WIN | 2 | 1 | 1 | 1 | 0 | 0 | 0 | 0 | 0 | 1 | 0 | **6** |
| AVG Rank | 7.12 | 3.33 | 6.92 | 7.25 | 7.08 | 11.5 | 5.33 | 8.83 | 7.25 | 4.04 | 7.5 | **1.83** |

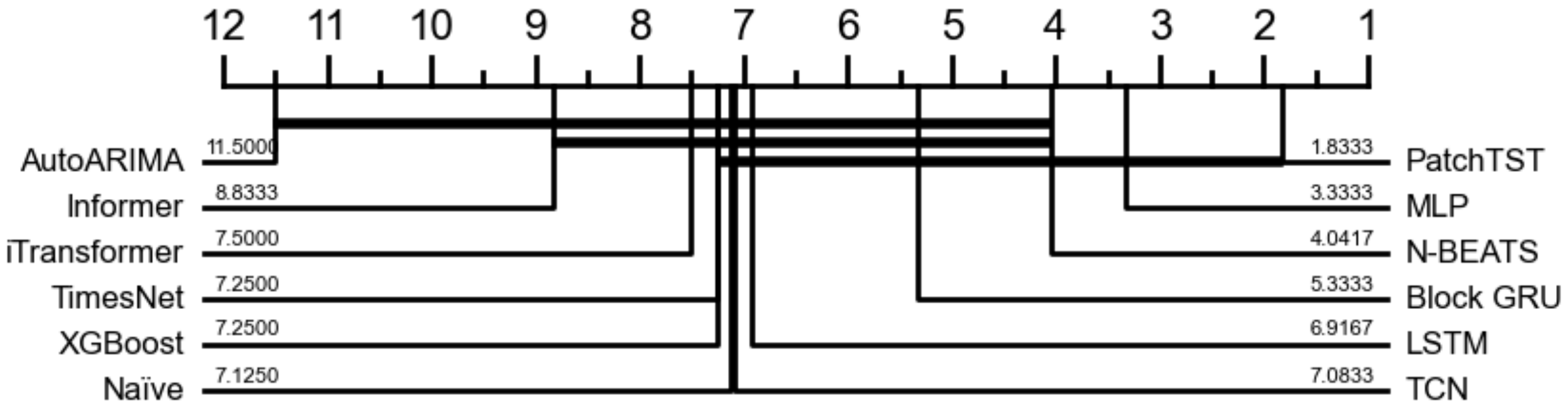

Figure 26: Critical difference diagrams for 12 univariate TSF algorithms on the 12 datasets for FH=96

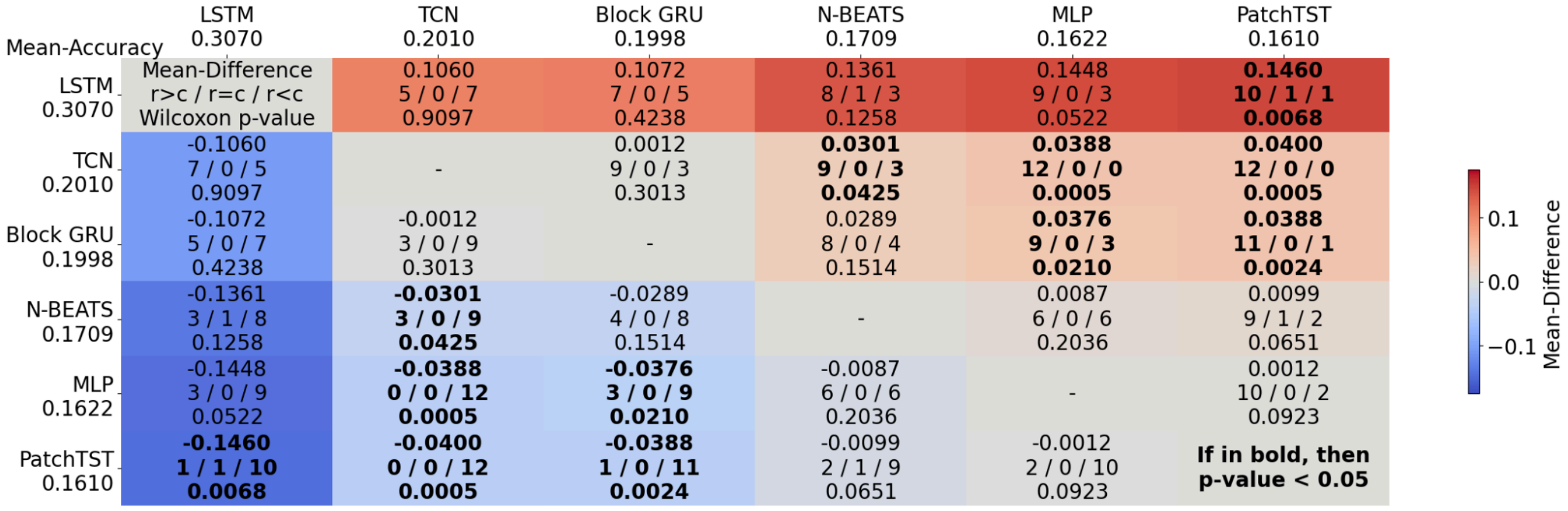

Figure 27: MCM for top six univariate TSF algorithms on the 12 datasets for FH=96

The results for scenario two forecasting horizon of 288 are:

Table 32. Scenario two Performance summary of 12 algorithms on 12 datasets for FH=288

| Metric | Algorithms | | | | | | | | | | | |
|---|---|---|---|---|---|---|---|---|---|---|---|---|
| | Naive | MLP | LSTM | XGBoost | TCN | AutoARIMA | Block GRU | Informer | Times Net | N-BEATS | iTransformer | PatchTST |
| AVG ERR | 0.2729 | 0.2004 | 0.3156 | 0.2286 | 0.2436 | 0.3718 | 0.2400 | 0.2754 | 0.2198 | 0.2251 | 0.2136 | **0.1985** |
| WIN | 1 | 0 | 1 | 2 | 0 | 0 | 1 | 0 | 0 | 0 | 1 | **6** |
| AVG Rank | 7.46 | 4.00 | 8.33 | 7.08 | 7.62 | 10.21 | 7.62 | 8.33 | 5.33 | 4.50 | 5.08 | **2.42** |

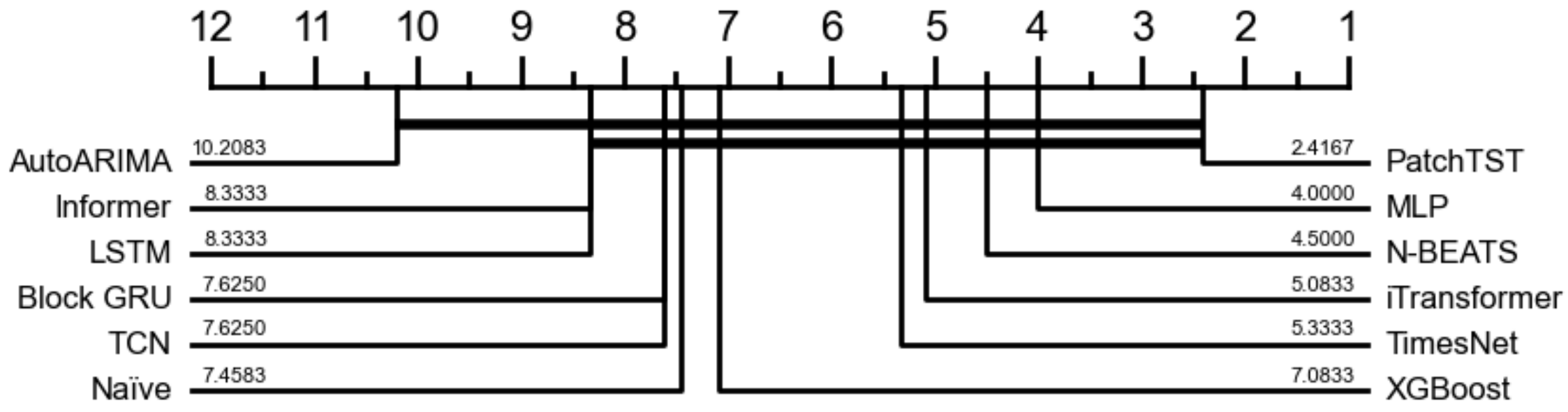

Figure 28: Critical difference diagrams for 12 univariate TSF algorithms on the 12 datasets for FH=288

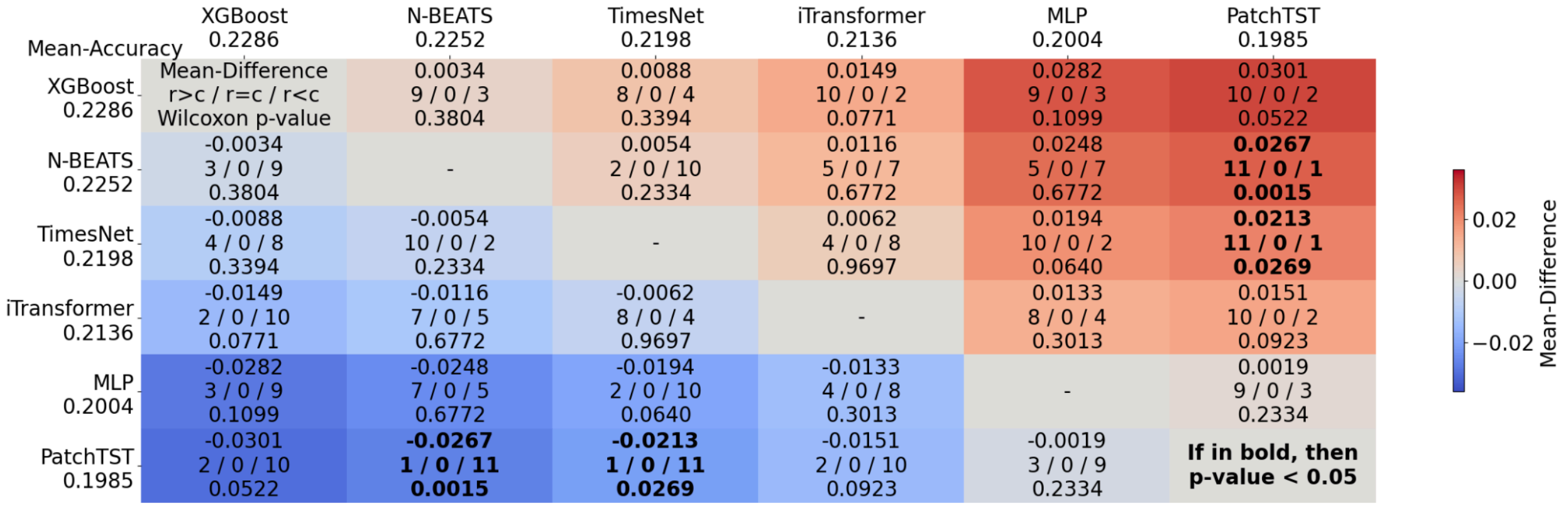

Figure 29: MCM for top six univariate TSF algorithms on the 12 datasets for FH=288

The results for scenario two forecasting horizon of 672 are:

Table 33. Scenario two Performance summary of 12 algorithms on 12 datasets for FH=672

| Metric | Algorithms | | | | | | | | | | | |
|---|---|---|---|---|---|---|---|---|---|---|---|---|
| | Naive | MLP | LSTM | XGBoost | TCN | AutoARIMA | Block GRU | Informer | Times Net | N-BEATS | iTransformer | PatchTST |
| AVG ERR | 0.2907 | 0.2479 | 0.6895 | 0.2528 | 0.2434 | 0.3799 | 0.2765 | 0.3593 | 0.2451 | 0.2690 | **0.2401** | 0.2615 |
| WIN | 2 | 1 | 0 | 2 | 1 | 0 | 0 | 1 | 0 | 1 | 1 | **3** |
| AVG Rank | 6.16 | 5.29 | 9.66 | 6.58 | 5.83 | 8.92 | 7.25 | 8.87 | 4.50 | 5.75 | 5.50 | **3.67** |

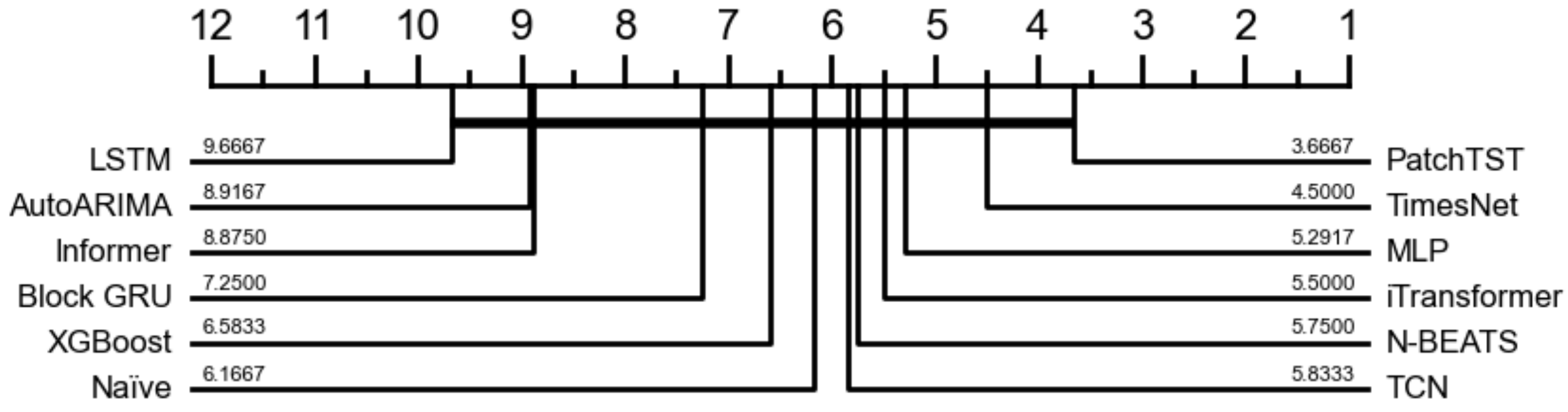

Figure 30: Critical difference diagrams for 12 univariate TSF algorithms on the 12 datasets for FH=672

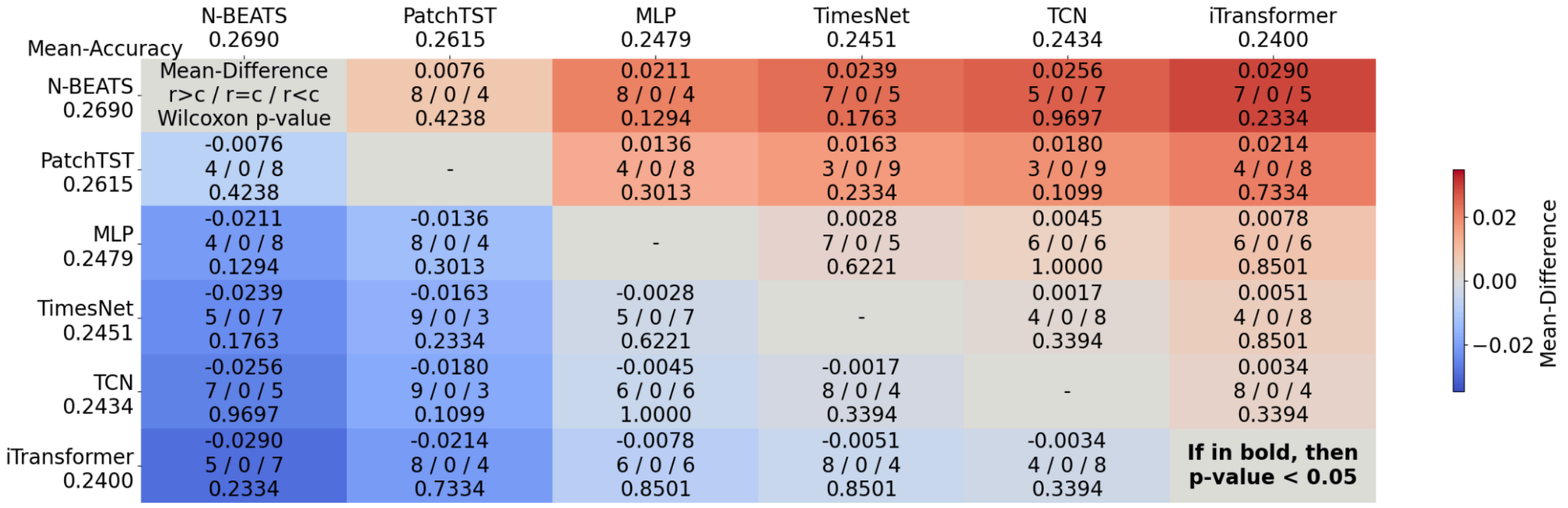

Figure 31: MCM for top six univariate TSF algorithms on the 12 datasets for FH=672

## Exp Scenario 3: Short-term Multivariate TSF

The results for scenario three forecasting horizon of 3 are:

Table 34. Scenario three Performance summary of 11 algorithms on 11 datasets for FH=3

| Metric | Algorithms | | | | | | | | | | |
|---|---|---|---|---|---|---|---|---|---|---|---|
| | Naive | MLP | TCN | Block GRU | XGBoost | Dlinear | BiTCN | TSMixerX | TiDE | N-HITS | TFT |
| AVG ERR | 0.1375 | 0.1210 | 0.3360 | 0.1611 | 0.1141 | 0.6400 | 0.2207 | 0.1799 | 0.2022 | **0.1091** | 0.1406 |
| WIN | 2 | 1 | 0 | 2 | 1 | 0 | 0 | 0 | 0 | **5** | 0 |
| AVG Rank | 4.32 | 3.59 | 8.5 | 6.41 | 4.04 | 10.14 | 8.64 | 5.95 | 8.18 | **1.86** | **4.36** |

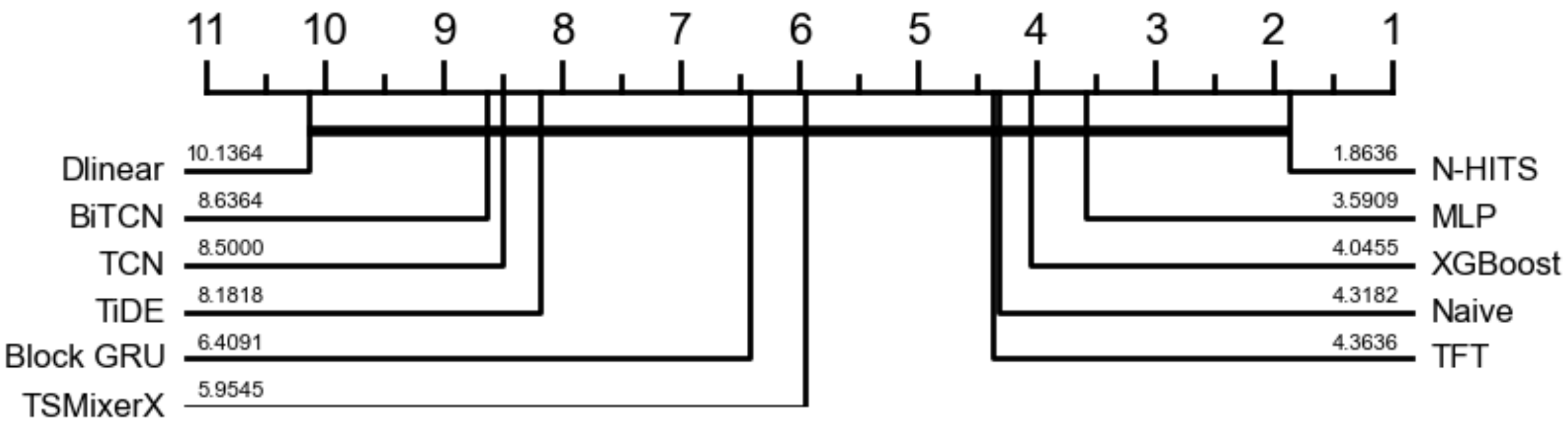

Figure 32: Critical difference diagrams for 11 multivariate TSF algorithms on the 11 datasets for FH=3

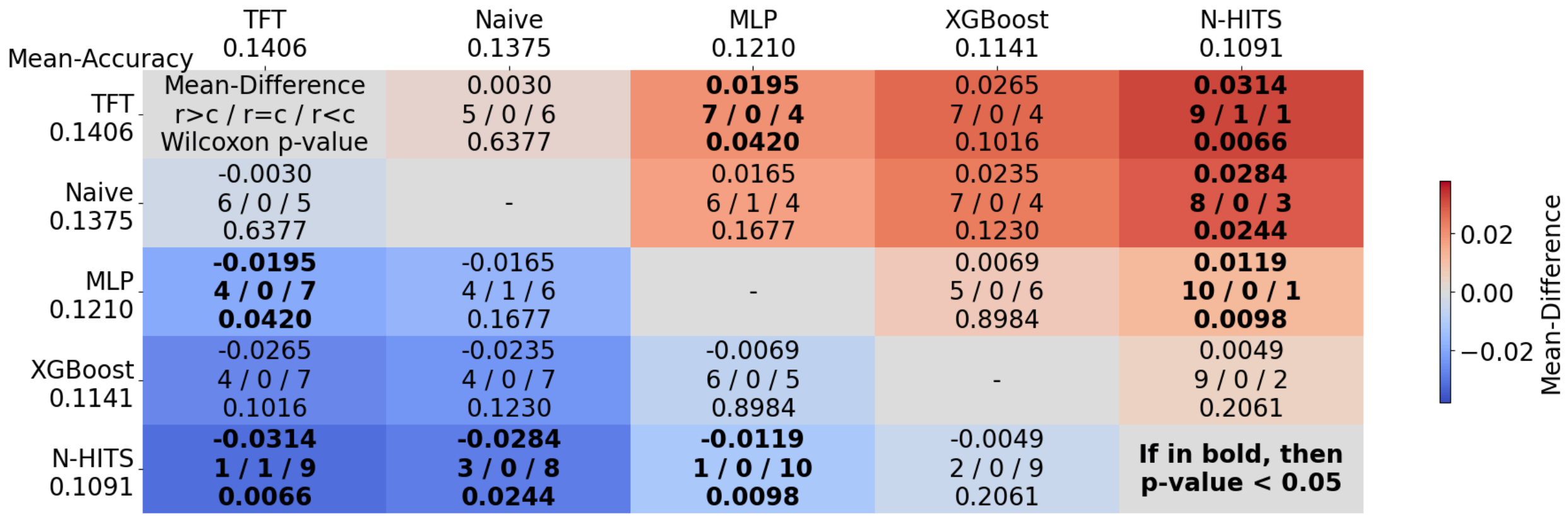

Figure 33: MCM for top five multivariate TSF algorithms on the 11 datasets for FH=3

The results for scenario three forecasting horizon of 6 are:

Table 35. Scenario three Performance summary of 11 algorithms on 11 datasets for FH=6

| Metric | Algorithms | | | | | | | | | | |
|---|---|---|---|---|---|---|---|---|---|---|---|
| | Naive | MLP | TCN | Block GRU | XGBoo st | Dlinear | BiTCN | TSMix erX | TiDE | N-HIT S | TFT |
| AVG ERR | 0.1846 | 0.1357 | 0.3064 | 0.1697 | 0.1387 | 0.6708 | 0.2148 | 0.1941 | 0.2330 | **0.1300** | 0.1707 |
| WIN | 2 | 2 | 0 | 1 | 1 | 0 | 0 | 0 | 0 | **5** | 0 |
| AVG Rank | 4.95 | 2.91 | 8.09 | 7.14 | 5.00 | 10.14 | 8.10 | 5.73 | 7.91 | **1.77** | **4.27** |

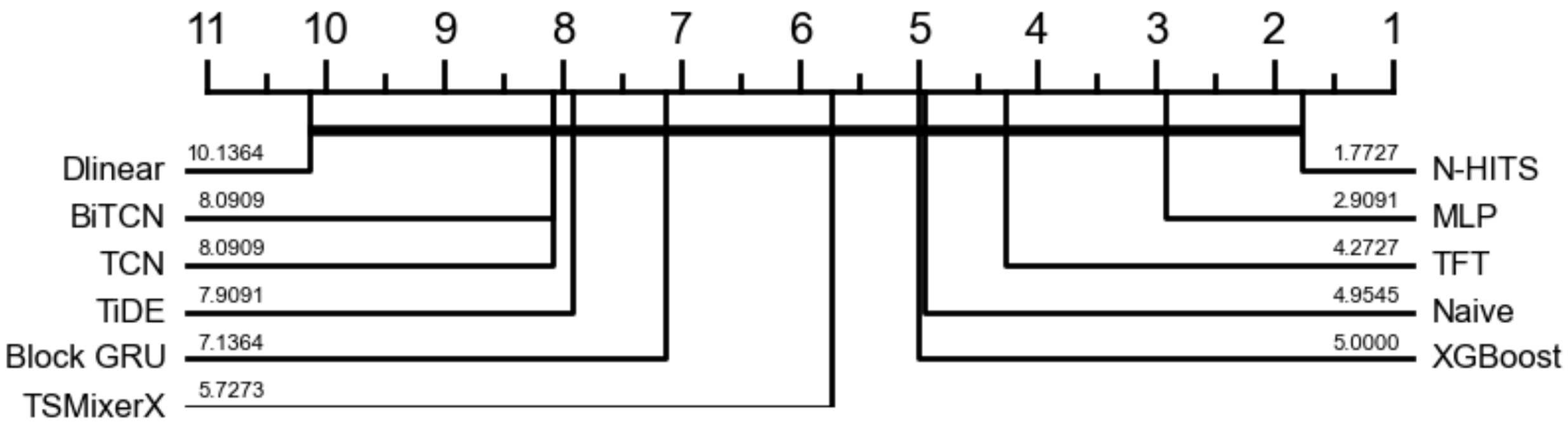

Figure 34: Critical difference diagrams for 11 multivariate TSF algorithms on the 11 datasets for FH=6

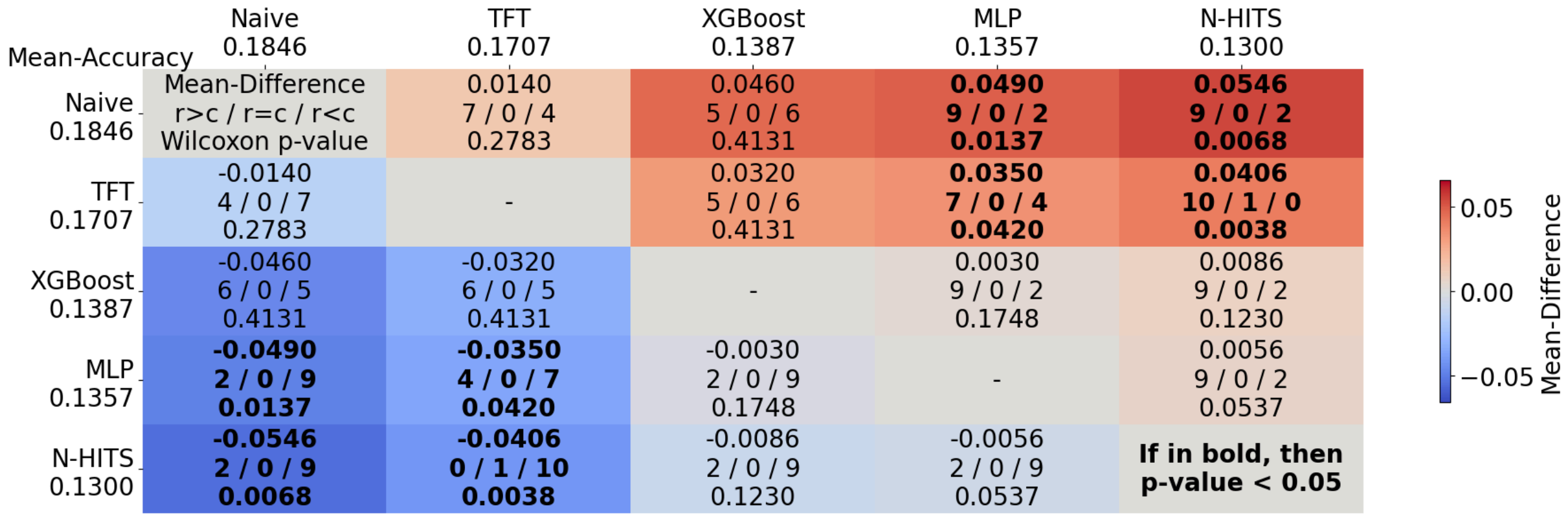

Figure 35: MCM for top five multivariate TSF algorithms on the 11 datasets for FH=6

The results for scenario three forecasting horizon of 12 are:

Table 36. Scenario three Performance summary of 11 algorithms on 11 datasets for FH=12

| Metric | Algorithms | | | | | | | | | | |
|---|---|---|---|---|---|---|---|---|---|---|---|
| | Naive | MLP | TCN | Block GRU | XGBoost | Dlinear | BiTCN | TSMixerX | TiDE | N-HITS | TFT |
| AVG ERR | 0.2080 | 0.1599 | 0.4610 | 0.1809 | 0.1655 | 0.6315 | 0.1909 | 0.2052 | 0.2381 | 0.1579 | **0.1500** |
| WIN | 1 | 0 | 0 | 2 | 1 | 0 | 0 | 0 | 0 | **3** | 2 |
| AVG Rank | 5.45 | 3.18 | 9.32 | 5.91 | 5.45 | 10.09 | 7.45 | 5.18 | 7.5 | **2.41** | **3.95** |

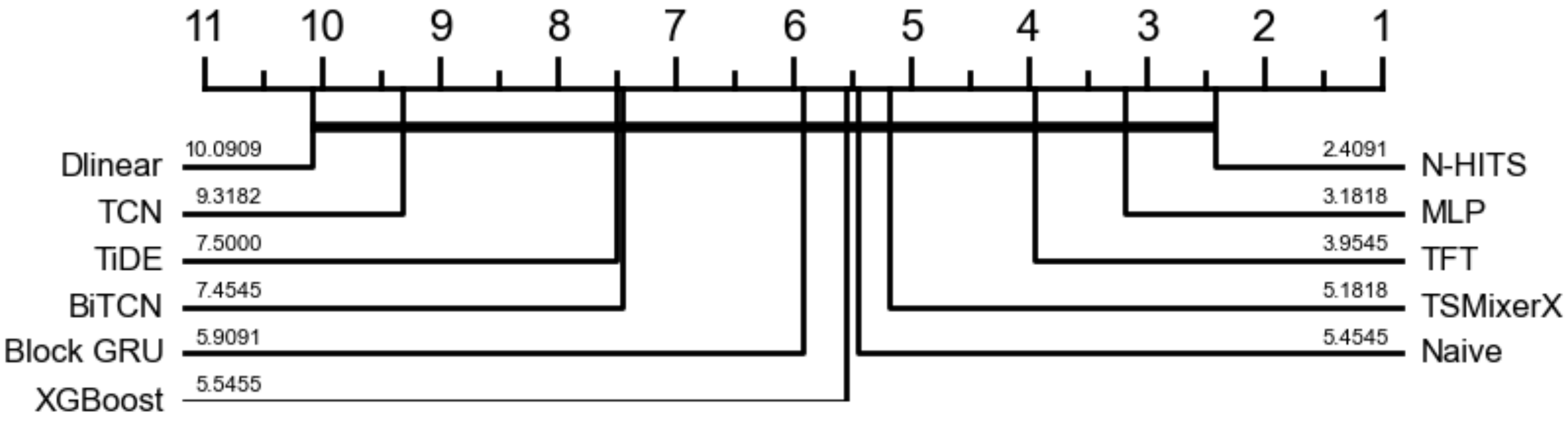

Figure 36: Critical difference diagrams for 11 multivariate TSF algorithms on the 11 datasets for FH=12

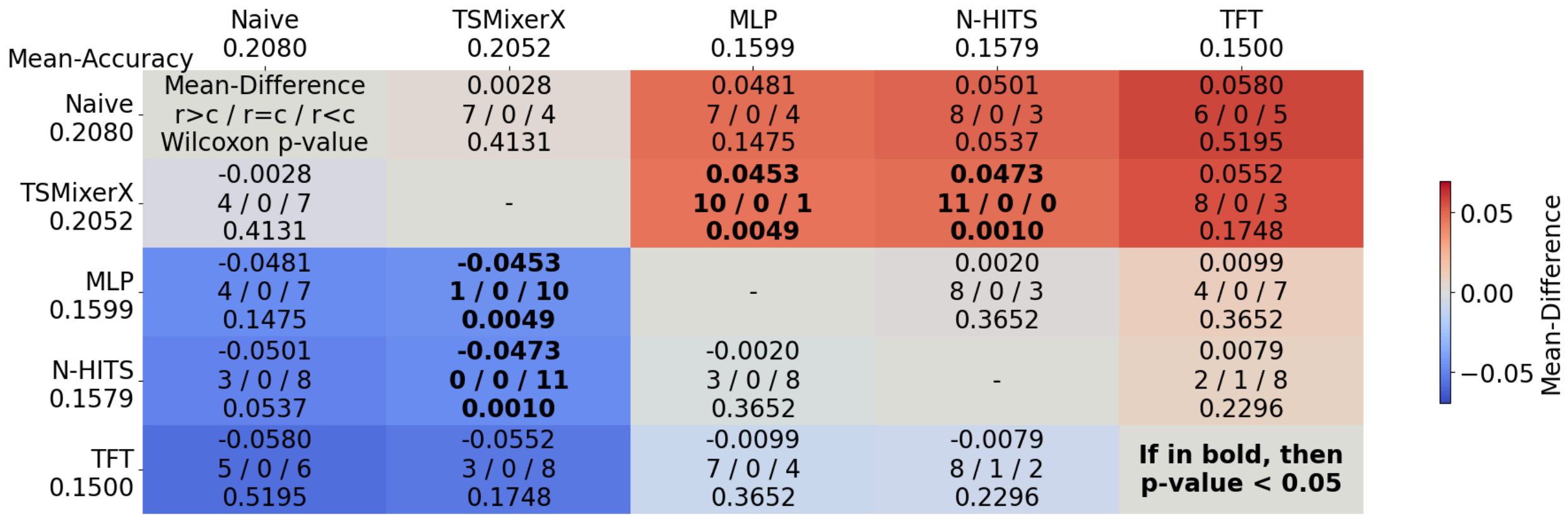

Figure 37: MCM for top five multivariate TSF algorithms on the 11 datasets for FH=12

# Exp Scenario 4: Long-term Multivariate TSF

The results for scenario four forecasting horizon of 96 are:

Table 37. Scenario four Performance summary of 11 algorithms on 11 datasets for FH=96

| Metric | Algorithms | | | | | | | | | | |
|---|---|---|---|---|---|---|---|---|---|---|---|
| | Naive | MLP | TCN | Block GRU | XGBoost | Dlinear | BiTCN | TSMixerX | TiDE | N-HITS | TFT |
| AVG ERR | 0.3149 | 0.2695 | 0.5071 | 0.2902 | 0.2245 | 0.9290 | 0.2538 | **0.2170** | 0.2204 | 0.2364 | 0.3029 |
| WIN | 1 | 0 | 0 | 0 | **4** | 0 | 3 | 2 | 0 | 1 | 0 |
| AVG Rank | 6.18 | 6.00 | 8.41 | 6.82 | 4.82 | 10.82 | 5.91 | **3.45** | 3.91 | 3.91 | 5.77 |

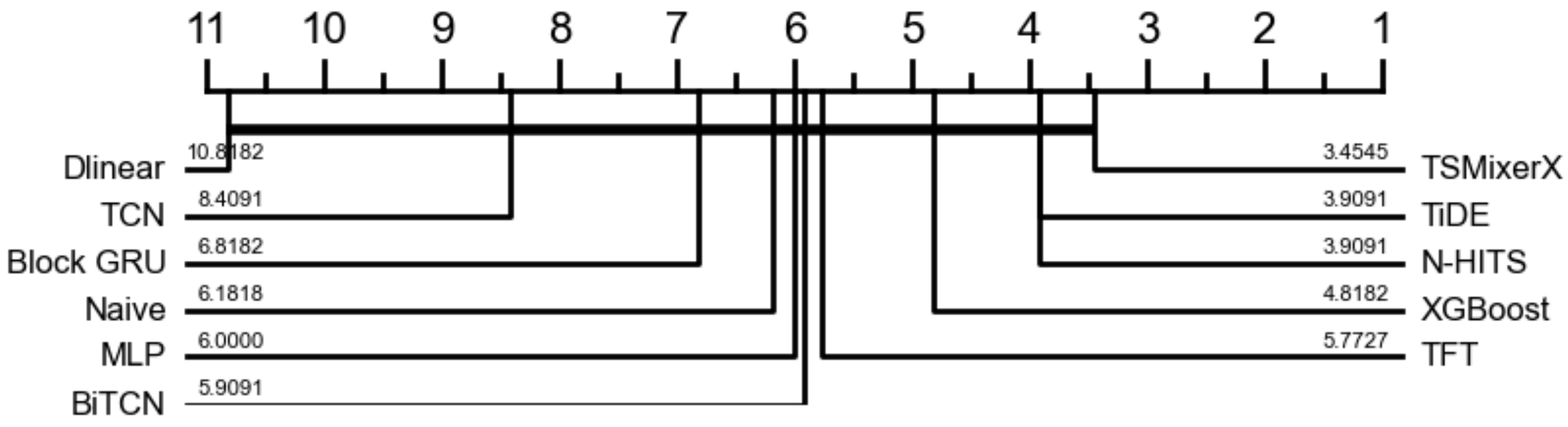

Figure 38: Critical difference diagrams for 11 multivariate TSF algorithms on the 11 datasets for FH=96

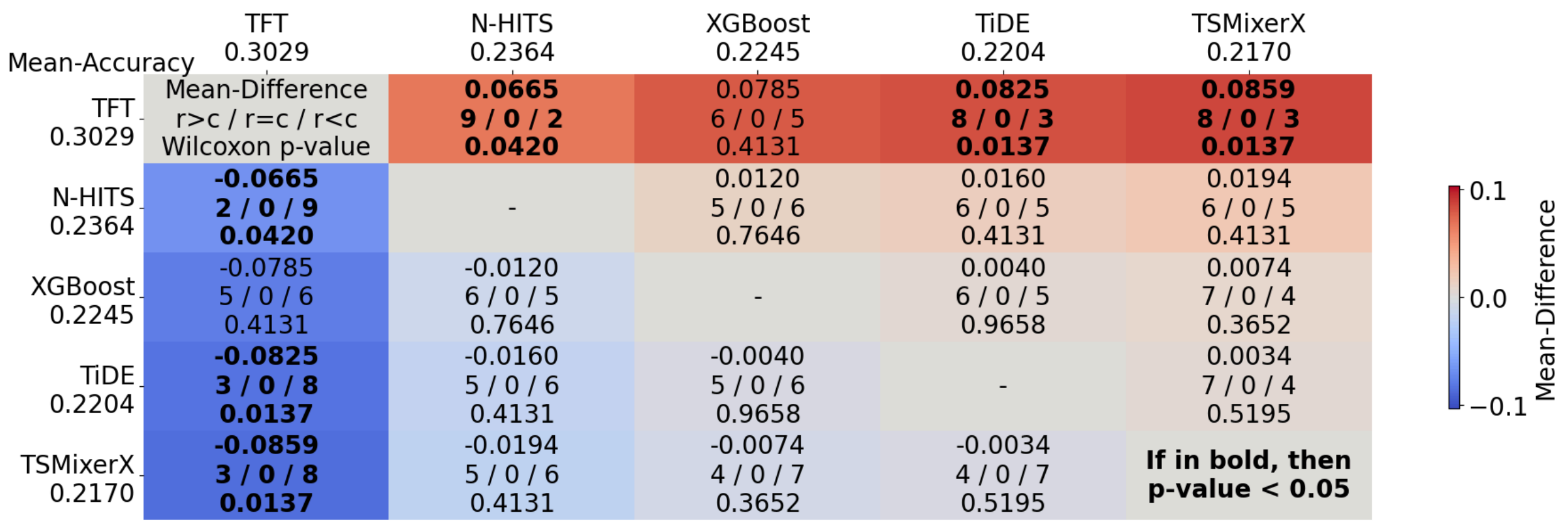

Figure 39: MCM for top five multivariate TSF algorithms on the 11 datasets for FH=96

The results for scenario four forecasting horizon of 288 are:

Table 38. Scenario four Performance summary of 11 algorithms on 11 datasets for FH=288

| Metric | Algorithms | | | | | | | | | | |
|---|---|---|---|---|---|---|---|---|---|---|---|
| | Naive | MLP | TCN | Block GRU | XGBoo st | Dlinear | BiTCN | TSMix erX | TiDE | N-HIT S | TFT |
| AVG ERR | 0.3405 | 0.2678 | 0.5269 | 0.2982 | 0.2335 | 0.9263 | 0.2792 | **0.2314** | 0.2396 | 0.2794 | 0.3336 |
| WIN | 0 | 1 | 0 | 0 | 4 | 0 | 3 | 0 | 1 | **1** | 1 |
| AVG Rank | 6.59 | 4.32 | 8.59 | 7.00 | 4.82 | 10.77 | 5.82 | **3.64** | 3.68 | 4.82 | 5.95 |

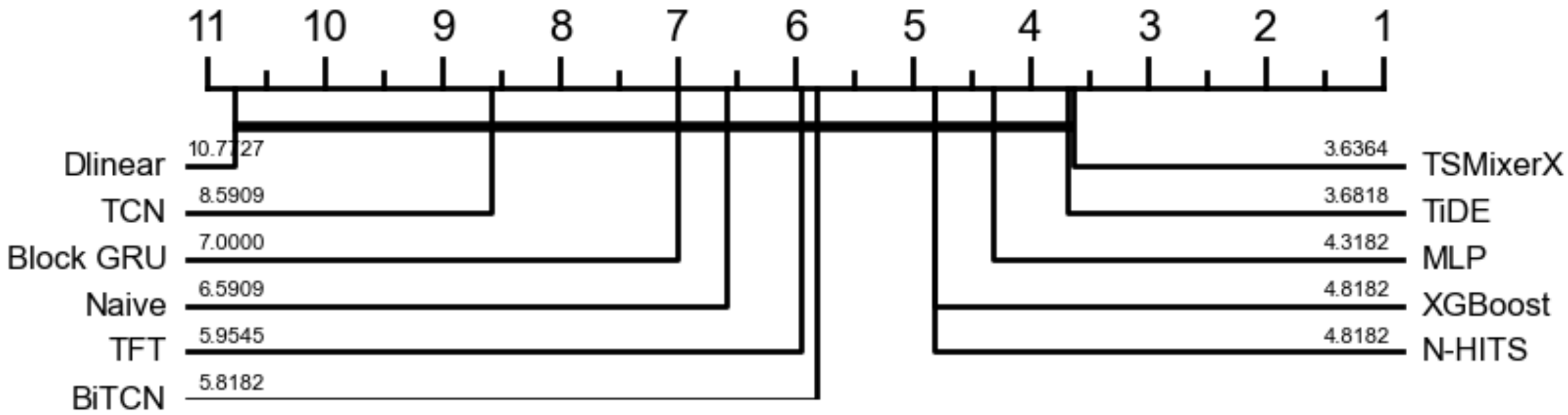

Figure 40: Critical difference diagrams for 11 multivariate TSF algorithms on the 11 datasets for FH=288

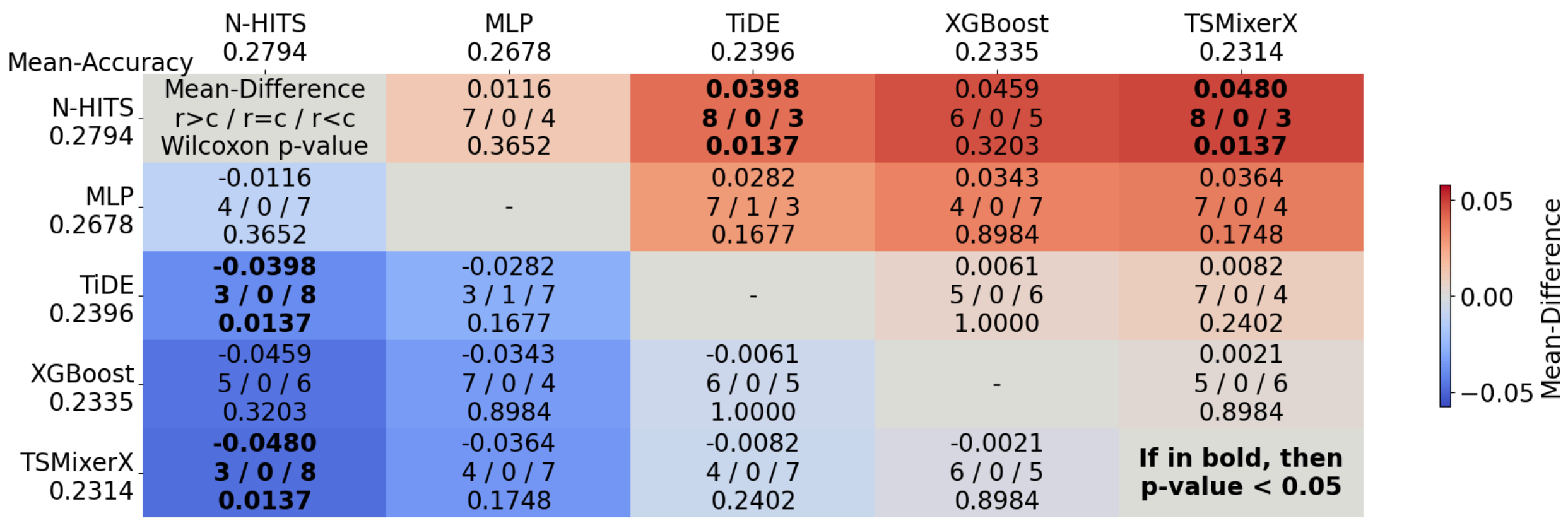

Figure 41: MCM for top five multivariate TSF algorithms on the 11 datasets for FH=288

The results for scenario four forecasting horizon of 672 are:

Table 39. Scenario four Performance summary of 11 algorithms on 11 datasets for FH=672

| Metric | Algorithms | | | | | | | | | | |
|---|---|---|---|---|---|---|---|---|---|---|---|
| | Naive | MLP | TCN | Block GRU | XGBoost | Dlinear | BiTCN | TSMixerX | TiDE | N-HITS | TFT |
| AVG ERR | 0.3461 | 0.2995 | 0.4569 | 0.3047 | 0.2659 | 0.9282 | 0.3076 | 0.2716 | **0.2591** | 0.3068 | 1.0000 |
| WIN | 2 | 0 | 1 | 0 | **3** | 0 | 3 | 1 | 1 | 0 | 0 |
| AVG Rank | 5.45 | 4.95 | 7.32 | 5.27 | 4.82 | 10.27 | 5.27 | 4.23 | **3.73** | 4.32 | 10.36 |

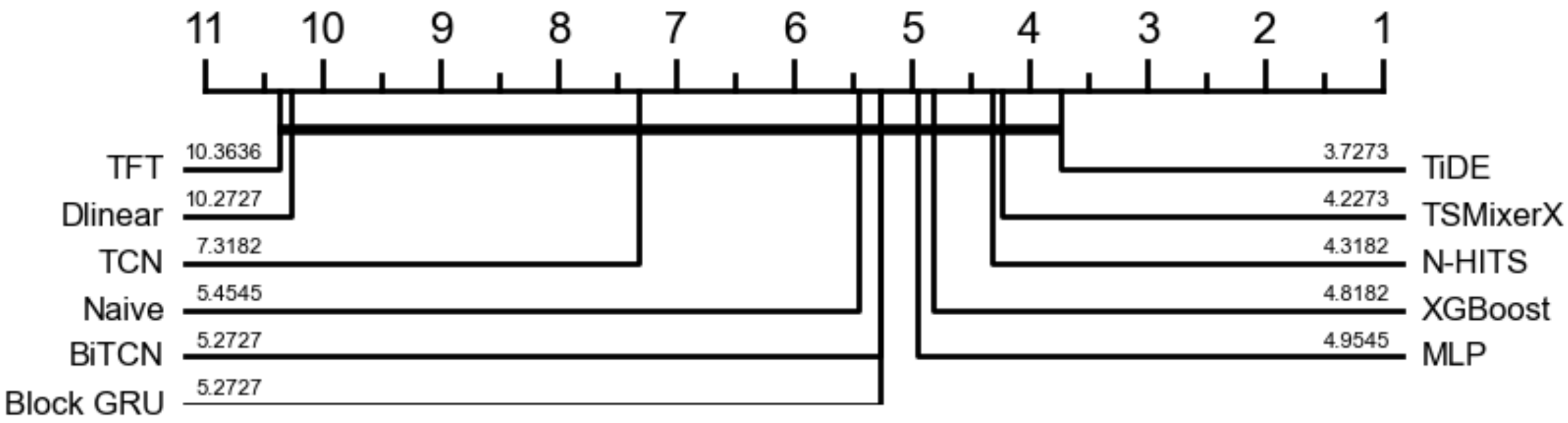

Figure 42: Critical difference diagrams for 11 multivariate TSF algorithms on the 11 datasets for FH=672

| Mean-Accuracy | N-HITS 0.3068 | MLP 0.2995 | TSMixerX 0.2716 | XGBoost 0.2660 | TiDE 0.2591 |
|---|---|---|---|---|---|
| N-HITS 0.3068 | Mean-Difference<br>r>c / r=c / r<c<br>Wilcoxon p-value | 0.0073<br>4 / 0 / 7<br>0.6377 | 0.0351<br>5 / 0 / 6<br>0.3203 | 0.0408<br>5 / 0 / 6<br>0.8311 | 0.0477<br>7 / 0 / 4<br>0.1016 |
| MLP 0.2995 | -0.0073<br>7 / 0 / 4<br>0.6377 | - | 0.0279<br>6 / 0 / 5<br>0.1748 | 0.0336<br>5 / 1 / 5<br>0.8239 | 0.0404<br>7 / 0 / 4<br>0.1748 |
| TSMixerX 0.2716 | -0.0351<br>6 / 0 / 5<br>0.3203 | -0.0279<br>5 / 0 / 6<br>0.1748 | - | 0.0057<br>4 / 1 / 6<br>0.6245 | 0.0126<br>6 / 0 / 5<br>1.0000 |
| XGBoost 0.2660 | -0.0408<br>6 / 0 / 5<br>0.8311 | -0.0336<br>5 / 1 / 5<br>0.8239 | -0.0057<br>6 / 1 / 4<br>0.6245 | - | 0.0069<br>6 / 0 / 5<br>0.7002 |
| TiDE 0.2591 | -0.0477<br>4 / 0 / 7<br>0.1016 | -0.0404<br>4 / 0 / 7<br>0.1748 | -0.0126<br>5 / 0 / 6<br>1.0000 | -0.0069<br>5 / 0 / 6<br>0.7002 | **If in bold, then p-value < 0.05** |

Figure 43: MCM for top five multivariate TSF algorithms on the 11 datasets for FH=672